\iftrue

\documentclass[11pt]{article}

\usepackage[utf8]{inputenc} 
\usepackage[T1]{fontenc}    
\usepackage{hyperref}       
\usepackage{url}            
\usepackage{amsfonts}       
\usepackage{microtype}      

\usepackage{subcaption}
\usepackage{algorithm}
\usepackage{algorithmic}

\usepackage{amsmath,amsfonts,bm}
\usepackage{amsthm}
\usepackage{mathtools}
\usepackage{xcolor}
\usepackage{nicefrac}       

\newcommand{\psd}{\succcurlyeq}
\newcommand{\tr}{\textrm{tr}}

\newcommand{\rank}{\textrm{rank}}

\allowdisplaybreaks

\newtheorem{assumption}{Assumption}[section]
\newtheorem{theorem}{Theorem}[section]

\newtheorem{lemma}[theorem]{Lemma}
{}

\newtheorem{prop}[theorem]{Proposition}

\makeatletter
\newtheorem*{rep@theorem}{\rep@title}
\newcommand{\newreptheorem}[2]{%
\newenvironment{rep#1}[1]{%
 \def\rep@title{#2 \ref{##1}}%
 \begin{rep@theorem}}%
 {\end{rep@theorem}}}
\makeatother

\newreptheorem{theorem}{Theorem}
\newreptheorem{lemma}{Lemma}
\newreptheorem{prop}{Proposition}
\newreptheorem{corollary}{Corollary}
\newreptheorem{definition}{Definition}

\newcommand{\ex}{\mathop{\mathbb{E}}\limits}

\newenvironment{itemize*}%
{\begin{itemize}[leftmargin=*,topsep=0pt]%
		\setlength{\itemsep}{0pt}%
		\setlength{\parskip}{0pt}}%
	{\end{itemize}}
\newenvironment{enumerate*}%
{\begin{enumerate}[leftmargin=*,topsep=0pt]%
		\setlength{\itemsep}{0pt}%
		\setlength{\parskip}{0pt}}%
	{\end{enumerate}}

\def\Figref#1{Figure~\ref{#1}}

\def\Secref#1{Section~\ref{#1}}

\def\eqref#1{equation~\ref{#1}}
\def\Eqref#1{Equation~(\ref{#1})}

\def\Thmref#1{Theorem~\ref{#1}}

\def\Lemref#1{Lemma~\ref{#1}}

\def\twoEqref#1#2{Equations (\ref{#1}) and (\ref{#2})}

\def\Propref#1{Proposition~\ref{#1}}

\def\Tableref#1{Table~\ref{#1}}

\def\1{\bm{1}}

\def\vzero{{\bm{0}}}
\def\vone{{\bm{1}}}

\def\va{{\bm{a}}}

\def\vu{{\bm{u}}}
\def\vv{{\bm{v}}}
\def\vw{{\bm{w}}}
\def\vx{{\bm{x}}}

\def\mA{{\bm{A}}}
\def\mB{{\bm{B}}}
\def\mC{{\bm{C}}}
\def\mD{{\bm{D}}}

\def\mI{{\bm{I}}}

\def\mM{{\bm{M}}}

\def\mP{{\bm{P}}}

\def\mS{{\bm{S}}}

\def\mU{{\bm{U}}}
\def\mV{{\bm{V}}}
\def\mW{{\bm{W}}}
\def\mX{{\bm{X}}}
\def\mY{{\bm{Y}}}

\DeclareMathAlphabet{\mathsfit}{\encodingdefault}{\sfdefault}{m}{sl}
\SetMathAlphabet{\mathsfit}{bold}{\encodingdefault}{\sfdefault}{bx}{n}

\def\gA{{\mathcal{A}}}

\def\gL{{\mathcal{L}}}

\def\gN{{\mathcal{N}}}
\def\gO{{\mathcal{O}}}

\def\gX{{\mathcal{X}}}
\def\gY{{\mathcal{Y}}}
\def\gZ{{\mathcal{Z}}}

\newcommand{\E}{\mathbb{E}}

\newcommand{\R}{\mathbb{R}}

\newcommand{\Var}{\mathrm{Var}}

\DeclareMathOperator*{\argmin}{arg\,min}

\usepackage{amssymb}
\usepackage{booktabs}       
\usepackage{multirow}
\usepackage{amsfonts}
\usepackage{dsfont}
\usepackage{mathtools}

\newcommand{\ns}[1]{\textcolor{orange}{[??: #1]}}

\newcommand{\trr}{\textmd{tr}}
\newcommand{\val}{\textmd{val}}
\newcommand{\te}{\textmd{test}}
\newcommand{\trtr}{\textmd{tr-tr}}
\newcommand{\trva}{\textmd{tr-val}}
\newcommand{\trtrtrva}{\textmd{\{tr-tr,tr-val\}}}
\newcommand{\rep}{\textmd{rep}}
\newcommand{\bad}{\textmd{bad}}
\newcommand{\good}{\textmd{good}}
\newcommand{\best}{\textmd{best}}

\newcommand{\Alg}{\gA}
\newcommand{\telambda}{\bar{\lambda}}
\newcommand{\nte}{\bar{n}_1}
\newcommand{\ntr}{n_1}
\newcommand{\nva}{n_2}
\newcommand{\Sto}{S}
\newcommand{\Str}{\Sto^{\trr}}
\newcommand{\Sva}{\Sto^{\val}}
\newcommand{\St}{\Sto}
\newcommand{\Xto}{\mX}
\newcommand{\Xtr}{\Xto^{\trr}}
\newcommand{\Xva}{\Xto^{\val}}
\newcommand{\Yto}{\mY}
\newcommand{\Ytr}{\Yto^{\trr}}
\newcommand{\Yva}{\Yto^{\val}}

\newcommand{\Ltest}{\gL^{\te}}
\newcommand{\Ltestrep}{\Ltest_{\telambda,\rep}}
\newcommand{\Ltrva}{\gL^{\trva}}
\newcommand{\Ltrvarep}{\Ltrva_{\lambda,\rep}}
\newcommand{\Ltrtr}{\gL^{\trtr}}
\newcommand{\Ltrtrrep}{\Ltrtr_{\lambda,\rep}}

\newcommand{\Lhatall}{\widehat{\gL}^{\trtrtrva}}
\newcommand{\Lhatallrep}{\Lhatall_{\lambda,\rep}}
\newcommand{\Lhattrva}{\widehat{\gL}^{\trva}}
\newcommand{\Lhattrvarep}{\Lhattrva_{\lambda,\rep}}
\newcommand{\Lhattrtr}{\widehat{\gL}^{\trtr}}
\newcommand{\Lhattrtrrep}{\Lhattrtr_{\lambda,\rep}}

\newcommand{\Clf}{\mW}
\newcommand{\clf}{\vw}
\newcommand{\param}{\theta}
\newcommand{\Param}{\Theta}
\newcommand{\repparam}{\theta^{\rep}}

\newcommand{\dist}{\rho}
\newcommand{\tclf}{{\vv}}
\newcommand{\veta}{\bm{\eta}}
\newcommand{\flayer}{{\mA}}
\newcommand{\slayer}{{\clf}}
\newcommand{\Astar}{{\flayer^*}}
\newcommand{\tdist}{\mu}
\newcommand{\tedist}{\bar{\tdist}}

\newcommand{\inner}{\textmd{in}}
\newcommand{\out}{\textmd{out}}
\newcommand{\instep}{T_{\inner}}
\newcommand{\outstep}{T_{\out}}
\newcommand{\inlr}{\eta_{\inner}}
\newcommand{\outlr}{\eta_{\out}}
\newcommand{\gradtrva}{\nabla^{\trva}}

\newcommand{\XtoU}{\Xto_{\mU}}
\newcommand{\XtoUp}{\Xto_{\mU^{\perp}}}

\usepackage{hyperref}
\hypersetup{colorlinks=true,citecolor=blue,linkcolor=blue}
\usepackage[parfill]{parskip}
\usepackage{natbib}
\usepackage[margin=0.9in]{geometry}
\usepackage{authblk}

\begin{document}

\title{A Representation Learning Perspective on the Importance of Train-Validation Splitting in Meta-Learning}
\date{}
\author[1]{{\large Nikunj Saunshi}}
\author[1]{\large Arushi Gupta}
\author[1]{\large Wei Hu}

\affil[1]{\small Department of Computer Science, Princeton University}
\affil[ ]{\texttt {\{nsaunshi, arushig, huwei\}@cs.princeton.edu}}
\affil[ ]{\small }
\affil[ ]{In proceedings of ICML 2021}
\maketitle

\begin{abstract}

An effective approach in meta-learning is to utilize multiple ``train tasks'' to learn a good initialization for model parameters that can help solve unseen ``test tasks'' with very few samples by fine-tuning from this initialization.
Although successful in practice, theoretical understanding of such methods is limited.
This work studies an important aspect of these methods: splitting the data from each task into train (support) and validation (query) sets during meta-training.
Inspired by recent work \citep{raghu2019rapid}, we view such meta-learning methods through the lens of representation learning and argue that the train-validation split encourages the learned representation to be {\em low-rank} without compromising on expressivity, as opposed to the non-splitting variant that encourages high-rank representations.
Since sample efficiency benefits from low-rankness, the splitting strategy will require very few samples to solve unseen test tasks.
We present theoretical results that formalize this idea for linear representation learning on a subspace meta-learning instance, and experimentally verify this practical benefit of splitting in simulations and on standard meta-learning benchmarks.
\end{abstract}

\section{Introduction}
\label{sec:intro}

Humans can learn from prior experiences, summarize them into skills or concepts and leverage those to solve new tasks with very few demonstrations \citep{ahn1993psychological}.
Since labeled data is often expensive to obtain, it would be desirable for machine learning agents to emulate this behavior of exploiting data from prior experiences to make solving new tasks sample efficient.
In this regard, we consider the meta-learning or learning-to-learn paradigm \citep{schmidhuber1987evolutionary,thrun1998learning}, where a learner utilizes data from many ``train'' tasks to learn a useful prior that can help solve new ``test'' tasks.
The goal is to do well on test tasks with way fewer samples per task than would be required without access to train tasks.
There is a rich history of successful methods in meta-learning and related fields of multi-task, lifelong and few-shot learning \citep{evgeniou2004regularized,ruvolo2013efficient,vinyals2016matching}.
The advent of the deep learning pipeline has led to more recent model-agnostic methods \citep{finn2017model} that learn a good initialization for model parameters by using train tasks (meta-training) and solve test tasks by fine-tuning from this initialization using little data (meta-testing).
Such methods have helped with many problems like few-shot classification in computer vision \citep{nichol2018first}, reinforcement learning \citep{finn2017model}, federated learning \citep{mcmahan2017communication}, neural machine translation \citep{gu2018meta}.

This empirical success has encouraged theoretical studies (cf. \Secref{sec:related}) of the statistical and optimization properties of such methods.
We are interested in the statistical aspect, where a meta-learner is evaluated based on the number of test task samples it requires to achieve small error.
Specifically, we focus our attention on the design choice of data splitting that is used in many meta-learning methods \citep{finn2017model,rajeswaran2019meta,raghu2019rapid}.
In this setting, a typical meta-learner learns a good initialization by 1) splitting the train task data into train (support) and validation (query) sets, 2) running the inner loop or {\em base-learner} of choice on the {\em train set} of the task starting from the current initialization, 3) minimizing the loss incurred on the {\em validation set} of the task by parameters from the inner loop, 4) updating the initialization in outer loop.
The learned initialization is used to solve an unseen {\em test task}, typically in the few-shot regime with very few samples available.
Here we study the sample complexity benefit offered by such train-validation (tr-val) splitting by arguing that it learns {\em low-rank \& expressive} representations.

We further compare this splitting strategy to the non-splitting variant where the entire task data is used for both, inner loop and outer loop updates \citep{nichol2018first,zhou2019efficient}.
Recent theoretical work \citep{bai2020important} makes the case for the non-splitting variant (referred to as tr-tr variant) with an analysis in the centroid learning setting for noiseless Gaussian linear regression and arguing that the non-splitting variant uses the available data more effectively in an asymptotic regime (cf. \Secref{subsec:data_splitting_bai}).
We instead take a representation learning perspective, inspired by \citet{raghu2019rapid} arguing that the power of modern meta-learning methods can be derived from the good representations they learn.
Our main contributions are summarized below:
\begin{itemize}
	\item {\bf Big picture:} We study the setting of representation learning, where the meta-learner trains a representation function (all weights before the linear classifier head) in the outer loop, while the inner loop learns a linear classifier on top of fixed representations.
	We show, theoretically and experimentally, that tr-val splitting helps with sample complexity of few-shot learning by inducing an {\em implicit regularization} that encourages the learning of {\em expressive low-rank representations}.
	On the theoretical side, we prove the sample efficiency of the tr-val objective for linear representations, going beyond previous theoretical results that require explicit regularization or low rank constraints.
	We also show potential failures of the tr-tr variant without explicit regularization, contrasting recent theoretical work \citep{bai2020important} that shows that tr-tr is better than tr-val in a different setting.
	Experiments support our hypothesis on simulations and standard benchmarks.

	\item {\bf Theory:} Our theoretical study is for a representation function that {\em linearly} maps $d$-dimensional inputs to $d$-dimensional\footnote{Dimensionality of representation can be smaller too. Larger makes the results stronger.} representations, while the inner loop performs ridge regression on fixed representation.
	We consider a meta-learning instance where tasks are linear regression problems with regressors on a $k$-dimensional subspace ($k\ll d$).
	Firstly we show that approximately minimizing the tr-val objective guarantees learning an expressive low-rank representation, which is precisely the underlying $k$-dimensional subspace.
	This results in a new task sample complexity of $\gO(k)$, i.e. only $\gO(k) \ll d$ samples are required from test tasks to achieve a small error.
	In contrast, no matter how well the tr-tr objective is optimized, one cannot guarantee learning a ``good'' representation and we may end up having an $\Omega(d)$ new task samples complexity.
	 By exploiting the practical design choice of tr-val splitting, we go beyond previous analysis that require explicit regularization or low-rank constraints on linear representations.
	 Our analysis allows for different number of samples per task in meta-training and testing, unlike previous excess risk based analyses \citep{baxter2000model,maurer2016benefit,denevi2018incremental}.

	\item {\bf Experiments:} Simulations on the subspace meta-learning instance verify that the tr-val splitting leads to smaller few-shot error by virtue of learning the right subspace.
	In contrast, the tr-tr objective trained using standard methods has poor performance on the same problem.
	We also test on standard few-shot classification benchmarks like Omniglot \citep{lake2015human} and MiniImageNet \citep{vinyals2016matching}, where again the tr-val objective turns out to be superior to tr-tr objective, for representation learning and a gradient based method implicit MAML \citep{rajeswaran2019meta}.
	We further find that while the tr-tr and tr-val representations are both expressive enough to separate all test classes with a lot of training data, the tr-val representation has lower effective rank.
	Thus although the theory is for the simplistic model of linear representations, it provide useful insights into the practice of meta-learning.
\end{itemize}

After discussing related work in \Secref{sec:related}, we present meta-learning preliminaries and describe the representation learning framework and the tr-val, tr-tr objectives in \Secref{sec:setup}.
\Secref{sec:meta_linrep} defines the subspace meta-learning instance for our theoretical analysis and formulates linear representation learning.
We present our theoretical results in \Secref{sec:theory} for the tr-tr and tr-val objectives, along with intuitive explanations and comparisons to previous resutls.
We present our experimental verifications and findings in \Secref{sec:exps}.
Proofs and additional experiment details are in the Appendix.


\section{Related Work}
\label{sec:related}

\textbf{Background and empirical work:}
Learning-to-learn or meta-learning has a rich history \citep{schmidhuber1987evolutionary,bengio1990learning,naik1992meta,caruana1997multitask,thrun1998learning}.
Existing deep learning based methods can be broadly classified as metric-based \citep{vinyals2016matching,snell2017prototypical}, model-based \citep{andrychowicz2016learning,ravi2017optimization} and more relevant to us, gradient-based \citep{finn2017model,nichol2018first} methods.
Recent empirical work \citep{raghu2019rapid} shows that most of the power of gradient-based methods like MAML \cite{finn2017model} can be derived by fixing representations and learning a linear classifier for new tasks.
Their proposed representation learning algorithm ANIL and other methods \citep{lee2019meta,bertinetto2018metalearning} show that  representation learning performs comparably to gradient-based methods on many benchmarks.
We note that \citet{oh2020does} highlights the importance of updating representations in the inner loop.
Recent works also try to demystify these methods by studying the role of depth \citep{arnold2019maml} and clustering of representations \citep{goldblum2020unraveling}.

\textbf{Theoretical work:}
The seminal work of \citet{baxter2000model} introduced a framework to study the statistical benefits of meta-learning, subsequently used to show excess risk bounds for ERM-like methods using techniques like covering numbers \citep{baxter2000model}, algorithmic stability \citep{maurer2005algorithmic} and Gaussian complexities \citep{maurer2016benefit}.
Sample complexity guarantees for {\em initialization-based} methods have been shown for inner loop losses that are convex in model parameters \citep{khodak2019adaptive,khodak2019provable,zhou2019efficient}; in particular for squared-error loss \citep{denevi2018learning,bai2020important}, a.k.a. centroid meta-learning.
The underlying structure exploited in these guarantees is the closeness of optimal task parameters for different tasks in some metric.

Another paradigm for analyzing meta-learning methods is representation learning \cite{maurer2016benefit,du2020few,tripuraneni2020theory}.
{\em Linear representation} learning is a popular playground to theoretically study various aspects of meta-learning \citep{argyriou2008convex,maurer2009transfer,maurer2013excess,bullins2019generalize,denevi2018incremental,du2020few,tripuraneni2020provable,tripuraneni2020theory}.
These analyses do not employ a train-validation split but instead require explicit regularizers or low-rank constraints on representations to show meaningful guarantees, which we do not need.
\citet{saunshi2020sample} show that Reptile \citep{nichol2018first} and gradient descent on linear representation learning (without data splitting or regularization) learn an underlying 1-dimensional subspace and show a separation from centroid meta-learning methods.

Other theoretical analyses show PAC-Bayes bounds \citep{pentina2014pac}, and regret bounds for life-long learning \citep{alquier2017regret} and study mixed linear regression \citep{kong2020meta}.
Methods like MAML are also studied from an optimization perspective \citep{rajeswaran2019meta,fallah2019convergence,wang2020global,collins2020does}, but we are interested in statistical aspects.

\textbf{Data splitting:}
The role of data splitting was theoretically considered in a recent work \citep{bai2020important} for the centroid learning problem.
While \citet{denevi2018learning} shows bounds for tr-val splitting, as noted in \citet{bai2020important}, their bounds cannot differentiate between tr-tr and tr-val objectives. 
\cite{wang2020guarantees} shows that tr-tr is worse than tr-val splitting for tuning learning rate using meta-learning.
We do a detailed comparison to data splitting and linear representation learning results in \Secref{subsec:comparison}.

\section{Meta-Learning Preliminaries}
\label{sec:setup}

\paragraph{Notation:}
\label{subsec:notation}
We use $\vx$ is used to denote a vector, $\Sto$ to denote a set, $\flayer$ to denote matrix, $\param$ to denote model parameters (like weights of neural network).
$\vx\sim\Sto$ denotes sampling an element $\vx$ from the uniform distribution over a finite set $\Sto$.
$\gN(\vzero_d,\Sigma)$ is Gaussian distribution with mean $\vzero_d$ and covariance $\Sigma$.
$I_d$ denotes a $d\times d$ identity matrix.
The ReLU function is denoted by $(x)_+ = \vone_{x>0}~x$.
For $\flayer\in\R^{d\times d}$, $\flayer^{\dagger}$ denotes the pseudo-inverse of $\flayer$ and $P_{\flayer} = (\flayer\flayer^{\dagger})$ denotes the projection matrix for the column span of $\flayer$.

\subsection{Data Splitting and Meta-Learning}
\label{subsec:splits}
We formalize the train-validation split that is used in practice and define the tr-tr and tr-val objective functions.
Let $\gZ=\gX\times\gY$  denote space of data points, where $\gX$ is the input space, $\gY$ is the label space.
Let $\param\in\Param\subseteq\R^D$ denote the model parameters (e.g. weights of a neural network) and $f_{\param}(\vx)\in\R^c$ denote the model prediction\footnote{For classification, model prediction is the logits and $c$ is number of classes. For regression, $c$ is target dimension.} for input $\vx\in\gX$.
A meta-learner has access to $T$ train tasks as datasets $\Sto_1,\dots, \Sto_T$, with each dataset $\Sto_t$ of $n$ points being split into $\Str_t=\{(\vx^{\trr}_{t,i}, y^{\trr}_{t,i})\}_{i=1}^{\ntr}$ and $\Sva_t=\{(\vx^{\val}_{t,i}, y^{\val}_{t,i})\}_{i=1}^{\nva}$, with $\Sto_t = \Str_t \cup \Sva_t$, where
$\Str_t$ and $\Sva_t$ refer to the train and validation splits respectively.
For a loss function $\ell$, e.g. logistic loss or squared error loss, we define the average loss incurred by model parameters $\param\in\Param$ on a dataset $\Sto$ as
\begin{align}
	\ell(\param;\Sto) &= \ex_{(\vx,y)\sim \Sto} \left[\ell(f_{\param}(\vx), y)\right]
\end{align}
Initialization-based meta-learning methods \citep{finn2017model} aim to learn an initialization $\param_0\in\Param$ for model parameters such that solving a task using its train data and the initialization $\param_0$ with an algorithm $\Alg$ will lead to parameters that do well on the test data for the task.
Formally, $\Alg : \Param\times\gZ^n\rightarrow\Param$ is an {\em inner algorithm} or base-learner, e.g. $\Alg(\param_0, \Sto)$ can run a few steps of gradient descent starting from $\param_0$ on the loss $\ell(\cdot;\Sto)$.
We now describe the train-validation (tr-val) and train-train (tr-tr) {\em outer algorithms} or meta-learners as ones that minimize the following objectives
\begin{align}
	\Lhattrva(\param; \Alg) &= \frac{1}{T} \sum\limits_{t=1}^T \ell(\Alg(\param, \Str_t) ; \Sva_t)\label{eqn:Ltrva}\\
	\Lhattrtr(\param; \Alg) &= \frac{1}{T} \sum\limits_{t=1}^T \ell(\Alg(\param,\St_t) ; \St_t)\label{eqn:Ltrtr}
\end{align}
Our tr-tr and tr-val objective definitions are similar to those from \cite{bai2020important}.
We now describe the representation learning objective and the corresponding inner algorithm $\gA$.


\subsection{Representation Learning}
\label{subsec:replearn}
As in \citet{raghu2019rapid}, we define a representation learning objective where the inner algorithm only learns a linear predictor on top of fixed learned representations.
Let $\repparam$ parametrize the representation function $f_{\repparam}:\gX\rightarrow\R^d$

\paragraph{Base-learner:}
For any task, we define the inner algorithm $\Alg_{\lambda,\rep}$ that only learns a linear classifier $\Clf\in\R^{d\times c}$ as
\begin{align}
	\ell_{\lambda,\rep}(\Clf; \repparam,\Sto) &= \ex_{(\vx,y)\sim \Sto} [\ell(\Clf^\top f_{\repparam}(\vx), y)]  +  \frac{\lambda}{2} \|\Clf\|_F^2\nonumber\\
	\Alg_{\lambda,\rep}(\repparam;\Sto) &= \argmin_{\Clf} \ell_{\lambda,\rep}(\Clf; \repparam,\Sto)
	\label{eqn:inner_alg}
\end{align}
\paragraph{Meta-learner:}
The corresponding tr-val and tr-tr objectives that use the base-learner $\Alg_{\lambda,\rep}$ are
\begin{align}
	\Lhatallrep(\repparam) \coloneqq \Lhatall(\repparam; \Alg_{\lambda,\rep})
	\label{eqn:replearn}
\end{align}
where $\Lhatall$ are defined in \twoEqref{eqn:Ltrva}{eqn:Ltrtr}.
The inner algorithm from \Eqref{eqn:inner_alg}, that uses $\|\cdot\|_F$ regularization, has been used in empirical work, for $\ell$ being the logistic and squared error losses in \citet{bertinetto2018metalearning} and the margin loss in \citet{lee2019meta}; also used in theoretical work \citep{saunshi2020sample}.
As evidenced in these works, minimizing $\Lhattrvarep$ performs very well in practice.
In the subsequent sections we will show, theoretically and experimentally, that $\Lhattrvarep$ learned representations are better than $\Lhattrtrrep$ for few-shot learning.

\section{Meta-Learning of Linear Representations}
\label{sec:meta_linrep}

In this section we discuss the subspace meta-learning instance for which we show theoretical guarantees.
We also define the tr-tr and tr-val objectives and the meta-testing metric for linear representation learning.

\subsection{Subspace Meta-Learning Instance}
\label{subsec:meta_instance}
We construct a simple meta-learning instance where each task is a regression problem.
We assume that the $T$ train tasks and the unseen test tasks will be sampled from an underlying distribution over tasks, as in many prior theoretical work \citep{baxter2000model,maurer2016benefit,denevi2018learning,bai2020important}.
Each task is a regression problem, with the input and target spaces being $\gX=\R^d$ and $\gY=\R$ respectively.
A task $\dist_{\tclf}$ is characterized by a vector $\tclf\in\R^d$ and has an associated distribution\footnote{We abuse notation and use $\dist_{\tclf}$ as a task and its distribution.} over $\gX\times\gY$ that satisfies
\begin{align}
	(\vx, y) \sim \dist_{\tclf} ~\equiv~ \vx\sim\gN(\vzero_d, I_d), y\sim\gN(\tclf^\top\vx, \sigma^2)
	\label{eqn:rho_def}
\end{align}
Thus the input distribution is standard normal, while the label is a linear function of input with some Gaussian noise\footnote{We only need $\ex[y|\vx]=\tclf^\top\vx$, $\Var[y|\vx]=\sigma^2 I_d$.} with variance $\sigma^2$ added to it.
We are interested in a meta-learning instance that only considers a subset of the tasks $\{\dist_{\tclf}\}$, namely for $\tclf\in\R^d$ that lie on an {\em unknown} $k$-dimensional subspace for some $k\ll d$.
Let $\Astar\in\R^{d\times k}$ be an orthogonal matrix\footnote{For simplicity. Results hold even if $\Astar$ is not orthonormal.} ($\Astar^\top\Astar = I_k$) whose columns define the basis for this subspace.
We define the distribution $\mu$ over tasks $\{\dist_{\tclf}\}$ that is supported on $\textmd{span}(\Astar)$ as follows
\begin{align}
	\dist_{\tclf} \sim\tdist ~\equiv~ \tclf \sim \gN(\vzero_d, \Astar\Astar^\top)
	\label{eqn:mu_def}
\end{align}
The optimal classifier for each regression task $\dist_{\tclf}$ is $\tclf$.
Since all classifiers $\tclf$ of interest will be on a $k$-dimensional subspace spanned by $\Astar$, we only need to know the projections of the $d$-dimensional inputs onto $\Astar$ to solve all tasks in $\tdist$.
Thus an expressive low rank linear representation $\vx \rightarrow \Astar\Astar^\top \vx$ exists for the inputs, but this subspace $\Astar$ is unknown to the meta-learner.
We note that a 1-dimensional subspace meta-learning instance ($k=1$) was considered in \citet{saunshi2020sample} to show guarantees for Reptile.

\vspace{-0.1in}

\paragraph{Meta-training dataset}
\label{subsec:meta_data}
We assume access to $T$ tasks $\dist_{\tclf_1}, \dots, \dist_{\tclf_T}$ sampled independently from the distribution $\tdist$ defined in \Eqref{eqn:mu_def}.
For the task $\dist_{\tclf_t}$ a dataset of $n$ points, $\Sto_t = (\Xto_t, \Yto_t)$, where $\Xto_t\in\R^{n\times d}, \Yto\in\R^{n}$ is sampled from $\dist_{\tclf_t}$.
Each dataset $\Sto_t = (\Xto_t,\Yto_t)$ is split into $\Str_t = (\Xtr_t,\Ytr_t)$ and $\Sva_t = (\Xva_t,\Yva_t)$ with sizes $\ntr$ and $\nva$ respectively, with $\ntr + \nva = n$.

\subsection{Linear Representation Learning}
\label{subsec:linreplearn}
We parametrize the predictor as a two-layer linear network, denoting the representation layer as $\flayer\in\R^{d\times d}$ and the linear classifier layer as $\slayer\in\R^d$.
The loss function is the squared-error loss $\ell(\hat{y}, y) = (\hat{y} - y)^2$.
We now describe the base-learner, meta-learners and meta-testing metric.

\paragraph{Base-learner:}
Recall that for each task $t\in[T]$, the task data of size $\Sto_t=(\Xto_t, \Yto_t)$ of size $n$ is split into $\Str_t=(\Xtr_t, \Ytr_t)$ and $\Sva_t=(\Xva_t, \Yva_t)$ of sizes $\ntr$ and $\nva$ respectively.
For the squared error loss, the inner loop for data $\Sto=(\Xto,\Yto)$ reduces to the following:
\begin{align}
	\ell_{\lambda,\rep}(\slayer; \flayer,\Sto) &= \frac{1}{n}\|\Xto\flayer\slayer - \Yto\|^2  +  \lambda \|\slayer\|_2^2
	\nonumber\\
	\slayer_{\lambda}(\flayer;\Sto) &= \argmin_{\slayer\in\R^d} ~\ell_{\lambda,\rep}(\slayer; \flayer,\Sto)
	\label{eqn:replearn_inner_alg}
\end{align}
%
%
%
\paragraph{Meta-learner:}
The tr-val and tr-tr objectives to learn representation layer $\flayer$ on $T$ train tasks are described below:
\begin{align}
	\Lhattrvarep(\flayer;(\ntr,\nva)) 
	&= \frac{1}{T\nva} \sum\limits_{t=1}^T \left\|\Xva_t\flayer\slayer_{\lambda}(\flayer; \Str_t) - \Yva_t\right\|^2
	\label{eqn:replearn_trva}\\
	\Lhattrtrrep(\flayer;n) 
	&= \frac{1}{Tn} \sum\limits_{t=1}^T \left\|\Xto_t\flayer\slayer_{\lambda}(\flayer; \Sto_t) - \Yto_t\right\|^2
	\label{eqn:replearn_trtr}
\end{align}
We note that the linear representation learning literature considers a similar objective to $\Lhattrtrrep$ \citep{bullins2019generalize,denevi2018incremental,du2020few,tripuraneni2020theory}, albeit with either a Frobenius norm constraint on $\flayer$ or a rank constraint with $\flayer\in\R^{d\times k}$ rather than $\R^{d\times d}$.
We show that such a constraint is not needed for the $\Lhattrvarep(\flayer)$ objective, since it implicitly induces such a regularization.

\paragraph{Meta-testing/evaluation}
We evaluate the learned representation layer $\flayer$ on a test task $\dist_{\tclf}$ with only $\nte$ train samples from the task.
The inner algorithm first uses $\flayer$ to learn $\slayer_{\telambda}(\flayer;\Sto)$\footnote{The regularization parameter $\telambda$ can also be different from $\lambda$.} from \Eqref{eqn:replearn_inner_alg} on $\nte$ samples $\Sto$ sampled from $\dist_{\tclf}$.
The final predictor $\flayer\slayer_{\telambda}(\flayer;\Sto)\in\R^d$ is then evaluated on $\dist_{\tclf}$: $\E_{(\vx,y)\sim\dist_{\tclf}}\|\vx^\top\flayer\slayer - y\|^2 = \sigma^2 + \|\flayer\slayer - \tclf\|^2$.
Formally we define the meta-testing metric as an average loss over a test tasks sampled from $\tedist$
\begin{align}
	\Ltestrep(\flayer;\nte) = \ex_{\dist_{\tclf}\sim\tedist} \left[\ex_{\Sto\sim\dist_{\tclf}^{\nte}} \left[\|\flayer\slayer_{\telambda}(\flayer;\Sto) - \tclf\|^2\right]\right] + \sigma^2
	\label{eqn:meta_test}
\end{align}
This metric is similar to the one used in prior work \citep{baxter2000model,maurer2016benefit,denevi2018incremental,saunshi2020sample}, however the distribution over test tasks $\tedist$ need not be the same as $\tdist$.
All we need is that test tasks lies on the subspace spanned on $\Astar$.
\begin{assumption}
	For all $\dist_{\tclf}\sim\tedist$, $\Astar\Astar^\top\tclf = \tclf$
\end{assumption}
Note also that samples per test task $\nte$ can also be different from $n$ or $\ntr$ available during meta-training.

\section{Theoretical Results}
\label{sec:theory}

We discuss theoretical results for linear representation learning on the meta-learning instance defined in \Secref{subsec:meta_instance}.
First we show that trying to minimize the tr-tr objective without any explicit regularization can end up learning a full rank representation, thus leading to $\Omega(d)$ sample complexity on a new task.
While this is not too surprising or hard to show, it suggests that for the non-splitting variant to succeed, it needs to depend on either some explicit regularization term or low-rank constraints as imposed in prior work, or rely on inductive biases of the training algorithm.
We then show our main result for the tr-val objective showing that minimizing it implicitly imposes a low-rank constraint on representations and leads to new task sample complexity of $\gO(k)$.
Proving this results entails deriving a closed form expression for the asymptotic tr-val objective using symmetry in the Gaussian distribution and properties of the inverse Wishart distribution \citep{mardia1979multivariate}.

\subsection{Results for Train-Train Objective}
\label{subsec:tr_tr_results}
For theoretical results in this section, we look at the asymptotic limit as the number of train tasks $T$ goes to infinity, but with each task having just $n$ samples\footnote{Standard benchmarks that solve $N$-way $K$-shot tasks with $C$ classes can effective produce ${C}\choose{N}$ tasks which can be large}.
\begin{align}
	\Ltrtrrep(\flayer;n) = \lim_{T\rightarrow\infty} \Lhattrtrrep(\flayer;n)
	 \label{eqn:Ltrtr_limit}
\end{align}
This limit gives an indication of the best one could hope to do with access to many tasks with small amount of data ($n$) per task.
Note that the results in this subsection can be extended to the non-asymptotic case of finite $T$ as well.

The goal of $\Ltrtrrep$ is to learn a first representation layer $\flayer$ such the linear classifier $\slayer_{\lambda}(\flayer; \Sto_t)$ learned using the task data $\Sto=(\Xto,\Yto)$ can fit the same task data $\Sto$ well.
While a low rank layer ($\flayer = \Astar$) can perfectly predict the signal component of the the labels ($\Xto_t\tclf_t$) for tasks in $\tdist$, fitting the random noise in the labels ($\Yto_t - \Xto_t\tclf_t$) requires a representation that is as expressive as possible.
However as we show subsequently, a full rank representation layer is bad for the final meta-evaluation $\Ltestrep$.
We formalize this idea below and show the existence of such ``bad'' full rank representations that make the $\Ltrtrrep$ arbitrarily small.

%

\begin{theorem}
\label{thm:tr_tr_bad_sol}
For every $\lambda,n>0$,  for every $\tau>0$, there exists a ``bad'' representation layer $\flayer_{\bad}\in\R^{d\times d}$ that satisfies $\Ltrtrrep(\flayer_{\bad};n)\le\inf\limits_{\flayer\in\R^{d\times d}} \Ltrtrrep(\flayer;n) + \tau$, but has the following lower bound on meta-testing loss
	\begin{align*}
		\inf_{\telambda>0} \Ltestrep(\flayer_\bad;\nte) - \sigma^2
		&\ge \min\left\{1 - \frac{\nte}{d(1+\sigma^2)}, \frac{d\sigma^2}{(1+\sigma^2)\nte}\right\}
	\end{align*}
\end{theorem}
The fixed error of $\sigma^2$ is unavoidable for any method due to noise in labels.
In the few-shot regime of $\nte \ll d$, the lower bound on the error is close to 1.
To get an error of $\sigma^2+\epsilon$ on a new test task for a very small $\epsilon$, the number of samples $\nte$ for the new tasks must satisfy $\nte =\Omega(\frac{d}{\epsilon})$.

\paragraph{Implications}
\Thmref{thm:tr_tr_bad_sol} implies that no matter how close $\Ltrtrrep(\flayer)$ gets to the optimal value, one cannot guarantee good performance for $\flayer$ in meta-testing $\Ltestrep(\flayer)$.
This does not, however, rule out the existence of ``good'' representations that have a small $\Ltrtrrep$ and small $\Ltestrep$.
However our result suggests that simply trying to minimize $\Ltrtrrep$ may not be enough to learn good representations.
We show in \Secref{sec:exps} that standard meta-learning algorithms with a tr-tr objective can end up learning bad representations, even if not the worst one.
We demonstrate this for the above meta-learning instance and also for standard benchmarks, thus suggesting that our result has practical bearing.

Prior work has shown guarantees for linear representation learning algorithms without any data splitting.
The main difference is that these methods either add a norm regularization/constraint on the representation layer \citep{argyriou2008convex,maurer2009transfer,maurer2013excess,bullins2019generalize,denevi2018incremental} to encourage low-rankness or an explicit low rank constraint on the representation \citep{du2020few,tripuraneni2020provable,tripuraneni2020theory}.
We show that even for this simple meta-learning instance, lack of rank constraint or norm regularization can lead to terrible generalization to new tasks, as also demonstrated in experiments in \Secref{sec:exps}.
In \Tableref{table:withexpreg} we find that adding a regularization term on the representation does lead to much better performance, albeit still slightly worse that the tr-val variant.

\paragraph{Proof sketch}
We prove \Thmref{thm:tr_tr_bad_sol} by first arguing that $\lim_{\kappa\rightarrow\infty} \Ltrtrrep(\kappa I_d; n) = \inf\limits_{\flayer\in\R^{d\times d}} \Ltrtrrep(\flayer;n)$.
Thus picking $\flayer_{\bad}=\kappa I_d$ for a large enough $\kappa$ suffices to get an $\epsilon$-good $\Ltrtrrep$ solution.
On the other hand since $\kappa I_d$ treats all directions alike, it does not encode anything about the $k$-dimensional subspace $\Astar$.
The test task, thus, reduces to linear regression on $d$-dimensional isotropic Gaussians data, which is known to have $\Omega\left(\frac{d}{\epsilon}\right)$ sample complexity to get loss at most $\epsilon$.
The lemma below shows that $\kappa I_d$ converges to the infimum of $\Ltrtrrep$, since a lower rank layer will not fit the noise in the train labels well enough.
\begin{lemma}
\label{lem:tr_tr_full_rank}
	For every $\lambda>0$ and $\flayer\in\R^{d\times d}$ with rank $r$,
	\begin{align*}
		\Ltrtrrep(\flayer;n) \ge \lim_{\kappa\rightarrow\infty}\Ltrtrrep(\kappa&\flayer;n) \ge \sigma^2\frac{(n-r)_+}{n}\\
		\text{  \&  }~\lim_{\kappa\rightarrow\infty}\Ltrtrrep(\kappa I_d;n) &= \sigma^2\frac{(n-d)_+}{n}
	\end{align*}
\end{lemma}
We note that this result can be shown without needing Gaussianity assumptions.
The ideas from this result are applicable beyond linear representations.
In general, making the $\Ltrtrrep$ objective small can be achieved by learning a very expressive representation can fit the training data perfectly.
There is nothing in the objective function that encourages forgetting redundant information in the representation.
We now present the main results for tr-val objective.

\subsection{Result for Train-Validation Objective}
\label{subsec:tr_va_results}

We again look at the asymptotic limit as the number of train tasks $T$ goes to infinity, with each task having an $(\ntr, \nva)$ train-validation split.
\begin{align}
	 \Ltrvarep(\flayer;(n_1,n_2)) = \lim_{T\rightarrow\infty} \Lhattrvarep(\flayer;(n_1,n_2))
	 \label{eqn:Ltrva_limit}
\end{align}
Here we demonstrate two benefits of the tr-val.

\paragraph{Correct objective to minimize}
The first result formalizes the intuitive benefit of the $\Ltrvarep$ objective: if number of train samples in train tasks and test tasks are the same, i.e. $\ntr=\nte$, then $\Ltrvarep$ is the the correct objective to minimize
\begin{prop}
\label{prop:Ltrva_equal_Ltest}
	$\Ltrvarep(\cdot;(\ntr,\nva))$ and $\Ltestrep(\cdot;\nte)$ are equivalent if $~\nte = \ntr$ and $\telambda=\lambda$
\end{prop}
Thus when $\ntr=\nte$, $\Lhattrvarep$ is an unbiased estimate of $\Ltest_{\lambda,\rep}$ and minimizing it makes sense.
This is easy to prove by exploiting the independence of $\Str$ and $\Sva$ in the expression for $\Lhattrvarep$ in \Eqref{eqn:Ltrva}.
However when $\nte\neq\ntr$ or $\lambda\neq\telambda$, the objective functions are different, so it is apriori unclear why minimizing $\Ltrvarep$ should be a good idea if we care about $\Ltest_{\lambda,\rep}$.
This is what we handle next.

\paragraph{Low rank representation}
We show here that for $\lambda=0$, the minimizers of $\Ltrvarep$ are the same for almost all $\ntr$ and are in fact rank-$k$ representations that span the subspace of interest $\Astar$.
This low rankness also ensures that the sample complexity of a new task is $\gO\left(\frac{k}{\epsilon}\right)$.
\begin{theorem}
\label{thm:trva_result1_robust}
	Let $\lambda=0$. If $\ntr\ge c_1k$, $\sigma^2\in (0,c_2)$, $\tau\le c_3\frac{\sigma^2}{\ntr}$ for small constants $c_1,c_2,c_3$, then any $\flayer_{\good}\in\R^{d\times d}$ that is $\tau$-optimal, i.e. $\Ltrvarep(\flayer_{\good};(\ntr,\nva)) \le \inf\limits_{\flayer\in\R^{d\times d}} \Ltrvarep(\flayer;(\ntr,\nva)) + \tau$, will satisfy
	\begin{align*}
		\rank(\flayer_{\good}) = k,~~~ \|P_{\flayer_{\good}}\Astar - \Astar\|^2 \le \tau
	\end{align*}
	and the meta-testing performance $\nte>2k+2$ satisfies
	\begin{align*}
		\inf_{\telambda\ge0}&\Ltestrep(\flayer_{\good};\nte) - \sigma^2 \le 2\tau + \sigma^2\frac{2k}{\nte}
	\end{align*}
\end{theorem}
Thus under mild conditions on the noise in labels and size of train split during meta-training, this result shows that minimizers ($\tau=0$) of $\Ltrvarep$ will have meta-testing loss smaller than $\epsilon$ with $\nte=\gO\left(\frac{k}{\epsilon}\right)$ samples per test task, even when $\ntr\neq\nte$.
It does so by learning a rank-$k$ representation that spans $\Astar$.
Compared to the result from \Thmref{thm:tr_tr_bad_sol}, the bad representation $\flayer_{\bad}$ learned using the $\Ltrtrrep$ objective will need $\nte=\Omega\left(\frac{d}{\epsilon}\right)$ samples to get the same error.
The above result also shows that getting close enough to optimality is also sufficient for good meta-testing performance; full version of the theorem and its proof can be found in \Secref{asec:trva}.
We reiterate that we did not need any explicit regularization to show this sample complexity benefit.
The proof sketch below highlights the implicit regularization in $\Ltrvarep$ and discusses the role of label noise $\sigma$ and number of train task samples $\ntr$ in ensuring the desired low-rankness.

\paragraph{Proof sketch}
The crux of the proof is in a stronger result in \Thmref{thm:trva_closedform} that provides a {\em closed form expression} for $\Ltrvarep(\flayer;(\ntr,\nva))$.
For any representation $\flayer$, let $\flayer = \mU\mS\mV^\top$ be the singular value decomposition (SVD) with $\mU,\mV\in\R^{d\times r}, \mS\in\R^{r\times r}$ and $r=\rank(\flayer)$.
The expression for $\Ltrvarep(\flayer;(\ntr,\nva))$ from \Thmref{thm:trva_closedform} is
\begin{align*}
	&\Ltrvarep(\flayer) \approx \underbrace{\min\left\{1,\frac{\ntr}{r}\right\}\|P_{\flayer}^\perp\Astar\|^2}_{\bm \alpha_1(\flayer)} + \underbrace{\frac{(r-\ntr)_+}{r}}_{\bm \alpha_2(\flayer)}\\
	& + \underbrace{\sigma^2\left[\mathds{1}_{r>\ntr+1}\frac{\ntr}{r - \ntr - 1} + \mathds{1}_{r<\ntr-1}\frac{r}{\ntr - r - 1}\right]}_{\bm \alpha_3(\flayer)}
\end{align*}
where $P_{\flayer}$ is the projection matrix for the column span of $\flayer$ and $P_{\flayer}^\perp = I_d - P_{\flayer}$.
We now look at how each term contributes to the final result.

1) The term $\bm{\alpha_1(\flayer)}$ penalizes inexpressiveness of the representation through the $\|P_{\flayer}^\perp\Astar\|^2$ term. When $r\ge k$, this can be made 0 by letting $\flayer$ span all the $k$ directions in $\Astar$; thus this term also encourages the rank being at least $k$.

2) The second term $\bm{\alpha_2(\flayer)}$ can be as low as 0 only when the rank satisfies $r\le\ntr$.
In the noiseless setting ($\sigma=0$), $\Ltrvarep$ already encourage learning an {\em expressive representation} with rank between $k$ and $\ntr$.
Thus if the number of samples $\ntr$ in the train split of train tasks is small, we are already guaranteed a low rank representation and new task sample complexity of $\gO(\ntr)$; thus it might be beneficial to use fewer samples in the train split and more in the validation split, similar to the result in \cite{bai2020important}.

3) Unlike $\sigma=0$, $\sigma>0$ can differentiate in $r\in[k,\ntr]$ through the $\bm{\alpha_3(\flayer)}$ term. Under the mild assumptions on $\sigma$ and $\ntr$, this term ensures that the loss is minimized exactly at $r=k$, and thus $\|P_{\flayer}^\perp\Astar\|^2 = 0$ from $\bm{\alpha_1(\flayer)}$.
Thus label noise helps learn low rank representations even for large $\ntr$.

This leads to the first part of \Thmref{thm:trva_result1_robust} showing that any minimizer $\flayer_{\good}$ of the tr-val objective will be low-rank and expressive.
The extension to $\tau$-optimal solutions also follows the same strategy.
The second part showing that such a representation $\flayer_{\good}$ can reduce the sample complexity of a new test task follows by noting that $\flayer_{\good}$ effectively projects inputs onto the $k$-dimensional subspace $\Astar$ and thus lets us use the upper bound on sample complexity for linear regression on $k$ dimensions, which is $\gO(\frac{k}{\epsilon})$.


The above result holds for any output dimensionality $D\ge k$ of representation $\flayer\in\R^{d\times D}$; the tr-val objective can automatically adapt and learn low rank and expressive representations, unlike the tr-tr objective.
This is demonstrated in experiments on Omniglot and MiniImageNet in \Secref{sec:exps} where increasing dimensionality of representations improves the performance of tr-val representations, but the added expressivity hurts the tr-tr performance (cf. Tables~\ref{table:widthtable} \& \ref{table:miniwidth}).
\Thmref{thm:trva_closedform} is proved using symmetry in Gaussian distribution and properties of Wishart distributions.
We now make a closer comparison to prior work on linear representation learning and role of tr-val splitting.

\subsection{Comparison to Prior Work and Discussions}
\label{subsec:comparison}

\textbf{Data splitting:}
\label{subsec:data_splitting_bai}
\citet{bai2020important} analyze the tr-tr and tr-val objectives for linear centroid meta-learning, where the prediction function for parameters $\param\in\R^D$ on input $\vx$ is $f_{\param}(\vx) = \param^\top\vx$.
Using random matrix theory, they show in the realizable setting that tr-tr objective learns parameters that have better new task sample complexity than tr-val parameters, by a constant factor, because it can use the data from train tasks more efficiently.
For experiments they use iMAML \citep{rajeswaran2019meta} for tr-val method and MiniBatchProx \citep{zhou2019efficient} for the tr-tr method and show that tr-tr outperforms tr-val on standard benchmarks.
We note that these findings do not contradict ours, since MiniBatchProx does not minimize the vanilla tr-tr objective method but adds a regularization term. 
In \Secref{subsec:imaml_exp} we verify that a tr-tr variant of iMAML without regularization indeed does poorly on the same dataset.

\textbf{Linear representation learning:}
We have already discussed differences in settings from many linear representation learning guarantees: lack of data splitting and presence of regularization or rank constraints.
Our work incorporates a practical choice of data splitting in linear representation learning theory.
Another such result is from \citet{saunshi2020sample} that employs a trajectory based analysis to show that Reptile and also gradient descent $\Ltrtr$ learn good representations without explicit regularization or data splitting.
This provides evidence that inductive biases of training algorithms is another way to avoid high rank representations.
Their result is for the subspace meta-learning instance with $k=1$ and $\ntr\rightarrow\infty$.
While interesting, it is not clear how to extend it to the more interesting case of $\ntr=\gO(k)$ that our analysis for $\Ltrvarep$ can handle.

\section{Experiments}
\label{sec:exps}

\begin{table}[!t] 
  \caption{Performance of meta-learning models trained using the tr-val and tr-tr objectives on Omniglot few-shot classification. RepLearn method is described in \Secref{asec:exps} and iMAML method is from \cite{rajeswaran2019meta}
  }
  \label{table:omnitable}
  \small
  \medskip
  \centering
  \small
  \begin{tabular}{lc|ccc}
    \toprule
    				&&  5-way 1-shot	& 20-way 1-shot \\
    \midrule
    \multirow{2}{*}{tr-val}	& RepLearn & 97.30 $\pm$ 0.01 & 92.30 $\pm$ 0.01 \\
    					& iMAML & 97.90 $\pm$ 0.58 & 91.00 $\pm$ 0.54\\
    \midrule
    \multirow{2}{*}{tr-tr}	& RepLearn & 89.00 $\pm$ 0.01 & 88.20 $\pm$ 0.01 \\
    					& iMAML & 49.20 $\pm$ 1.91 & 18.30 $\pm$ 0.67\\
    \bottomrule
  \end{tabular}
\end{table}

\begin{table}[!t] 
  \caption{Performance of linear representations $\flayer$ on simulation dataset with $d=50, k=5, \sigma=0.5$ (refer to \Secref{subsec:meta_instance}). tr-tr and tr-val correspond to representations trained using $\Lhattrtrrep$ and $\Lhattrvarep$ respectively. Representations are evaluated on test tasks with $\nte$ samples using $\Ltestrep(\flayer;\nte)$; ($\telambda$ is tuned for all entries). Includes performance of $\flayer=I_d$ and $\flayer = \Astar\Astar^\top$ that are the ``worst'' and ``best'' representations for this task respectively.
  }
  \label{table:simul}
  \medskip
  \centering
  \small
  \begin{tabular}{c|ccc}
    \toprule
    			& \multicolumn{3}{c}{$\Ltestrep(\cdot;\nte)$}\\
    $\flayer$ 	& $\nte=5$	& $\nte=15$ 	& $\nte=25$\\
    \midrule
    $I_d$				& 1.10 $\pm$ 0.09	& 0.99 $\pm$ 0.08	& 0.97 $\pm$ 0.17 \\
    tr-tr ($\lambda=0$)	& 1.04 $\pm$ 0.08	& 0.85 $\pm$ 0.06 	& 0.82 $\pm$ 0.06\\
    tr-tr ($\lambda=1$) 	& 0.94 $\pm$ 0.07	& 0.69 $\pm$ 0.05	& 0.66 $\pm$ 0.05\\
    tr-tr ($\lambda=10$) 	& 0.92 $\pm$ 0.07	& 0.69 $\pm$ 0.05	& 0.69 $\pm$ 0.05\\
    \midrule
    tr-val ($\lambda=0$) & 0.72 $\pm$ 0.06	& 0.40 $\pm$ 0.03	& 0.38 $\pm$ 0.03\\
    $\Astar\Astar^\top$ & 0.70 $\pm$ 0.05	& 0.38 $\pm$ 0.03	& 0.36 $\pm$ 0.03 \\
    \bottomrule
  \end{tabular}
\end{table}

\begin{figure*}[!t]
\centering
\begin{subfigure}[t]{0.22\textwidth}
\includegraphics[scale=.14]{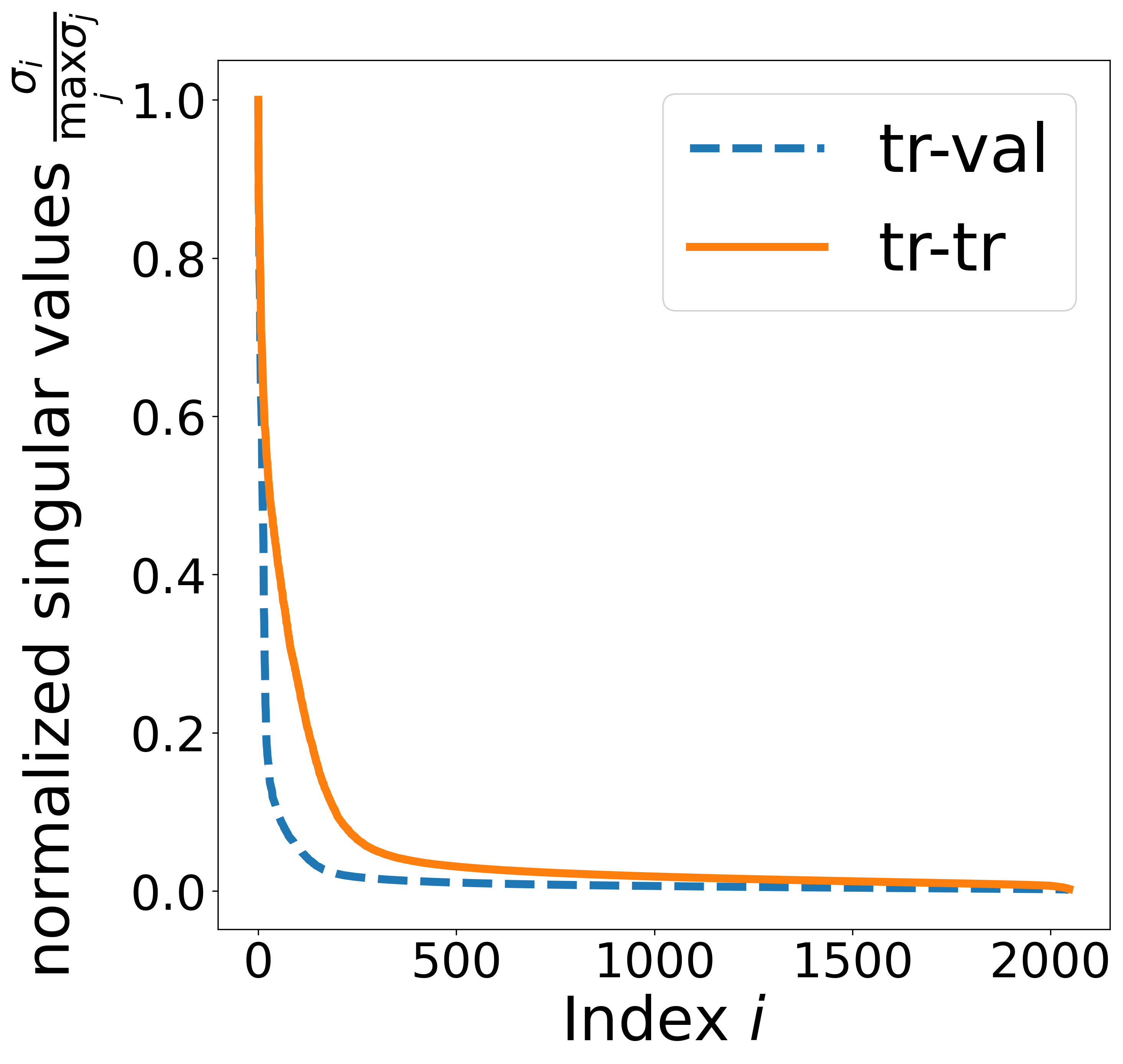}
 \caption{}
 \label{subfig:sing}
 \end{subfigure}%
 \begin{subfigure}[t]{0.22\textwidth}
\includegraphics[scale=.14]{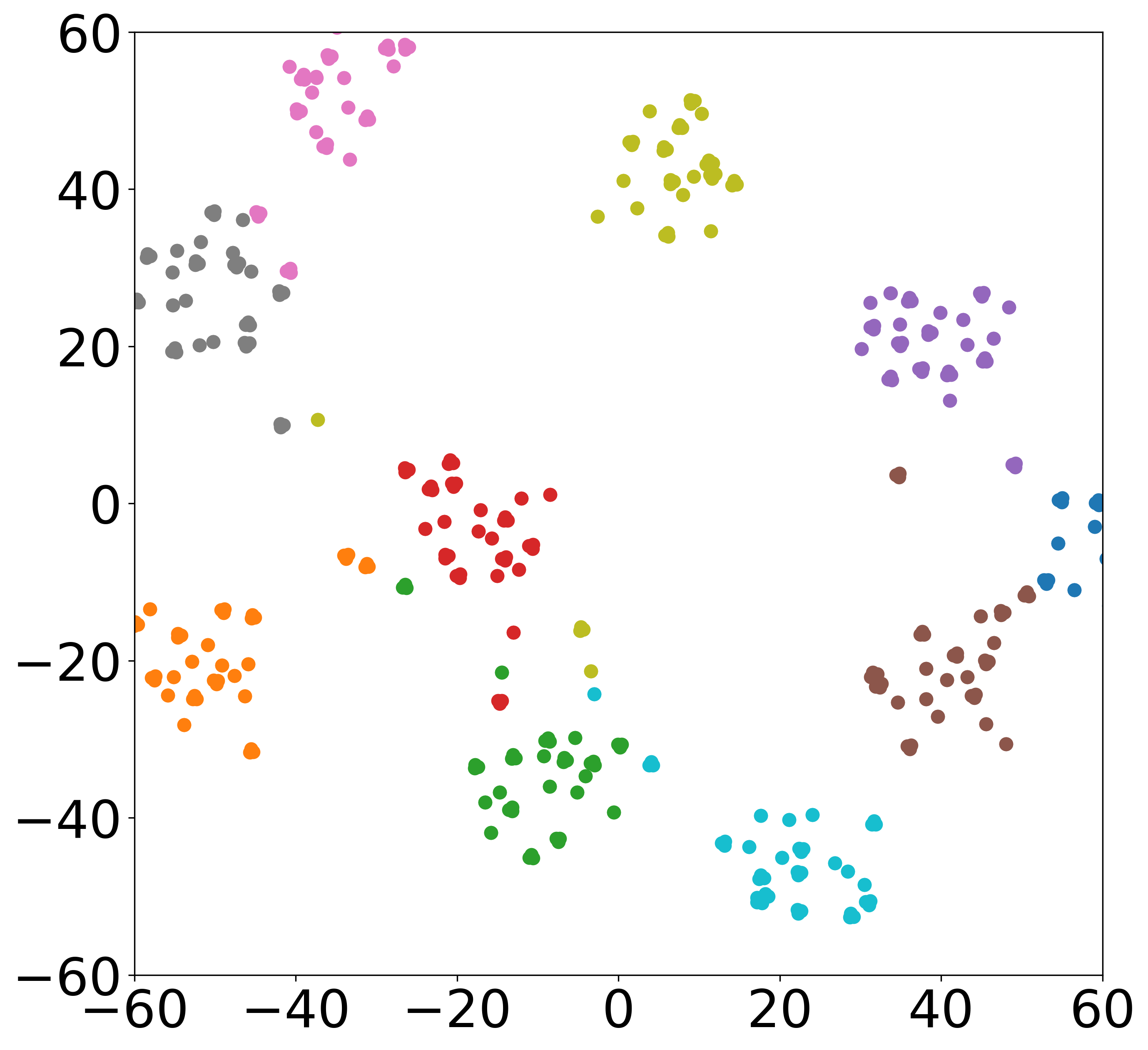}
 \caption{}
 \label{subfig:tsnetrva}
 \end{subfigure}%
  \begin{subfigure}[t]{0.22\textwidth}
\includegraphics[scale=.14]{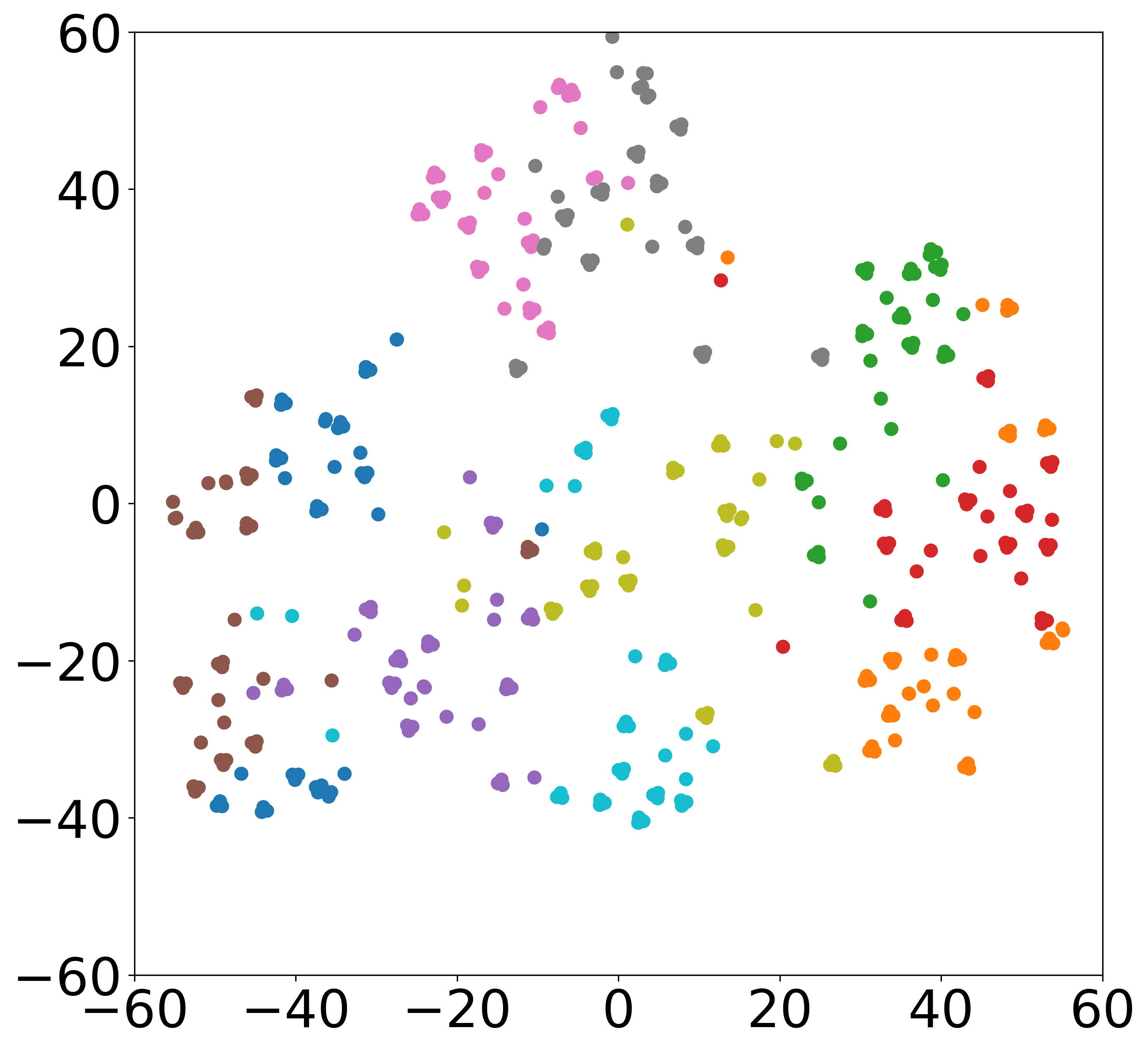}
 \caption{}
 \label{subfig:tsnetrtr}
 \end{subfigure}%
  \begin{subfigure}[t]{0.26\textwidth}
\includegraphics[scale=.14]{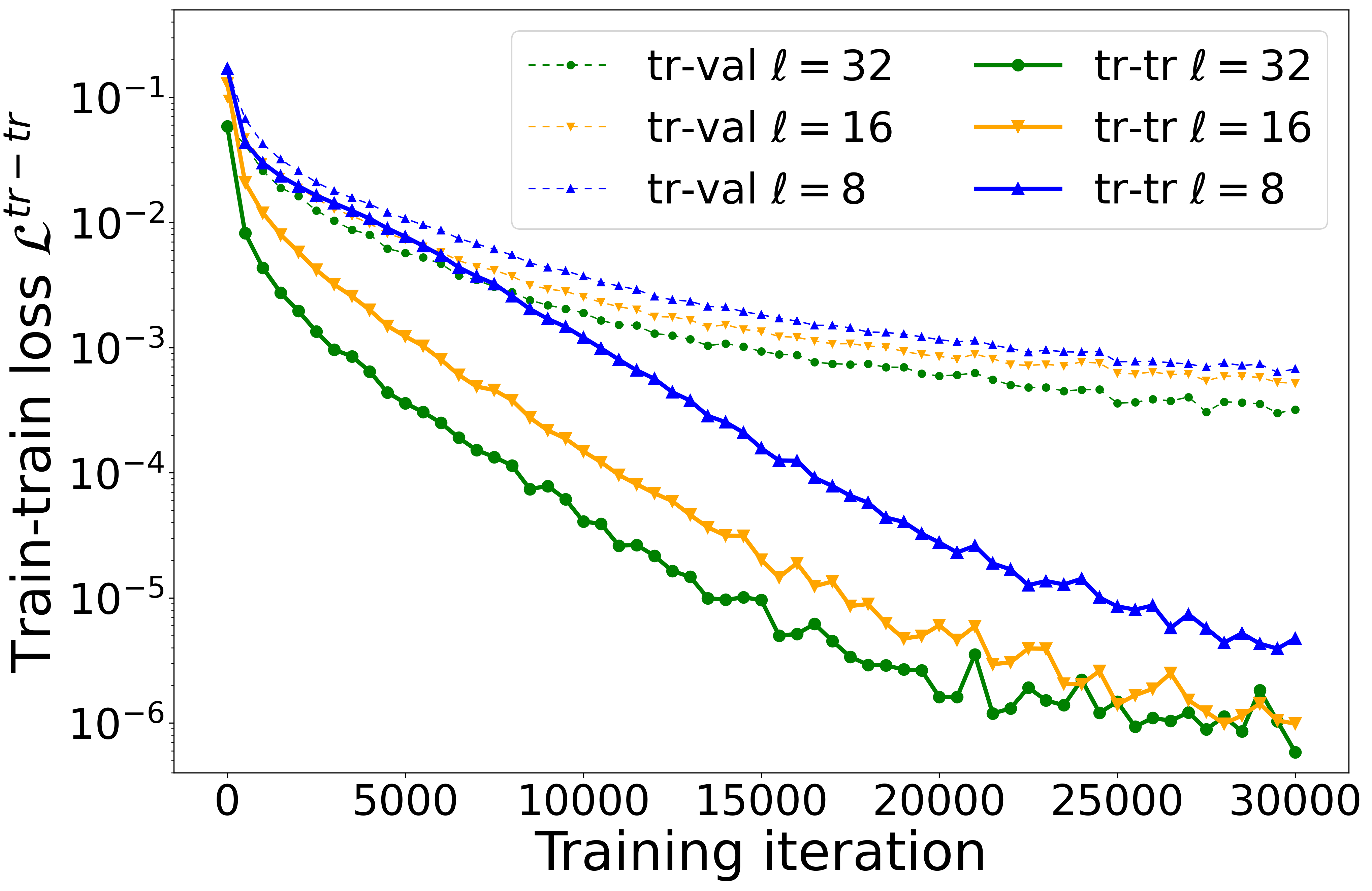}
 \caption{}
 \label{subfig:trtr}
 \end{subfigure}%
\caption{\textbf{(a)}: Normalized singular values for tr-val versus tr-tr. Singular value decay for tr-val is faster, indicating that it has lower effective rank than tr-tr. \textbf{(b,c)}: tSNE for tr-val and tr-tr representations respectively for inputs from 10 randomly selected test classes from Omniglot dataset. Dots with the same color correspond to points from the same class. We find that the tr-val representations are more clustered than the tr-tr representations. \textbf{(d)}: tr-tr loss plotted in log scale versus training iteration for tr-val versus tr-tr runs at varying width factor $\ell$. This verifies that tr-tr models are indeed optimizing the loss that they were meant to minimize.}
\label{fig:lowrank}
\end{figure*}

In this section we present experimental results\footnote{Code available at \url{https://github.com/nsaunshi/meta_tr_val_split}} that demonstrate the practical relevance of our theoretical results.
For simulations we use the subspace learning meta-learning instance defined in \Secref{subsec:meta_instance} and verify that the tr-val objective is indeed superior to the tr-tr variant due to learning low-rank representation.
We also perform experiments on the standard meta-learning benchmarks of Omniglot \citep{lake2015human} and MiniImageNet \citep{vinyals2016matching} datasets, where we again find that representation learning benefits from tr-val splitting.
Furthermore we show that iMAML \cite{rajeswaran2019meta}, a non-representation learning meta-learning algorithm, also benefits from tr-val splitting.
iMAML was also used by \cite{bai2020important} to conclude the tr-tr is better than tr-val.
We also perform a more detailed study of the learned representations on these benchmarks and find that while tr-val and tr-tr representations are almost equally expressive, tr-val representations have lower effective rank and thus have better few-shot performance.
In \Tableref{table:withexpreg} we verify that adding a Frobenius norm regularization on the representation with the tr-tr objective helps significantly, but is still slightly worse than tr-val splitting.

For our representation learning experiments, similar to ANIL algorithm \citep{raghu2019rapid}, we optimize the objective defined in \Eqref{eqn:replearn} that fixes the representation in the inner loop.
Details of our first-order variant of ANIL (akin to first-order MAML \citep{finn2017model}) are deferred to \Secref{asec:replearn}; Algorithm~\ref{alg:outer} summarizes this RepLearn algorithm.
Please refer to \Secref{asec:exps} for additional experiments.

\textbf{Simulation:}
We simulate the meta-learning instance from \Secref{subsec:meta_instance} with $d=50$, $k=5$ and $\sigma=0.5$.
The tr-tr models are trained by minimizing $\Lhattrtrrep(\cdot;n)$ with $n=16$ and tr-val models trained with $\Lhattrvarep(\cdot;(\ntr,\nva))$ with $\ntr=8,\nva=8$. We tune $\telambda$ for before meta-evaluation using validation tasks for each model.
For meta-testing we evaluate with $\nte\in\{5,15,25\}$ train samples per test task in \Tableref{table:simul}.
We find that the tr-val objective, trained with $\lambda=0$ as in \Thmref{thm:trva_result1_robust}, outperforms all tr-tr models, and does almost as well as the correct projection $\Astar\Astar^\top$.
We notice that the tr-tr objective does not do as badly as $\flayer=I_d$, indicating that the training algorithm could have some useful inductive biases, but not enough.
For the $\flayer\in\R^{50\times 50}$ learned using tr-val in the simulation experiment, the top $k=5$ singular values explain 95\% of its total norm, thus implying that it is almost rank $k$.
We also check the alignment of the top-$k$ singular directions of $\flayer$ with the optimal subspace and find the principle angle to be as small as $0.1^{\circ}$. 
For reference, the corresponding numbers for the best tr-tr learned $\flayer$ are $24\%$ and $5^{\circ}$ respectively.

\textbf{RepLearn on Omniglot:}
We investigate the performance of tr-val and tr-tr objectives for representation learning on the Omniglot dataset for the 5-way 1-shot and 20-way 1-shot variants with a 4 layer convolution backbone as in \cite{finn2017model}; results presented in \Tableref{table:omnitable}.
Following the same protocol as \cite{bai2020important}, for meta-training we use $n=2N$ for tr-tr objective and $\ntr=\nva=N$ for the tr-val objective for the $N$-way 1-shot task. For meta-testing we use $\nte = N$. We find that tr-val outperforms tr-tr. Additional details are in \Secref{asec:exps}.

\textbf{Capacity and tr-val versus tr-tr:} 
We examine the performance gap between tr-val versus tr-tr as we increase the expressive power of the representation network.
We use a baseline 4 hidden layer fully connected network (FCN) inspired by \cite{rajeswaran2019meta}, but with $64 \ell$ nodes at all 4 layers for different values of $\ell\in\{1,4,8,16,32\}$.
Using FCNs gets rid of inductive biases of CNNs and helps distill the role of data splitting better.
The models are trained using tr-tr and tr-val representation learning objectives on the Omniglot 5-way 1-shot dataset; results presented in \Tableref{table:widthtable}..
We find that increasing width tends to slightly improve the performance of tr-val models but hurts the performance of tr-tr models, thus increasing the gap between tr-val and tr-tr.
Additionally, we note that the tr-tr model succeeds in minimizing the tr-tr loss $\Ltrtr$ as evident in \Figref{subfig:trtr}.
Thus its bad performance on meta-learning is not an optimization issue, but can be attributed to learning overly expressive representations due to lack of explicit regularization, as predicted by \Thmref{thm:tr_tr_bad_sol}.
This demonstrates that the tr-val objective is more robust to choice of model architecture. 

\begin{table}
\caption{Accuracies in \% of representations parameterized by fully connected networks of varying widths on Omniglot 5-way 1-shot meta-testing and on a supervised dataset that is constructed using the 413 test classes from Omniglot. Representations trained using tr-val objective consistently outperforms those learned using tr-tr objective, and the gap increases as width increases.}
\label{table:widthtable}
    \medskip
  \centering
  \small
\begin{tabular}{ c | c c | c c}
    \toprule
  width $=$	& \multicolumn{2}{c}{Omniglot 5-way 1-shot} & \multicolumn{2}{c}{Supervised 413-way} \\
  $64^*\ell$	& tr-val 			& tr-tr 			& tr-val 	& tr-tr\\
  \midrule
 $\ell = 1$ 		& 87.8 $\pm$ 1.1 	& 80.6 $\pm$ 1.4	& 87.4 	& 73.4\\  
 $\ell = 4$ 		& 90.8 $\pm$ 1.0 	& 77.8 $\pm$ 1.4	& 100.0  	& 100.0\\  
 $\ell = 8$ 		& 91.6 $\pm$ 1.0 	& 73.9 $\pm$ 1.5	& 100.0 	& 100.0\\  
 $\ell = 16$ 	& 91.7 $\pm$ 0.9 	& 70.6 $\pm$ 1.6	& 100.0  	& 100.0\\  
 $\ell = 32$ 	& 91.6 $\pm$ 1.0 	& 67.8 $\pm$ 1.6	& 100.0	& 100.0\\
 \bottomrule
\end{tabular}
\end{table}

\textbf{Low-rankness and expressivity:}
We investigate the representations from the trained FCNs with $\ell = 32$.
To ascertain expressivity of representations, we combine all data from the test classes of Omniglot to produce a supervised learning dataset with 413 classes.
We measure expressivity of representations by evaluating its linear classification performance on the 413-way supervised task; results in \Tableref{table:widthtable}.
We find that the tr-val representations do at least as well as tr-tr ones and thus are expressive enough.
To compare the effective ranks, we plot the normalized singular values of both sets of representations in \Figref{subfig:sing} and find that the tr-val representations have a steeper drop in singular values than the tr-tr representations, thus confirming their lower effective rank.
t-SNE plots (\Figref{subfig:tsnetrva} and \Figref{subfig:tsnetrtr}) on the tr-val and tr-tr representations for data points from 10 randomly selected classes suggest that tr-val representations are better clustered than the tr-tr representations.

\textbf{iMAML on Omniglot:}
\label{subsec:imaml_exp}
We go beyond representation learning and consider a meta-learning method iMAML \citep{rajeswaran2019meta} that updates all parameters in the inner loop. 
We modify the authors code\footnote{\url{https://github.com/aravindr93/imaml_dev}} to implement a tr-tr version and we investigate the performance of tr-val and tr-tr objectives on the Omniglot dataset for both 5-way 1-shot and 20-way 1-shot settings in Table \ref{table:omnitable}. 
We find that tr-val outperforms tr-tr again, this time by a significant margin.
We do not tune hyper-parameters for tr-val and just use the defaults, but for tr-tr we tune them to give it an edge.

\textbf{RepLearn on MiniImageNet:}
We performs similar experiments on MiniImageNet using a standard CNN backbone with 32, 64, 128, and 128 filters, respectively.
\Tableref{table:miniImageNet} shows again that RepLearn with tr-val splitting is much better than tr-tr on 5-way 1-shot and 5-way 5-shot settings, similar to the findings with Omniglot.
We also perform the capacity experiment by increasing the number of filters by a factor $\ell$; the convolutional layers contain $32 \ell$, $64 \ell$, $128 \ell$, and $128 \ell$ output filters, respectively.  Results in \Tableref{table:miniwidth} suggest that increasing the network capacity improves the performance of tr-val representations, but slightly hurts tr-tr performance, just like the findings for Omniglot dataset with fully-connected networks.
The drop might be lower here due to a CNN being used instead of fully connected network in \Tableref{table:widthtable}.
Thus the tr-val method is more robust to architecture choice/capacity and datasets.



\begin{table}[!t]
\caption{Accuracies in \% of representations parameterized by CNN networks with varying number of filters on MiniImageNet. Representations trained using tr-val objective consistently outperforms those learned using tr-tr objective; gap increases as width increases.}
\label{table:miniwidth}
    \medskip 
  \centering
  \small
\begin{tabular}{ c | c c }
    \toprule
  capacity $=$	& \multicolumn{2}{c}{MiniImageNet 5-way 1-shot} \\
  $num\_filters^*\ell$	& tr-val 			& tr-tr 			\\
  \midrule
 $\ell = 0.5$ 	& 46.66  $\pm$ 1.69	  & 26.25 $\pm$ 1.45		\\  
 $\ell = 1$ 		& 48.44 $\pm$ 1.62  	  & 26.81 $\pm$ 1.44		\\  
 $\ell = 4$ 		& 52.22 $\pm$ 1.68.	  & 24.66 $\pm$ 1.26		\\  
 $\ell = 8$ 		& 52.25 $\pm$ 1.71	  &25.28 $\pm$ 1.37		\\  
 \bottomrule
\end{tabular}
\end{table}

\begin{center}
\begin{table}[!t]
\caption{Tr-val v/s tr-tr meta-test accuracies in \% for a CNN model trained with RepLearn on MiniImageNet 5-way 1-shot  and 5-way 5-shot for varying values of the regularization parameter, $\lambda$. 
}
 \medskip
  \centering
  \small
\begin{tabular}{ c | c | c }
    \toprule
  & 5-way 1-shot&  5-way 5-shot \\
      \midrule
$\lambda=$ 0.0 tr-val 	& 46.16 $\pm$ 1.67		& 65.36 $\pm$ 0.91 \\
\midrule
$\lambda=$ 0.0 tr-tr  		& 25.53 $\pm$ 1.43     	& 33.49 $\pm$ 0.82  \\
$\lambda=$ 0.1 tr-tr  		& 24.69 $\pm$ 1.32       	& 34.91 $\pm$ 0.85\\
$\lambda=$ 1.0 tr-tr  		& 25.88 $\pm$ 1.45    	& 40.19 $\pm$ 1.12 \\
\bottomrule
\end{tabular}
\label{table:miniImageNet}
\end{table}
\end{center}

\section{Discussions and Future Work}
\label{sec:conclusion}

We study the implicit regularization effect of the practical design choice of train-validation splitting popular in meta-learning, and show that it encourages learning low-rank but expressive enough representations.
This is contrasted with the non-splitting variant that is shown to fail without explicit regularization.
Both of these claims are justified theoretically for linear representation learning and experimentally for standard meta-learning benchmarks.
Train-validation splitting provides a new mechanism for sample efficiency through implicit regularization in the objective, as opposed to explicit regularization and implicit bias of training algorithm, as discussed in \Secref{subsec:comparison}.
We show learning of exact low rank representations in our setting as opposed to approximate low-rankness observed in practice.
Relaxing assumptions of Gaussianity, common input distribution across tasks and linearity of representations might explain the observed {\em effective} low-rankness.
Finally an interesting problem is to get the best of all worlds, data efficiency from tr-tr style objective, explicit regularization and the implicit low rank regularization from tr-val splitting in a principled way.
Identifying and understanding other training paradigms that intrinsically use data efficiently, even without explicit regularization is also an interesting direction.

\textbf{Acknowlegments:}
We thank Mikhail Khodak for comments on an earlier draft of the paper. Nikunj Saunshi, Arushi Gupta and Wei Hu are supported by NSF, ONR, Simons Foundation, Amazon Research, DARPA and SRC.

\clearpage
\bibliography{references}
\bibliographystyle{plainnat}

\appendix
\onecolumn

\section{Appendix Overview}
\label{asec:overview}

The Appendix is structured as follows
\begin{itemize}
	\item \Secref{asec:trtr} proves the result for linear representation learning with the tr-tr objective, \Thmref{thm:tr_tr_bad_sol}, that shows that ``bad'' full rank solutions exist arbitrarily close to the optimal value of the tr-tr objective $\Ltrtrrep$.
	The proof works by arguing that smaller values of $\lambda$ are desirable for the tr-tr objective, and this can be simulated by effectively making the norm of the representation layer very high.
	Furthermore a higher rank representation is preferable over lower rank ones to fit the noise in the labels in the training data better.
	
	\item \Secref{asec:trva} proves the main result for linear representation learning with tr-val objective, \Thmref{thm:trva_result1_robust}, that proves that the optimal solutions to the tr-val objective $\Ltrvarep(\flayer;(\ntr,\nva))$ for most $\ntr$ and $\sigma$ will be {\em low-rank} representations that are also {\em expressive enough}.
	The result is also extended to solutions that are $\tau$-optimal in the $\Ltrvarep$ objective for a small enough $\tau$.
	The crux of the proof for this result is in \Thmref{thm:trva_closedform} that provides a closed form expression for the tr-val objective that disentangles the expressivity and the low-rankness of the representation.
	
	\item \Secref{asec:exps} presents additional experimental details and results, including results for the MiniImageNet dataset.
\end{itemize}



\section{More on Train-Train split}
\label{asec:trtr}

\subsection{Proof of main result}
\begin{reptheorem}
{thm:tr_tr_bad_sol}
For every $\lambda,n>0$,  for every $\tau>0$, there exists a ``bad'' representation layer $\flayer_{\bad}\in\R^{d\times d}$ that satisfies $\Ltrtrrep(\flayer_{\bad};n)\le\inf\limits_{\flayer\in\R^{d\times d}} \Ltrtrrep(\flayer;n) + \tau$, but has the following lower bound on meta-testing loss
	\begin{align*}
		\inf_{\telambda>0} &\Ltestrep(\flayer_\bad;\nte) \ge \sigma^2 + \min\left\{1 - \frac{\nte}{d(1+\sigma^2)}, \frac{d\sigma^2}{(\nte+d\sigma^2)}\right\}
	\end{align*}
\end{reptheorem}
\begin{proof}
	For most of the proof, we will leave out the $n$ in the expression for $\Ltrtrrep$, i.e. we will denote the tr-tr loss as $\Ltrtrrep(\flayer)$.
	We prove this result by using \Lemref{lem:tr_tr_full_rank} first, and later prove this lemma.
	\begin{replemma}
	{lem:tr_tr_full_rank}
		For every $\lambda>0$ and $\flayer\in\R^{d\times d}$ with rank $r$,
		\begin{align*}
			\Ltrtrrep(\flayer;n) \ge \lim_{\kappa\rightarrow\infty}\Ltrtrrep(\kappa&\flayer;n) \ge \sigma^2\frac{(n-r)_+}{n}\\
			\text{  \&  }~\lim_{\kappa\rightarrow\infty}\Ltrtrrep(\kappa \mI_d;n) &= \sigma^2\frac{(n-d)_+}{n}
		\end{align*}
	\end{replemma}
	This tells us that for any matrix $\flayer$, $\Ltrtrrep(\flayer) \ge \sigma^2\frac{(n-\rank(\flayer))_+}{n} \ge \sigma^2\frac{(n-d)_+}{n}$.
	Also the lower bound of $\sigma^2\frac{(n-d)_+}{n}$ can be achieved by $\kappa \mI_d$ in the limit of $\kappa\rightarrow\infty$, thus $\sigma^2\frac{(n-d)_+}{n} = \inf_{\flayer\in\R^{d\times d}}\Ltrtrrep(\flayer)$.
	Also since $\lim_{\kappa\rightarrow\infty}\kappa \mI_d = \sigma^2\frac{(n-d)_+}{n} = \inf_{\flayer\in\R^{d\times d}}\Ltrtrrep(\flayer)$, for a large enough $\kappa = \kappa(\tau)$, $\Ltrtrrep(\kappa(\tau)\mI_d)$ can be made lesser than $\inf_{\flayer\in\R^{d\times d}}\Ltrtrrep(\flayer) + \tau$.
	Thus we pick $\flayer_{\bad} = \kappa(\tau)\mI_d$.
	With this choice, the new task is essentially linear regression in $d$-dimension with isotropic Gaussians.
	
	To show that $\flayer_{\bad}$ is indeed bad, we will use the lower bound for ridge regression on isotropic Gaussian linear regression from Theorem 4.2(a) in \citet{saunshi2020sample}.
	They show that the excess risk for $\mI_d$ (and thus $\kappa(\tau)\mI_d$), regardless of the choice of regularizer $\telambda$, for a new task $\dist_{\tclf}$ will be lower bounded by 
	\begin{align*}
		\inf_{\telambda>0}\ex_{\Sto\sim\dist_{\tclf}^{\nte}} \left[\|\flayer\slayer_{\telambda}(\flayer;\Sto) - \tclf\|^2\right]
		& \ge \begin{cases}
			\frac{d\|\tclf\|^2\sigma^2}{\nte\|\tclf\| + \sigma^2d} & \text{if } \nte \ge d\\
			&\\
			\frac{\nte}{d} \frac{\|\tclf\|^2\sigma^2}{\|\tclf\| + \sigma^2} + \frac{d - \nte}{d}\|\tclf\|^2 & \text{if } \nte < d\\
		\end{cases}
	\end{align*}
	Their proof can be easily modified to replace $\|\tclf\|^2$ with $\ex_{\tclf\sim\tedist}\|\tclf\|^2$.
	The lower bound can be simplified for the the $\nte < d$ case to $\frac{\nte}{d} \frac{\|\tclf\|^2\sigma^2}{\|\tclf\| + \sigma^2} + \frac{d - \nte}{d}\|\tclf\|^2 = \|\tclf\|^2 - \frac{\nte}{d}\frac{\|\tclf\|^2}{\|\tclf\|^2 + \sigma^2}$.
	Plugging in $\|\tclf\|=1$ completes the proof.
\end{proof}

	We now prove the lemma
	\begin{replemma}
	{lem:tr_tr_full_rank}
		For every $\lambda>0$ and $\flayer\in\R^{d\times d}$ with rank $r$,
		\begin{align*}
			\Ltrtrrep(\flayer;n) \ge \lim_{\kappa\rightarrow\infty}\Ltrtrrep(\kappa&\flayer;n) \ge \sigma^2\frac{(n-r)_+}{n}\\
			\text{  \&  }~\lim_{\kappa\rightarrow\infty}\Ltrtrrep(\kappa \mI_d;n) &= \sigma^2\frac{(n-d)_+}{n}
		\end{align*}
	\end{replemma}
	\begin{proof}
		We first prove that having $\lambda=0$ will lead to the smallest loss $\Ltrtrrep(\flayer)$ for every $\flayer$. We then observe that $\lambda=0$ can be simulated by increasing the norm of $\flayer$.
	These claims mathematically mean that, (a) $\Ltrtr_{\lambda,\rep}(\flayer) \ge \Ltrtr_{\lambda',\rep}(\flayer)$ whenever $\lambda\ge\lambda')$ and (b) $\Ltrtr_{\lambda,\rep}(\kappa\flayer) = \Ltrtr_{\frac{\lambda}{\kappa^2},\rep}(\flayer)$.
	This will give us that $\lim_{\kappa\rightarrow\infty} \Ltrtr_{\lambda,\rep}(\kappa\flayer) = \lim_{\lambda\rightarrow 0}\Ltrtr_{0,\rep}(\kappa\flayer) \ge \Ltrtr_{\lambda,\rep}(\flayer)$.
	Intuitively, $\Ltrtrrep$ is trying to learn a linear classifier on top of data that is linear transformed by $\flayer$ with the goal of fitting the same data well.
	
	\begin{lemma}
	\label{lem:trtr_limit}
		For any representation layer $\flayer\in\R^{d\times d}$ and $\lambda>0$, we have the following
		\begin{align*}
			\lim_{\kappa\rightarrow\infty}\Ltrtrrep(\kappa\flayer) = \lim_{\lambda\rightarrow 0}\Ltrtrrep(\flayer) \le \Ltrtrrep(\flayer)
		\end{align*}
	\end{lemma}
	Fitting the data is better when there is less restriction on the norm of the classifier, which in this case means when $\lambda$ is smaller. Furthermore, increasing the norm of the representation layer $\flayer$ effectively reduces the impact the regularizer will have.
	We first prove this lemma later, first we use it to prove \Lemref{lem:tr_tr_full_rank} that shows that the loss for low rank matrices will be high.

		\Lemref{lem:trtr_limit} already shows that $\Ltrtrrep(\flayer;n) \ge \lim_{\kappa\rightarrow\infty}\Ltrtrrep(\kappa\flayer;n)$.
		Also using \Lemref{lem:trtr_limit}, we can replace $\lim_{\kappa\rightarrow\infty}\Ltrtrrep(\kappa\flayer)$ with $\lim_{\lambda\rightarrow 0}\Ltrtrrep(\flayer)$.
		Using \Eqref{eqn:replearn_trva} and from central limit theorem, we have
		\begin{align}
			\Ltrtrrep(\flayer) = \lim_{T\rightarrow\infty}\Lhattrtr(\flayer)
			&= \ex_{\dist_{\tclf}\sim\tdist}\left[\ex_{\Sto\sim\dist_{\tclf}^{n}}\left[\frac{1}{n}\|\Xto\flayer\slayer_{\lambda}(\flayer; \Sto) - \Yto\|^2\right]\right]\\
			&= \ex_{\tclf\sim\gN(\vzero_d, \Astar\Astar^\top)}\left[\ex_{\Sto\sim\dist_{\tclf}^{n}}\left[\frac{1}{n}\|\Xto\flayer\slayer_{\lambda}(\flayer; \Sto) - \Yto\|^2\right]\right]
			\label{eqn:Ltrtr_exp}
		\end{align}
		This is because $\Lhattrtrrep$ is an average loss for $T$ train tasks, and the limit when $T\rightarrow\infty$ it converges to the expectation over the task distribution $\tdist$.
		We first observe that $\Sto\sim\dist_{\tclf}$ is equivalent to sampling $\Xto\sim\gN(\vzero_d,\mI_d)$, $\veta\sim\gN(\vzero_n,\sigma^2 I_n)$ which gives us $\Yto = \Xto\tclf + \veta$, where $\Xto\in\R^{n\times d},\veta\in\R^n,\Yto\in\R^n$.
		Using the definition of $\slayer_{\lambda}(\flayer; \Sto)$ from \Eqref{eqn:replearn_inner_alg} the standard KKT condition for linear regression, we can write a closed form solution for 
		\begin{align}
			\slayer_{\lambda}(\flayer; \Sto)
			&= \argmin_{\slayer\in\R^d} \frac{1}{n} \|\Xto\flayer\slayer - \Yto\|^2 + \lambda \|\slayer\|^2 = \left(\frac{\flayer^\top\Xto^\top\Xto\flayer}{n} + \lambda \mI_d\right)^{-1} \frac{\flayer^\top\Xto^\top}{n}\Yto
			\label{eqn:ridge_soln}\\
			\lim_{\lambda\rightarrow0} \slayer_{\lambda}(\flayer; \Sto)
			&= \left(\frac{\flayer^\top\Xto^\top\Xto\flayer}{n}\right)^{\dagger} \frac{\flayer^\top\Xto^\top}{n}\Yto
		\end{align}
		where the last step is folklore that the limit of ridge regression as regularization coefficient goes to 0 is the minimum $\ell_2$-norm linear regression solution.
		Plugging this into \Eqref{eqn:Ltrtr_exp} and taking the limit 
		\begin{align*}
			\lim_{\lambda\rightarrow 0} \Ltrtrrep(\flayer)
			&= \ex_{\tclf,\Sto}\left[\frac{1}{n}\left\|\Xto\flayer\left(\lim_{\lambda\rightarrow 0}\slayer_{\lambda}(\flayer; \Sto)\right) - \Yto\right\|^2\right]\\
			&= \ex_{\tclf,\Sto}\left[\frac{1}{n}\left\|\left(\Xto\flayer \left(\frac{\flayer^\top\Xto^\top\Xto\flayer}{n}\right)^{\dagger} \frac{\flayer^\top\Xto^\top}{n} - I_n\right)\Yto\right\|^2\right]\\
			&= \frac{1}{n}\ex_{\tclf,\Sto}\left[\left\|P^{\perp}_{\Xto\flayer}\Yto\right\|^2\right]
		\end{align*}
		where for any matrix $\mB\in\R^{n\times d}$, we denote $P_{\mB}\in\R^{n\times n} = \mB\mB^{\dagger}$ to denote the projection matrix onto the span of columns of $\mB$, and $P^{\perp}_{\mB} = I_n - P_{\mB}$ is the projection matrix onto the orthogona subspace.
		Note that if $\rank(\mB) = n$, then $P_{\mB} = I_n$ and $P^{\perp}_{\mB} = 0$.
		Thus the error incurred is the amount of the label $\mY$ that the representation $\Xto\flayer$ cannot predict linearly.
		We further decompose this
		\begin{align}
			\lim_{\lambda\rightarrow 0} \Ltrtrrep(\flayer)
			&= \frac{1}{n}\ex_{\tclf,\Sto}\left[\left\|P^{\perp}_{\Xto\flayer}\Yto\right\|^2\right]
			= \frac{1}{n}\ex_{\tclf,\Sto}\left[\left\|P^{\perp}_{\Xto\flayer}\Xto\tclf + P^{\perp}_{\Xto\flayer}\veta\right\|^2\right]\nonumber\\
			&= \underbrace{\frac{1}{n}\ex_{\tclf,\Sto}\left[\left\|P^{\perp}_{\Xto\flayer}\Xto\tclf\right\|^2\right]}_{\text{fitting signal}} + \underbrace{\frac{1}{n}\ex_{\tclf,\Sto}\left[\left\|P^{\perp}_{\Xto\flayer}\veta\right\|^2\right]}_{\text{fitting noise}} + \underbrace{\frac{2}{n}\ex_{\tclf,\Sto}\left[\veta^{\top}P^{\perp}_{\Xto\flayer}\Xto\tclf\right]}_{\text{cross term}}
			\label{eqn:trtr_decomposition}\\
			&= \frac{1}{n}\ex_{\Sto}\ex_{\tclf\sim\gN(\vzero_d,\Astar\Astar^{\top})}\left[\left\|P^{\perp}_{\Xto\flayer}\Xto\tclf\right\|^2\right] + \ex_{\Sto}\frac{1}{n}\tr\left(P^{\perp}_{\Xto\flayer}\ex_{\veta}\left[\veta\veta^\top\right]P^{\perp}_{\Xto\flayer}\right)\nonumber\\
			&= \underbrace{\frac{1}{n}\ex_{\Sto}\left[\left\|P^{\perp}_{\Xto\flayer}\Xto\Astar\right\|^2\right]}_{\alpha(\flayer)} + \underbrace{\ex_{\Sto}\frac{\sigma^2}{n}\tr\left(P^{\perp}_{\Xto\flayer}\right)}_{\beta(\flayer)}
		\end{align}
		We first note, due to independence of $\veta$ and $\Xto$ that the cross term in \Eqref{eqn:trtr_decomposition} is 0 in expectation.
		Using the distributions for $\tclf\sim\gN(\vzero_d,\Astar\Astar^\top)$ and $\veta\sim\gN(\vzero_n,\sigma^2 I_n)$, we get the final expressions
		Firstly, we note that $\alpha(\flayer) \ge 0$ for every $\flayer$, and a sufficient condition for $\alpha(\flayer) = 0$ is that $\Astar$ lies in the span of $\flayer$, i.e. $P^{\perp}_{\flayer}\Astar = 0$.
		More importantly, it is clear that $\alpha(\mI_d) = 0$.
		Next we look at the $\beta(\flayer)$ term which is proportional to the trace of $P^{\perp}_{\Xto\flayer}$ which, for a projection matrix, is also equal to the rank of the matrix.
		Note that since the rank of $\Xto\flayer\in\R^{d\times n}$ is at most $\min\{n,d\}$.
		Thus we get
		\begin{align*}
			\beta(\flayer) = \sigma^2 \frac{\rank(P^{\perp}_{\Xto\flayer})}{n} = \sigma^2 \frac{(n - \rank(P^{\perp}_{\Xto\flayer}))}{n} \ge \sigma^2 \frac{(n - \rank(\flayer))_+}{n}
		\end{align*}
		where $(x)_+ = \mathds{1}_{x\ge0} x$.
		This along with $\alpha(\flayer)\ge 0$ proves the first part of the result.
		For the second part $\alpha(\mI_d)=0$ along with noticing that $\tr(P^{\perp}_{\Xto \mI_d}) = n - \rank(\Xto) = n-\min\{n,d\}$ since a Gaussian matrix is full rank with measure 1, thus giving us $\beta(\mI_d) = \sigma^2 \frac{(n-d)_+}{n}$ and completing the proof.
	\end{proof}
	
	Thus \Lemref{lem:tr_tr_full_rank} shows that $\lim_{\kappa\rightarrow\infty}\Ltrtrrep(\kappa \mI_d) = \inf_{\flayer} \Ltrtrrep(\flayer)$ and so picking a large enough $\kappa(\tau)$ can always give us $\Ltrtrrep(\kappa(\tau) \mI_d) \le \inf_{\flayer} \Ltrtrrep(\flayer) + \tau$ which completes the proof of \Thmref{thm:tr_tr_bad_sol}.
	The only thing left to prove is  \Lemref{lem:trtr_limit} which we do below
	
	\begin{proof}[Proof of \Lemref{lem:trtr_limit}]
		Using \Eqref{eqn:ridge_soln} we get the following expression for $\Ltrtrrep$
		\begin{align}
			\Ltrtrrep(\flayer)
			&= \ex_{\tclf,\Sto}\left[\frac{1}{n}\left\|\Xto\flayer\slayer_{\lambda}(\flayer; \Sto) - \Yto\right\|^2\right]\nonumber\\
			&= \ex_{\tclf,\Sto}\left[\frac{1}{n}\left\|\left(\Xto\flayer \left(\frac{\flayer^\top\Xto^\top\Xto\flayer}{n} + \lambda \mI_d\right)^{-1} \frac{\flayer^\top\Xto^\top}{n} - I_n\right)\Yto\right\|^2\right]\label{eqn:Ltrtr_exp}
		\end{align}
		From this we get
		\begin{align*}
			\Ltrtrrep(\kappa\flayer)
			&= \ex_{\tclf,\Sto}\left[\frac{1}{n}\left\|\left(\kappa\Xto\flayer \left(\frac{\kappa^2\flayer^\top\Xto^\top\Xto\flayer}{n} + \lambda \mI_d\right)^{-1} \frac{\kappa\flayer^\top\Xto^\top}{n} - I_n\right)\Yto\right\|^2\right]\\
			&= \ex_{\tclf,\Sto}\left[\frac{1}{n}\left\|\left(\Xto\flayer \left(\frac{\flayer^\top\Xto^\top\Xto\flayer}{n} + \frac{\lambda}{\kappa^2} \mI_d\right)^{-1} \frac{\flayer^\top\Xto^\top}{n} - I_n\right)\Yto\right\|^2\right]\\
			&= \Ltrtr_{\frac{\lambda}{\kappa^2},\rep}(\flayer)
		\end{align*}
		Thus $\lim\limits_{\kappa\rightarrow\infty}\Ltrtrrep(\kappa\flayer) = \lim\limits_{\kappa\rightarrow\infty}\Ltrtr_{\frac{\lambda}{\kappa^2},\rep}(\flayer) = \lim\limits_{\lambda\rightarrow0} \Ltrtr_{\lambda,\rep}(\flayer)$.
		Note that we have used the fact that $\Ltrtrrep$ is continuous in $\lambda$ for $\lambda>0$.
		To prove the remaining result, i.e. $\lim\limits_{\lambda\rightarrow0} \Ltrtr_{\lambda,\rep}(\flayer) \le \Ltrtr_{\lambda,\rep}(\flayer)$, we just need to prove that $\Ltrtrrep(\flayer)$ is a increasing function of $\lambda$.
		Suppose $\frac{\Xto\flayer}{\sqrt{n}} = \mU_{\Xto}\mS_{\Xto}\mV_{\Xto}^\top$ is the singular value decomposition.
		For any $\lambda'<\lambda$, we can rewrite $\Ltrtrrep(\flayer)$ from \Eqref{eqn:Ltrtr_exp} as follows
		\begin{align*}
			\Ltrtrrep(\flayer)
			&= \ex_{\tclf,\Sto}\left[\frac{1}{n}\left\|\left(\Xto\flayer \left(\frac{\flayer^\top\Xto^\top\Xto\flayer}{n} + \lambda \mI_d\right)^{-1} \frac{\flayer^\top\Xto^\top}{n} - I_n\right)\Yto\right\|^2\right]\\
			&= \ex_{\tclf,\Sto}\left[\frac{1}{n}\left\|\left(\mV_{\Xto}\mS_{\Xto}\mU_{\Xto}^\top \left(\mU_{\Xto}\mS_{\Xto}\mV_{\Xto}^\top\mV_{\Xto}\mS_{\Xto}\mU_{\Xto}^\top + \lambda \mI_d\right)^{-1} \mU_{\Xto}\mS_{\Xto}\mV_{\Xto}^\top - I_n\right)\Yto\right\|^2\right]\\
			&= \ex_{\tclf,\Sto}\left[\frac{1}{n}\left\|\mV_{\Xto}\left(\mS_{\Xto}\left(\mS^2_{\Xto} + \lambda \mI_d\right)^{-1} \mS_{\Xto} - I_n\right)\mV_{\Xto}^\top\Yto\right\|^2\right]\\
			&= \ex_{\tclf,\Sto}\left[\frac{1}{n}\sum_{i=1}^{d}\left[\left(\frac{\mS_{\Xto}[i]^2}{\lambda + \mS_{\Xto}[i]^2} - 1\right)^2\mV_{\Xto}^\top\Yto[i]^2\right]\right]\\
			&= \ex_{\tclf,\Sto}\left[\frac{1}{n}\sum_{i=1}^{d}\left[\left(\frac{\lambda}{\lambda + \mS_{\Xto}[i]^2}\right)^2\mV_{\Xto}^\top\Yto[i]^2\right]\right]\\
			&\ge \ex_{\tclf,\Sto}\left[\frac{1}{n}\sum_{i=1}^{d}\left[\left(\frac{\lambda'}{\lambda' + \mS_{\Xto}[i]^2}\right)^2\mV_{\Xto}^\top\Yto[i]^2\right]\right]\\
			&= \Ltrtr_{\lambda',\rep}(\flayer)
		\end{align*}
		where the only inequality in the above sequence follows from the observation that $\frac{\lambda}{\lambda+a}$ is an increasing function for $a,\lambda>0$.
		This completes the proof.
	\end{proof}


\section{More on Train-Validation split}
\label{asec:trva}

\subsection{Proof of main results}
We first prove the result that for $\ntr=\nte$ and $\lambda=\telambda$, $\Ltrvarep \equiv \Ltestrep$.
\begin{repprop}
{prop:Ltrva_equal_Ltest}
	$\Ltrvarep(\cdot;(\ntr,\nva))$ and $\Ltestrep(\cdot;\nte)$ are equivalent if $~\nte = \ntr$ and $\telambda=\lambda$
\end{repprop}
\begin{proof}
	We again note using the central limit theorem and \Eqref{eqn:replearn_trva} that
	\begin{align*}
		\Ltrvarep(\flayer) = \lim\limits_{T\rightarrow\infty} \Lhattrvarep(\flayer) 
		&= \ex_{\dist_{\tclf}\sim\tedist} \left[\ex_{(\Str,\Sva)\sim\dist_{\tclf}^{n}} \left[\frac{1}{\nva}\left\|\Xva\flayer\slayer_{\lambda}(\flayer; \Str) - \Yva\right\|^2\right]\right]\\
		&=^{(a)} \ex_{\dist_{\tclf}\sim\tedist} \left[\ex_{\Str\sim\dist_{\tclf}^{\ntr}} \left[\ex_{\Sva\sim\dist_{\tclf}^{\nva}}\frac{1}{\nva}\left\|\Xva\flayer\slayer_{\lambda}(\flayer; \Str) - \Yva\right\|^2\right]\right]\\
		&= \ex_{\dist_{\tclf}\sim\tedist} \left[\ex_{\Str\sim\dist_{\tclf}^{\ntr}} \left[\ex_{(\vx,y)\sim\dist_{\tclf}}\left(\vx^\top\flayer\slayer_{\lambda}(\flayer; \Str) - y\right)^2\right]\right]\\
		&= \ex_{\dist_{\tclf}\sim\tedist} \left[\ex_{\Str\sim\dist_{\tclf}^{\ntr}} \left[\left\|\flayer\slayer_{\lambda}(\flayer; \Str) - \tclf\right\|^2\right]\right] + \sigma^2\\
		&=^{(b)} \Ltest_{\lambda,\rep}(\flayer;\ntr)
	\end{align*}
	where $(a)$ follows by noticing that sample $\Sto\sim\dist_{\tclf}^{n}$ and splitting randomly into $\Str$ and $\Sva$ is equivalent to independently sampling $\Str\sim\dist_{\tclf}^{\ntr}$ and $\Sva\sim\dist_{\tclf}^{\nva}$ and $(b)$ follows from the definition of $\Ltest$ from \Eqref{eqn:meta_test}.


\end{proof}

We now prove the main benefit of the tr-val split: learning of low rank linear representations.
For that, we use this general result that computes closed form expression for $\Ltrvarep$.
\begin{theorem}
\label{thm:trva_closedform}
	Let $\lambda=0$. For a first layer $\flayer\in\R^{d\times d}$ with $r = \rank(\flayer)$, let $\flayer = \mU\mS\mV^\top$ be the SVD, where $\mU,\mV\in\R^{d\times r}$ and $\mS\in\R^{r\times r}$.
	Furthermore let $P_{\mU}=\mU\mU^\top$ denote the span of columns of $\mU$ (and thus $\flayer$) and let $P^{\perp}_{\mU} = I_d - P_{\mU}$. Then the tr-val objective has the following form
	\begin{align*}
                	\Ltrvarep(\flayer;(\ntr,\nva)) - \sigma^2 = 
		\begin{cases}
			(1+\alpha(\ntr,r))\|P_{\mU}^{\perp}\Astar\|^2 + \alpha(\ntr,r)\sigma^2 & \text{if } r < \ntr-1\\
			&\\
			(\frac{\ntr}{r}+\alpha(r,\ntr))\|P_{\mU}^{\perp}\Astar\|^2 + \frac{r-\ntr}{r} + \beta(\mS) + \alpha(r,\ntr)\sigma^2 & \text{if } r > \ntr+1\\
			&\\
			\infty & \text{otherwise}
		\end{cases}
	\end{align*}
	where $\beta(\mS)\ge0$ and $\beta(\mS)=0$ when $\mS = \kappa\mI_r$ for some $\kappa>0$. 
	Also $\alpha$ is defined as
	\begin{align}
		\alpha(a,b) = \frac{b}{a-b-1}
		\label{eqn:alpha_def}
	\end{align}
\end{theorem}
We prove this theorem in \Secref{asubsec:trva_closedform}
First we prove the main result by using this theorem.
We note that similar results can be shown for different regimes of $k,d,\ntr$ and $\sigma$.
The result below is for one reasonable regime where $\ntr=\Omega(k)$ and $\sigma=\gO(1)$.
Similar results can be obtained if we further assume that $k \ll d$ with weaker conditions on $\sigma$, we leave that for future work.
Furthermore, this result can be extended to $\tau$-optimal solutions to the tr-val objective rather than just the optimal solution.

\begin{reptheorem}
{thm:trva_result1_robust}
	Let $\lambda=0$. Suppose $\ntr\ge 2k+2$ and $\sigma^2\in (0,\frac{\ntr-k-1}{3k})$, then any optimal solution $\flayer_{\best}\in\argmin\limits_{\flayer\in\R^{d\times d}} \Ltrvarep(\flayer;(\ntr,\nva))$ will satisfy 
	\begin{align*}
		\rank(\flayer_{\best}) = k,~~~ P_{\flayer_{\best}}\Astar = \Astar,~~~ \Ltrvarep(\flayer_{\best};(\ntr,\nva)) - \sigma^2 = \sigma^2\frac{k}{\ntr - k - 1}
	\end{align*}
	where $P_{\flayer} = \flayer\flayer^{\dagger}$ is the projection matrix onto the columnspace of $\flayer$.
	For any matrix $\flayer_{\good}\in\R^{d\times d}$ that is $\tau$-optimal, i.e. it satisfies $\Ltrvarep(\flayer_{\good};(\ntr,\nva)) \le \inf\limits_{\flayer\in\R^{d\times d}} \Ltrvarep(\flayer;(\ntr,\nva)) + \tau$ for $\tau\in(0,\frac{\sigma^2}{\ntr-k-1})$, then we have
	\begin{align*}
		\rank(\flayer_{\good}) = k,~~~ \|P_{\flayer_{\good}}\Astar - \Astar\|^2 \le \tau
	\end{align*}
	The meta-testing performance of $\flayer_{\good}$ on a new task with $\nte>2k+2$ samples satisfies
	\begin{align*}
		\inf_{\telambda\ge0}&\Ltestrep(\flayer_{\good};\nte) - \sigma^2 \le 2\tau + \sigma^2\frac{2k}{\nte}
	\end{align*}
\end{reptheorem}
\begin{proof}
	Suppose the optimal value of $\Ltrvarep(\flayer;(\ntr,\nva))$ is $\gL^*$, i.e.
	\begin{align}
		\gL^* = \inf\limits_{\flayer\in\R^{d\times d}} \Ltrvarep(\flayer;(\ntr,\nva))\label{eqn:opt_Ltrval}
	\end{align}
	Let $\flayer_{\good}\in\R^{d\times d}$ be the ``good'' matrix that is $\tau$-optimal, i.e. $\Ltrvarep(\flayer_{\good};(\ntr,\nva)) - \gL^* \le \tau$.
	Note that the result for $\flayer_{\best}$ follows from the result for $\tau=0$.

	We use the expression for $\Ltrvarep(\cdot; (\ntr,\nva)$ from \Thmref{thm:trva_closedform} to find properties for $\flayer_{\good}$ that can ensure that it is $\tau$-optimal.
	Consider a representation $\flayer\in\R^{d\times d}$ and let $\flayer=\mU\mS\mV^\top$ be its SVD with $\mU,\mV\in\R^{d\times r}$ and $\mS\in\R^{r\times r}$ where $r = \rank(\flayer)$.
	We consider 3 cases for $r$: $r < k$, $k \le r \le \ntr$ and $r > \ntr$, and find properties that can result in $\flayer$ being $\tau$-optimal in each of the 3 cases.
	For the ranges of $\ntr,\sigma,\tau$ in the theorem statement, it will turn out that when $r<k$ or $r>\ntr$, $\flayer$ must satisfy $\Ltrvarep(\flayer; (\ntr,\nva) > \gL^* + \tau$, thus concluding that $\rank(\flayer_{\good})$ cannot be in these ranges.

	To do so we analyze the optimal value of $\Ltrvarep(\cdot; (\ntr,\nva))$ that can be achieved for all three cases.
	We first start analyzing the most promising case of $k\le r \le \ntr$.
	
	\paragraph{Case 1: $k\le r \le \ntr$}

	For this case we can use the $r < \ntr-1$ regime from \Thmref{thm:trva_closedform}.
	Note that for $r\in\{\ntr-1,\ntr\}$, $\Ltrvarep(\flayer;(\ntr,\nva))$ is unbounded.
	Thus we get
	\begin{align}
		\Ltrvarep(\flayer;(\ntr,\nva)) - \sigma^2 = (1+\alpha(\ntr,r))\|P_{\mU}^{\perp}\Astar\|^2 + \alpha(\ntr,r)\sigma^2\label{eqn:basic_eqn_case1}
	\end{align}
	where $\alpha(\ntr,r)$ is defined in \Eqref{eqn:alpha_def}.
	Note that $\|P_{\mU}^{\perp}\Astar\| \ge 0$ and in fact equality can be achieved for every $r$ in this case since $r = \rank(\mU) \ge k = \rank(\Astar)$.
	Furthermore $\alpha(\ntr,r) = \frac{r}{\ntr - r - 1}$ is an increasing function of $r$, and thus $\alpha(\ntr,r) \ge \alpha(\ntr,k)$, which can also be achieved by picking $r = k$.
	Thus for the range of $k\le r \le \ntr$, we get
	\begin{align}
		\gL^* - \sigma^2 \le \min_{\substack{\flayer\text{ s.t. }\\k\le r \le \ntr}}\Ltrvarep(\flayer;(\ntr,\nva)) - \sigma^2 = \sigma^2\alpha(\ntr, k) = \sigma^2 \frac{k}{\ntr - k - 1}\label{eqn:case1_trval_bound}
	\end{align}
	with the minimum being achieved when $\rank(\flayer) = k$ and $\|P_{\mU}^{\perp}\Astar\|$, which is the same as $P_{\mU}\Astar = \Astar$.
	
	Suppose $r_{\good} = \rank(\flayer_{\good})$ lies in this range, i.e. $k\le r_{\good} \le \ntr$.
	Using the fact that $\flayer_{\good}$ is $\tau$-optimal, we can show an upper bound the $r_{\good}$ and $\|P_{\flayer_{\good}}\Astar - \Astar\|$.
	From the $\tau$-optimality condition and \Eqref{eqn:case1_trval_bound}, we get
	\begin{align}
		\tau 
		&\ge \Ltrvarep(\flayer_{\good};(\ntr,\nva)) - \gL^* \ge \Ltrvarep(\flayer_{\good};(\ntr,\nva)) - \sigma^2\alpha(\ntr, k)\\
		&\ge^{(a)} (1+\alpha(\ntr,r))\|P_{\mU}^{\perp}\Astar\|^2 + \sigma^2\alpha(\ntr, r_{\good}) - \sigma^2\alpha(\ntr, k)\\
		&= (1+\alpha(\ntr,r))\|P_{\flayer_{\good}}^{\perp}\Astar\|^2 + \sigma^2(\alpha(\ntr, r_{\good}) - \alpha(\ntr, k))\\
		&=^{(b)} \|P_{\flayer_{\good}}^{\perp}\Astar\|^2 + \sigma^2 \alpha'(\ntr, r') (r_{\good} - k)
		=\|P_{\flayer_{\good}}^{\perp}\Astar\|^2 + \sigma^2 \frac{r'(\ntr - 1)}{(\ntr - r' - 1)^2} (r_{\good} - k)\\
		&\ge^{(c)} \|P_{\flayer_{\good}}^{\perp}\Astar\|^2 + \sigma^2 \frac{k(\ntr - 1)}{(\ntr - k - 1)^2} (r_{\good} - k)
		\ge \|P_{\flayer_{\good}}^{\perp}\Astar\|^2 + \sigma^2 \frac{k}{(\ntr - k - 1)} (r_{\good} - k)\label{eqn:r_good_bound}
	\end{align}
	where $(a)$ follows from \Eqref{eqn:basic_eqn_case1} instantiated for $\flayer_{\good}$, $(b)$ follows from the mean value theorem applied to the continuous function $\alpha(\ntr, \cdot)$ with $r' \in [k, r_{\good}]$, $(c)$ follows by $k > r'$ and the monotonocity of $\alpha'(\ntr, \cdot)$.
	Thus from the inequality in \Eqref{eqn:r_good_bound}, we can get an upper bound on $r_{\good}$ as follows
	\begin{align}
		\tau
		\ge \sigma^2 \frac{k}{(\ntr - k - 1)} (r_{\good} - k)
		&\implies r_{\good}
		\le k + \frac{\tau (\ntr - k - 1)}{\sigma^2 k}
		\le^{(a)} k + \frac{1}{k}\\
		&\implies r_{\good} = k
	\end{align}
	where $(a)$ follows from the upper bound on $\tau$ in the theorem statement.
	We can also get an upper bound on $\|P_{\flayer_{\good}}^{\perp}\Astar\|$ using \Eqref{eqn:r_good_bound} as $\|P_{\flayer_{\good}}^{\perp}\Astar\|^2 \le \tau$.
	Thus any $\flayer_{\good}$ with $k\le r_{\good}\le \ntr$ that is $\tau$-optimal will satisfy $r_{\good} = k$ and $\|P_{\flayer_{\good}}^{\perp}\Astar\|^2\le \tau$.
	
	Additionally the minimum achievable value for this range of rank is $\sigma^2 \frac{k}{\ntr - k - 1}$ fromr \Eqref{eqn:case1_trval_bound}.
	We now analyze the other two cases, where we will show that no $\flayer$ can even be $\tau$-optimal.


	\paragraph{Case 2: $r < k$}

	In this case, since $k < \ntr$, we are still in the $r < \ntr$ regime.
	The key point here is that when $r<k$, it is impossible for $\flayer$ to span all of $\Astar$.
	In fact for rank $r$, it is clear that $\flayer$ can cover only at most $r$ out of $k$ directions from $\Astar$.
	Thus the inexpressiveness term $\|P^{\perp}_{\mU}\Astar\|^2$ will be at least $\frac{k - r}{k}\|\Astar\|^2 = \frac{k-r}{k}$.
	Using the $r < \ntr$ expression from \Thmref{thm:trva_closedform}, we get
	\begin{align}
		\Ltrvarep(\flayer &;(\ntr,\nva)) - \sigma^2
		= (1+\alpha(\ntr,r))\|P_{\mU}^{\perp}\Astar\|^2 + \alpha(\ntr,r)\sigma^2\nonumber\\
		&\ge^{(a)} \frac{k-r}{k} + \sigma^2\frac{r}{\ntr - r - 1}\nonumber\\
		&= \frac{k-r}{k} + \sigma^2\frac{r}{\ntr - r - 1} - \sigma^2\frac{k}{\ntr - r - 1} + \sigma^2\frac{k}{\ntr - r - 1} - \sigma^2\frac{k}{\ntr - k - 1} + \sigma^2\frac{k}{\ntr - k - 1}\nonumber\\
		&= \frac{k-r}{k} - \sigma^2\frac{k-r}{\ntr - r - 1} - \sigma^2k\left(\frac{1}{\ntr - k - 1} - \frac{1}{\ntr - r - 1}\right) + \sigma^2\frac{k}{\ntr - k - 1}\nonumber\\
		&= \frac{k-r}{k} - \sigma^2\frac{k-r}{\ntr - r - 1} - \sigma^2\frac{k(k-r)}{(\ntr - r - 1)(\ntr - k - 1)} + \sigma^2\frac{k}{\ntr - k - 1}\nonumber\\
		&\ge^{(b)} \frac{k-r}{k} - \sigma^2\frac{k-r}{\ntr - k - 1} - \sigma^2\frac{(k-r)}{(\ntr - k - 1)} + \sigma^2\frac{k}{\ntr - k - 1}\nonumber\\
		&= \frac{k-r}{k} - \sigma^2\frac{2(k-r)}{\ntr - k - 1} + \sigma^2\frac{k}{\ntr - k - 1}\label{eqn:case2_trval_bound}
	\end{align}
	where for $(a)$ we use $\alpha(\ntr,r)\ge 0$ and for $(b)$, we use that $k < \ntr - k -1 < \ntr - r - 1$.
	Since we assume $\sigma^2 < \nicefrac{(\ntr-k-1)}{2k}$, the difference between the first 2 terms is at least 0.
	Thus the error when $r<k$ is at least $\Ltrvarep(\flayer;(\ntr,\nva)) - \sigma^2 > \sigma^2\frac{k}{\ntr - k - 1}$, which is larger than the previous case.
	So the optimal solution cannot have $r < k$.

	We now check if such an $\flayer$ can be a $\tau$-optimal solution instead.
	The answer is negative due to the following calculation
	\begin{align*}
		\Ltrvarep(\flayer;(\ntr,\nva)) &- \gL^*
		\ge^{(a)} \Ltrvarep(\flayer;(\ntr,\nva)) - \sigma^2 \frac{k}{\ntr - k - 1}\\
		&\ge^{(b)} \frac{k-r}{k} - \sigma^2\frac{2(k-r)}{\ntr - k - 1}\\
		&\ge^{(c)}	\frac{3k\sigma^2}{\ntr - k - 1}\frac{k-r}{k} - \sigma^2\frac{2(k-r)}{\ntr - k - 1}\\
		&= \frac{\sigma^2(k-r)}{\ntr - k - 1}
		\ge \frac{\sigma^2}{\ntr - k - 1}
		> \tau
	\end{align*}
	where $(a)$ follows from \Eqref{eqn:case1_trval_bound}, $(b)$ follows from \Eqref{eqn:case2_trval_bound}, $(c)$ follows from $1 \ge \frac{3k\sigma^2}{\ntr - k - 1}$ from the condition on $\sigma$ in the theorem statement and the last inequalities follow from $r < k$ condition and the range of $\tau$ in the theorem statement respectively.
	Thus we cannot even have a $\tau$-optimal solution with $r < k$.
	Next we show the same for the case of $r > \ntr$.

	\paragraph{Case 3: $r > \ntr$}

	For this case we can use the $r > \ntr+1$ regime from \Thmref{thm:trva_closedform}.
	Note that for $r\in\{\ntr+1,\ntr\}$, $\Ltrvarep(\flayer;(\ntr,\nva))$ is unbounded.
	Thus we have $\Ltrvarep(\flayer;(\ntr,\nva)) - \sigma^2 = (\frac{\ntr}{r}+\alpha(r,\ntr))\|P_{\mU}^{\perp}\Astar\|^2 + \frac{r-\ntr}{r} + \beta(\mS) + \sigma^2\alpha(r,\ntr)$.
	We again note that since $r  > \ntr > k$, we can make $\|P_{\mU}^{\perp}\Astar\|^2 = 0$ by simply picking an ``expressive'' rank-$r$ subspace $\mU$.
	Further more $\beta(\mS)=0$ is easy to achieve by making $\mS = \mI_r$ which can be done independently of $\mU$.
	Thus we can lower bound the error as follows
	\begin{align}
		\Ltrvarep(\flayer;(\ntr,\nva)) - \sigma^2 \ge \frac{r-\ntr}{r} + \sigma^2\alpha(r,\ntr)
		&= 1 - \frac{\ntr}{r} + \sigma^2\frac{\ntr}{r - \ntr - 1}\label{eqn:trval_bound_case3}
	\end{align}
	where equality can be achieved by letting columns of $\flayer$ span the subspace $\Astar$ and by picking $\mS = I_r$.
	We now lower bound this value further to show that we cannot have a $\tau$-optimal solution.

	Firstly we note that if $\sigma\ge 1$, then \Eqref{eqn:trval_bound_case3} gives us
	\begin{align*}
		\Ltrvarep(\flayer;(\ntr,\nva)) - \sigma^2
		&\ge 1 - \frac{\ntr}{r} + \frac{\ntr}{r - \ntr - 1}
		\ge 1
		\ge^{(a)} \sigma^2 \frac{3k}{\ntr - k - 1}\\
		&\ge^{(b)} (\gL^* - \sigma^2) + \sigma^2 \frac{2k}{\ntr - k - 1}
		\ge^{(c)} (\gL^* - \sigma^2) + \tau (\ntr - k - 1) \frac{2k}{\ntr - k - 1}\\
		&> (\gL^* - \sigma^2) + \tau
	\end{align*}
	where $(a)$ follows from the upper bound on $\sigma^2$ from the theorem statement, $(b)$ follows from \Eqref{eqn:case1_trval_bound} and $(c)$ follows from the upper bound on $\tau$ from the theorem statement.
	Thus we cannot have a $\tau$-optimal solution in this case.

	If $\sigma<1$ on the other hand, we can lower bound the error as
	\begin{align*}
		\Ltrvarep(\flayer;(\ntr,\nva)) - \sigma^2 = 1 - \frac{\ntr}{r} + \sigma^2\frac{\ntr}{r - \ntr - 1}
		&> 1- \frac{\ntr}{r} + \sigma^2\frac{\ntr}{r - \ntr}\\
		&\ge \min_{r>\ntr} 1- \frac{\ntr}{r} + \sigma^2\frac{\ntr}{r - \ntr}
	\end{align*}
	Setting the derivate w.r.t. $r$ to 0 for the above expression, we get that this is minimized at $r = \nicefrac{\ntr}{(1-\sigma)}$.
	Plugging in this value and simplifying, we get
	\begin{align*}
		\Ltrvarep(\flayer;(\ntr,\nva)) - \sigma^2
		&\ge 1 - (1 - \sigma)^2 = 2\sigma - \sigma^2 \ge \sigma^2\\
		&= \sigma^2\frac{\ntr - 2k - 1}{\ntr - k - 1} + \sigma^2\frac{k}{\ntr - k - 1}
		>^{(a)} \frac{\sigma^2}{\ntr - k - 1} + (\gL^* - \sigma^2)\\
		&>^{(b)} \tau + (\gL^* - \sigma^2)\\
	\end{align*}
	where $(a)$ follows from the condition that $\ntr > 2k + 2$ and $(b)$ follows again from \Eqref{eqn:case1_trval_bound}.
	Thus the $r > \ntr$ case can never give us a $\tau$-optimal solution, let alone the optimal solution and so $\flayer_{\good}$ cannot have rank $r_{\good} > \ntr$ either.
	Combining all 3 cases, we can conclude that any $\tau$-optimal solution $\flayer_{\good}$ will satisfy $\rank(\flayer_{\good}) = k$ and $\|P_{\flayer_{\good}}^{\perp}\Astar\|^2 = \|P_{\flayer_{\good}}\Astar - \Astar\|^2 \le \tau$.

	
	
	For the second part, we use \Propref{prop:Ltrva_equal_Ltest} to first get that for $\lambda=\telambda=0$, $\Ltestrep(\flayer_{\good};\nte) = \Ltrvarep(\flayer_{\good}; (\nte,\cdot))$.
	Since $\nte > 2k+2$ and $\rank(\flayer_{\good}) = k < \nte$, we can use \Thmref{thm:trva_closedform} to get 
	\begin{align*}
		\Ltestrep(\flayer_{\good};\nte) - \sigma^2
		&= \Ltrvarep(\flayer_{\good}; (\nte,\cdot)) - \sigma^2
		= (1+\alpha(\nte,k))\|P_{\flayer_{\good}}^{\perp}\Astar\|^2 + \alpha(\nte,k)\sigma^2\\
		&\le^{(a)} 2\|P_{\flayer_{\good}}^{\perp}\Astar\|^2 + \sigma^2\frac{k}{\nte - k - 1}
		\le^{(b)} 2\tau + \sigma^2\frac{k}{\nte - \nicefrac{\nte}{2}}
		= 2\tau + \sigma^2\frac{2k}{\nte}
	\end{align*}
	where $(a)$ follows by noting that $\alpha(\nte, k) < 1$ and $(b)$ follows from $k+1 < \nte/2$.
	This completes the proof of result for $\tau$-optimal solution $\flayer_{\good}$.
	Setting $\tau=0$ gives results for optimal solution $\flayer_{\best}$ of $\Ltrvarep(\cdot,(\ntr,\nva))$.
\end{proof}

\subsection{Proof of \Thmref{thm:trva_closedform}}
\label{asubsec:trva_closedform}

We now prove the crucial result that gives a closed form solution for the tr-val objective
\begin{reptheorem}
{thm:trva_closedform}
	Let $\lambda=0$. For a first layer $\flayer\in\R^{d\times d}$ with $r = \rank(\flayer)$, let $\flayer = \mU\mS\mV^\top$ be the SVD, where $\mU,\mV\in\R^{d\times r}$ and $\mS\in\R^{r\times r}$.
	Furthermore let $P_{\mU}=\mU\mU^\top$ denote the span of columns of $\mU$ (and thus $\flayer$) and let $P^{\perp}_{\mU} = I_d - P_{\mU}$. Then the tr-val objective has the following form
	\begin{align*}
                	\Ltrvarep(\flayer;(\ntr,\nva)) - \sigma^2 = 
		\begin{cases}
			(1+\alpha(\ntr,r))\|P_{\mU}^{\perp}\Astar\|^2 + \alpha(\ntr,r)\sigma^2 & \text{if } r < \ntr-1\\
			&\\
			(\frac{\ntr}{r}+\alpha(r,\ntr))\|P_{\mU}^{\perp}\Astar\|^2 + \frac{r-\ntr}{r} + \beta(\mS) + \alpha(r,\ntr)\sigma^2 & \text{if } r > \ntr+1\\
			&\\
			\infty & \text{otherwise}
		\end{cases}
	\end{align*}
	where $\beta(\mS)\ge0$ and $\beta(\mS)=0$ when $\mS = \kappa\mI_r$ for some $\kappa>0$. 
	Also $\alpha$ is defined as
	\begin{align}
		\alpha(a,b) = \frac{b}{a-b-1}
		\label{eqn:alpha_def}
	\end{align}
\end{reptheorem}

\begin{proof}
	We will prove the result for the cases $r < \ntr$ and $r \ge \ntr$. Firstly, using \Propref{prop:Ltrva_equal_Ltest}, we already get that
	\begin{align}
		\Ltrvarep(\flayer;(\ntr,\nva)) = \Ltest_{\lambda,\rep}(\flayer;\ntr)
		&= \sigma^2 + \ex_{\dist_{\tclf}\sim\tedist} \left[\ex_{(\Xto,\Yto)\sim\dist_{\tclf}^{\ntr}} \left[\|\flayer\slayer_{\lambda}(\flayer;(\Xto,\Yto)) - \tclf\|^2\right]\right]\nonumber\\
		&= \sigma^2 + \ex_{\tclf\sim\gN(\vzero_d,\mI_d)} \left[\ex_{\substack{\Xto\sim\gN(\vzero_d,\mI_d)^{\ntr},\\\veta\sim\gN(\vzero_d,\sigma^2 \mI_n)}} \left[\|\flayer\slayer_{\lambda}(\flayer;(\Xto,\Xto\tclf+\veta)) - \tclf\|^2\right]\right]\label{eqn:trva_expansion}
	\end{align}
	We will use this expression in the rest of the proof.
	
	\paragraph{\underline{Case 1: $r \le \ntr$}}

	In this case, the rank of the representations $\Xto\flayer\in\R^{\ntr\times d}$ for training data is higher than the number of samples.
	Thus the unique minimizer for the dataset $(\Xto,\Yto)$ is
	\begin{align*}
	 	\lim_{\lambda\rightarrow0}\slayer_{\lambda}(\flayer;(\Xto,\Yto)) = \argmin_{\slayer\in\R^d} ~\ell_{\lambda,\rep}(\slayer; \flayer,(\Xto,\Yto)) = \left(\flayer^\top\Xto^\top\Xto\flayer\right)^{\dagger} \flayer^\top\Xto^\top\Yto
	\end{align*}
	Plugging this into \Eqref{eqn:trva_expansion}, we get
	\begin{align*}
		\Ltrvarep(\flayer &;(\ntr,\nva)) - \sigma^2
		= \ex_{\tclf\sim\gN(\vzero_d,\mI_d)} \left[\ex_{\substack{\Xto\sim\gN(\vzero_d,\mI_d)^{\ntr},\\\veta\sim\gN(\vzero_d,\sigma^2 \mI_{\ntr})}} \left[\|\flayer\left(\flayer^\top\Xto^\top\Xto\flayer\right)^{\dagger} \flayer^\top\Xto^\top(\Xto\tclf+\veta) - \tclf\|^2\right]\right]\\
		&=^{(a)}  \underbrace{\ex_{\tclf,\Xto} \left[\|\flayer\left(\flayer^\top\Xto^\top\Xto\flayer\right)^{\dagger} \flayer^\top\Xto^\top\Xto\tclf - \tclf\|^2\right]}_{\text{bias}(\flayer)}
		+ \underbrace{\ex_{\veta,\Xto} \left[\|\flayer\left(\flayer^\top\Xto^\top\Xto\flayer\right)^{\dagger} \flayer^\top\Xto^\top\veta\|^2\right]}_{\text{variance}(\flayer)} \\
	\end{align*}
	where $(a)$ follows from the independence of $\veta$ and $(\Xto,\tclf)$ and that $\ex_{}[\veta] = \vzero_n$.
	We will analyze the bias and variance terms separately below

	\textbf{Bias:}
	Recall that $\flayer=\mU\mS\mV^{\top}$ is the singular value decomposition, with $\mU,\mV\in\R^{d\times r}$ and $\mS\in\R^{r\times r}$.
	We set $\XtoU = \Xto\mU\in\R^{\ntr\times r}$
	The two key ideas that we will exploit are that $\XtoU\sim\gN(\vzero_d,\mI_r)^{\ntr}$ and that $\XtoU$ is independent of $\Xto P^{\perp}_{\mU}$ since $\Xto$ is Gaussian-distributed.
	\begin{align*}
		&\text{bias}(\flayer)
		= \ex_{\tclf,\Xto} \left[\|\flayer\left(\flayer^\top\Xto^\top\Xto\flayer\right)^{\dagger} \flayer^\top\Xto^\top\Xto\tclf - \tclf\|^2\right]\\
		&=^{(a)} \ex_{\tclf,\Xto} \left[\|\mU\mS\mV^{\top}\left(\mV\mS\mU^\top\Xto^\top\Xto\mU\mS\mV^{\top}\right)^{\dagger} \mV\mS\mU^\top\Xto^\top\Xto\tclf - \tclf\|^2\right]\\
		&=^{(b)} \ex_{\tclf,\Xto} \left[\|\mU\mS\left(\mS(\Xto\mU)^\top\Xto\mU\mS\right)^{-1}\mS(\Xto\mU)^\top\Xto\tclf - \tclf\|^2\right]\\
		&=^{(c)} \ex_{\tclf,\Xto} \left[\|\mU\mS\left(\mS\XtoU^\top\XtoU\mS\right)^{-1}\mS\XtoU^\top\Xto\tclf - \tclf\|^2\right]
		=^{(d)} \ex_{\tclf,\Xto} \left[\|\mU\mS\mS^{-1}(\XtoU^\top\XtoU)^{-1}\mS^{-1}\mS\XtoU^\top\Xto\tclf - \tclf\|^2\right]\\
		&=^{(e)} \ex_{\tclf,\Xto} \left[\|\mU(\XtoU^\top\XtoU)^{-1}\XtoU^\top\Xto\tclf - \tclf\|^2\right]
		= \ex_{\tclf,\Xto} \left[\|\mU(\XtoU^\top\XtoU)^{-1}\XtoU^\top\Xto(\mU\mU^\top + P^{\perp}_{\mU})\tclf - (\mU\mU^\top + P^{\perp}_{\mU})\tclf\|^2\right]\\
		&=^{(f)} \ex_{\tclf,\Xto} \left[\|\mU(\XtoU^\top\XtoU)^{-1}\XtoU^\top\Xto\mU\mU^\top\tclf + \mU(\XtoU^\top\XtoU)^{-1}\XtoU^\top\Xto P^{\perp}_{\mU}\tclf - \mU\mU^\top\tclf - P^{\perp}_{\mU}\tclf\|^2\right]\\
		&=^{(g)} \ex_{\tclf,\Xto} \left[\|\mU\mU^\top\tclf - \mU\mU^\top\tclf + \mU(\XtoU^\top\XtoU)^{-1}\XtoU^\top\Xto P^{\perp}_{\mU}\tclf\|^2 + \|P^{\perp}_{\mU}\tclf\|^2\right]\\
		&=^{(h)} \ex_{\tclf,\Xto} \left[\tr\left((P^{\perp}_{\mU}\tclf)^\top P^{\perp}_{\mU}\Xto^\top\XtoU (\XtoU^\top\XtoU)^{-2} \XtoU^\top\Xto P^{\perp}_{\mU}(P^{\perp}_{\mU}\tclf)\right)\right] + \|P^{\perp}_{\mU}\Astar\|^2\\
		&=^{(i)} \ex_{\tclf,\XtoU}\left[\tr\left(\XtoU (\XtoU^\top\XtoU)^{-2} \XtoU^\top \ex_{P^{\perp}_{\mU}\Xto}\left[\Xto P^{\perp}_{\mU}(P^{\perp}_{\mU}\tclf)(P^{\perp}_{\mU}\tclf)^\top P^{\perp}_{\mU}\Xto^\top\right]\right)\right] + \|P^{\perp}_{\mU}\Astar\|^2
	\end{align*}
	while $(a)$ just uses the SVD of $\flayer$, $(b)$ uses the simple fact that $\mV^\top(\mV\mB\mV^\top)^{\dagger}\mV = \mB^{\dagger}$ for an orthogonal matrix $\mV$.
	$(d)$ follows from the fact that $\XtoU^\top\XtoU\in\R^{r \times r}$ is full rank with probability 1, and thus invertible.
	$(f)$ simply decomposes $\tclf = \mU\mU^\top\tclf + P^{\perp}_{\mU}\tclf$, while $(g)$ follows from the orthogonality of $P^{\perp}_{\mU}\tclf$ and any vector in the span of $\mU$.
	$(h)$ uses the simple facts that $\|\va\|^2 = \tr(\va\va^\top)$ and $P^{\perp}_{\mU}P^{\perp}_{\mU} = P^{\perp}_{\mU}$ and $(i)$ uses the crucial observation that $\XtoU$ is independent of $\Xto P^{\perp}_{\mU}$, since for Gaussian distribution, all subspaces are independent of its orthogonal subspace, and that $\tr$ is a linear operator.
	
	We now look closer at the term $\mM = \ex\left[\Xto P^{\perp}_{\mU}(P^{\perp}_{\mU}\tclf)(P^{\perp}_{\mU}\tclf)^\top P^{\perp}_{\mU}\Xto^\top\right]$ which is a matrix in $\R^{\ntr\times\ntr}$.
	\begin{align}
		\mM_{i,j}
		&= \ex_{P^{\perp}_{\mU}\Xto} \left[\vx_i^\top (P^{\perp}_{\mU}\tclf)(P^{\perp}_{\mU}\tclf)^\top \vx_j\right]\nonumber\\
		&= \tr\left((P^{\perp}_{\mU}\tclf)(P^{\perp}_{\mU}\tclf)^\top \ex\left[ \vx_j\vx_i^\top\right]\right)
		= \begin{cases}
			0 & \text{if } i\neq j\\
			\|P^{\perp}_{\mU}\tclf\|^2 & \text{if } i=j
		\end{cases}
		\label{eqn:perp_term}
	\end{align}
	This gives us that $\mM = \|P^{\perp}_{\mU}\tclf\|^2 ~\mI_{\ntr}$.
	We can now complete the computation for $\text{bias}(\flayer)$.
	\begin{align*}
		\text{bias}(\flayer)
		&= \ex_{\tclf,\XtoU}\left[\tr\left(\XtoU (\XtoU^\top\XtoU)^{-2} \XtoU^\top\|P^{\perp}_{\mU}\tclf\|^2\right)\right] + \|P^{\perp}_{\mU}\Astar\|^2\\
		&= \ex_{\XtoU}\left[\tr\left(\XtoU (\XtoU^\top\XtoU)^{-2} \XtoU^\top\right)\right] \ex_{\tclf}\left[\|P^{\perp}_{\mU}\tclf\|^2\right] + \|P^{\perp}_{\mU}\Astar\|^2\\
		&= \ex_{\XtoU}\left[\tr\left((\XtoU^\top\XtoU)^{-1}\right)\right] \|P^{\perp}_{\mU}\Astar\|^2 + \|P^{\perp}_{\mU}\Astar\|^2\\
	\end{align*}
	We will deal with the $\ex_{\XtoU}\left[\tr\left((\XtoU^\top\XtoU)^{-1}\right)\right]$ term later and show that it is equal to $\alpha(\ntr,r)$.

	\textbf{Variance:}
	We will use many ideas that were used for the bias term. Again using the SVD, we get
	\begin{align*}
		\text{variance}(\flayer)
		&= \ex_{\veta,\Xto} \left[\|\flayer\left(\flayer^\top\Xto^\top\Xto\flayer\right)^{\dagger} \flayer^\top\Xto^\top\veta\|^2\right]\\
		&=^{(a)} \ex_{\veta,\XtoU} \left[\|\mU\mS\mV^\top\left(\mV\mS\XtoU^\top\XtoU\mS\mV^\top\right)^{\dagger} \mV\mS\XtoU^\top\veta\|^2\right]\\
		&=^{(b)} \ex_{\veta,\XtoU} \left[\|\mU\mS\mS^{-1}(\XtoU^\top\XtoU)^{-1}\mS^{-1}\mS\XtoU^\top\veta\|^2\right]
		=^{(c)} \ex_{\veta,\XtoU} \left[\|(\XtoU^\top\XtoU)^{-1}\XtoU^\top\veta\|^2\right]\\
		&=^{(d)} \ex_{\veta,\XtoU} \left[\tr\left((\XtoU^\top\XtoU)^{-1}\XtoU^\top\veta\veta^\top\XtoU(\XtoU^\top\XtoU)^{-1}\right)\right]\\
		&=^{(e)} \sigma^2\ex_{\XtoU} \left[\tr\left((\XtoU^\top\XtoU)^{-1}\XtoU^\top\XtoU(\XtoU^\top\XtoU)^{-1}\right)\right]\\
		&=^{(f)} \sigma^2\ex_{\XtoU}\left[\tr\left((\XtoU^\top\XtoU)^{-1}\right)\right]
	\end{align*} 
	Here $(a)$ uses SVD, $(b)$ uses the fact that as before $\XtoU^\top\XtoU$ is full rank with probability 1, $(d)$ follows from the norm and trace relationship, $(e)$ follows from the noise distribution $\veta\sim\gN(\vzero_{\ntr}, \sigma^2\mI_{\ntr})$ and its independence from $\Xto$.
	
	Thus combining the bias and variance terms, we get 
	\begin{align*}
		\Ltrvarep(\flayer;(\ntr,\va)) -\sigma^2 
		&= \text{bias}(\flayer) + \text{variance}(\flayer)\\
		&=\left(1 + \ex_{\XtoU}\left[\tr\left((\XtoU^\top\XtoU)^{-1}\right)\right]\right) \|P^{\perp}_{\mU}\|^2 + \ex_{\XtoU}\left[\tr\left((\XtoU^\top\XtoU)^{-1}\right)\right] \sigma^2
	\end{align*}

	Thus the only thing remaining to show is that $\ex_{\XtoU}\left[\tr\left((\XtoU^\top\XtoU)^{-1}\right)\right] = \alpha(\ntr, r)$.	
	To show this, we will use the fact that $(\XtoU^\top\XtoU)^{-1}$ is from the inverse Wishart distribution, and that $\ex_{\XtoU}(\XtoU^\top\XtoU)^{-1} = \frac{\mI_{\ntr}}{r - \ntr -1}$ when $r > \ntr + 1$ and unbounded when $r \in \{\ntr, \ntr+1\}$ \citep{mardia1979multivariate,belkin2020two}.
	For $r < \ntr - 1$, this gives us $\ex_{\XtoU}\left[\tr\left((\XtoU^\top\XtoU)^{-1}\right)\right] = \tr\left(\ex_{\XtoU}\left[(\XtoU^\top\XtoU)^{-1}\right]\right) = \tr\left(\frac{\mI_{\ntr}}{r - \ntr -1}\right) = \frac{\ntr}{r - \ntr - 1}$, which completes the proof.
	We now prove the result for $r> \ntr$.

	\paragraph{\underline{Case 2: $r > \ntr$}}
	In this case, the rank of the representations $\Xto\flayer\in\R^{\ntr\times d}$ for training data is lower than the number of samples.
	Thus we can use the dual formulation to get the minimum $\ell_2$ norm solution for dataset $(\Xto,\Yto)$
	\begin{align*}
	 	\lim_{\lambda\rightarrow0}\slayer_{\lambda}(\flayer;(\Xto,\Yto)) = \argmin_{\substack{\Xto\flayer\slayer = \Yto}} \|\slayer\|_2 = \flayer^\top\Xto^\top\left(\Xto\flayer\flayer^\top\Xto^\top\right)^{-1} \Yto
	\end{align*}
	Plugging this into \Eqref{eqn:trva_expansion}, we get
	\begin{align*}
		\Ltrvarep(\flayer &;(\ntr,\nva)) - \sigma^2
		= \ex_{\tclf\sim\gN(\vzero_d,\mI_d)} \left[\ex_{\substack{\Xto\sim\gN(\vzero_d,\mI_d)^{\ntr},\\\veta\sim\gN(\vzero_d,\sigma^2 \mI_{\ntr})}} \left[\|\flayer\flayer^\top\Xto^\top\left(\Xto\flayer\flayer^\top\Xto^\top\right)^{-1}(\Xto\tclf+\veta) - \tclf\|^2\right]\right]\\
		&=^{(a)}  \underbrace{\ex_{\tclf,\Xto} \left[\|\flayer\flayer^\top\Xto^\top\left(\Xto\flayer\flayer^\top\Xto^\top\right)^{-1}\Xto\tclf - \tclf\|^2\right]}_{\text{bias}(\flayer)}
		+ \underbrace{\ex_{\veta,\Xto} \left[\|\flayer\flayer^\top\Xto^\top\left(\Xto\flayer\flayer^\top\Xto^\top\right)^{-1}\veta\|^2\right]}_{\text{variance}(\flayer)} \\
	\end{align*}
	We again handle the bias and variance terms separately. Let $\XtoUp = \Xto P^{\perp}_{\mU}$ and we will again use the fact that $\XtoU=\Xto\mU\mU^\top$ and $\XtoUp$ are independent
	
	\textbf{Bias:}
	\begin{align*}
		&\text{bias}(\flayer)
		= \ex_{\tclf,\Xto} \left[\|\flayer\flayer^\top\Xto^\top\left(\Xto\flayer\flayer^\top\Xto^\top\right)^{-1}\Xto\tclf - \tclf\|^2\right]\\
		&= \ex_{\tclf,\Xto} \left[\|\mU\mS^2\mU^\top\Xto^\top\left(\Xto\mU\mS^2\mU^\top\Xto^\top\right)^{-1}\Xto(\mU\mU^\top + P^{\perp}_{\mU})\tclf - (\mU\mU^\top + P^{\perp}_{\mU})\tclf\|^2\right]\\
		&= \ex_{\tclf,\Xto} \left[\|\mU(\mS^2\XtoU^\top\left(\XtoU\mS^2\XtoU^\top\right)^{-1}\XtoU - \mI_r)\mU^\top\tclf + \mU\mS^2\XtoU^\top\left(\XtoU\mS^2\XtoU^\top\right)^{-1}\Xto P^{\perp}_{\mU}\tclf - P^{\perp}_{\mU}\tclf\|^2\right]\\
		&= \ex_{\tclf,\Xto} \left[\|\mU(\mS^2\XtoU^\top\left(\XtoU\mS^2\XtoU^\top\right)^{-1}\XtoU - \mI_r)\mU^\top\tclf + \mU\mS^2\XtoU^\top\left(\XtoU\mS^2\XtoU^\top\right)^{-1}\XtoUp P^{\perp}_{\mU}\tclf\|^2 + \|P^{\perp}_{\mU}\tclf\|^2\right]\\
		&= \ex_{\tclf,\Xto} \left[\|\mB_{\XtoU}\mU^\top\tclf + \mC_{\XtoU}\XtoUp P^{\perp}_{\mU}\tclf\|^2\right] + \|P^{\perp}_{\mU}\Astar\|^2\\
	\end{align*}
	where $\mB_{\XtoU}=\mU(\mS^2\XtoU^\top\left(\XtoU\mS^2\XtoU^\top\right)^{-1}\XtoU - \mI_r)$ and $\mC_{\XtoU}=\mU\mS^2\XtoU^\top\left(\XtoU\mS^2\XtoU^\top\right)^{-1}$ only depends on the components of $\Xto$ in the direction of $\mU$, i.e $\XtoU$.
	The main difference from the $r < \ntr$ case is that here $\mB_{\XtoU}\in\R^{r\times r}$ is not zero, since the rank of $\XtoU$ is $\min\{\ntr,r\} = \ntr$.
	We can expand the bias term further
	\begin{align}
		&\text{bias}(\flayer)
		= \ex_{\tclf,\Xto} \left[\|\mB_{\XtoU}\mU^\top\tclf\|^2 + \|\mC_{\XtoU}\XtoUp P^{\perp}_{\mU}\tclf\|^2 + 2\tr\left(\tclf^\top\mU\mB_{\XtoU}^\top \mC_{\XtoU}\XtoUp P^{\perp}_{\mU}\tclf\right)\right] + \|P^{\perp}_{\mU}\Astar\|^2\nonumber\\
		&=^{(a)} \ex_{\tclf,\XtoU} \left[\|\mB_{\XtoU}\mU^\top\tclf\|^2\right] 
			+\ex_{\tclf,\Xto}\left[\|\mC_{\XtoU}\XtoUp P^{\perp}_{\mU}\tclf\|^2\right] \nonumber\\
			&~~~~~~~+ 2\ex_{\tclf,\XtoU} \left[\tr\left(\tclf^\top\mU\mB_{\XtoU}^\top \mC_{\XtoU}\ex_{\XtoUp}\left[\XtoUp\right] P^{\perp}_{\mU}\tclf\right)\right]
			+ \|P^{\perp}_{\mU}\Astar\|^2\nonumber\\
		&=^{(b)} \ex_{\tclf,\XtoU} \left[\|\mB_{\XtoU}\mU^\top\tclf\|^2\right] 
			+ \ex_{\tclf,\Xto}\left[\|\mC_{\XtoU}\XtoUp P^{\perp}_{\mU}\tclf\|^2\right]
			+ \|P^{\perp}_{\mU}\Astar\|^2\nonumber\\
		&=^{(c)} \ex_{\tclf,\XtoU} \left[\|\mB_{\XtoU}\mU^\top\tclf\|^2\right] 
			+ \ex_{\tclf,\Xto}\left[\tr\left(\tclf^{\top} P^{\perp}_{\mU}\XtoUp^\top \mC_{\XtoU}^{\top}\mC_{\XtoU}\XtoUp P^{\perp}_{\mU}\tclf\right)\right]
			+ \|P^{\perp}_{\mU}\Astar\|^2\nonumber\\
		&=^{(d)} \ex_{\tclf,\XtoU} \left[\|\mB_{\XtoU}\mU^\top\tclf\|^2\right] 
			+ \ex_{\tclf,\XtoU}\left[\tr\left(\mC_{\XtoU}^{\top}\mC_{\XtoU} \ex_{\XtoUp}\left[\XtoUp P^{\perp}_{\mU}\tclf\tclf^{\top}P^{\perp}_{\mU}\XtoUp^{\top}\right]\right)\right]
			+ \|P^{\perp}_{\mU}\Astar\|^2\nonumber\\
		&=^{(e)} \ex_{\tclf,\XtoU} \left[\|\mB_{\XtoU}\mU^\top\tclf\|^2\right] 
			+ \ex_{\XtoU}\left[\tr\left(\mC_{\XtoU}^{\top}\mC_{\XtoU}\right)\right] \|P^{\perp}_{\mU}\Astar\|^2
			+ \|P^{\perp}_{\mU}\Astar\|^2\label{eqn:trva_case2_total}
	\end{align}
	where $(a)$ uses the fact that $\XtoUp$ is independent of $\XtoU$ and $\tclf$, $(b)$ uses the Gaussianity of $\XtoUp$ and that it has mean $\vzero$, $(c)$ uses $\|\va\|^2 = \tr(\va\va^\top)$, $(d)$ uses the independence of $\XtoU$ and $\XtoUp$ and $(e)$ uses the calculation from \Eqref{eqn:perp_term}.
	We first tackle the $\mC_{\XtoU}$ term
	\begin{align}
		\ex_{\XtoU}\left[\tr\left(\mC_{\XtoU}^{\top}\mC_{\XtoU}\right)\right]
		&= \ex_{\XtoU}\left[\tr\left(\left(\XtoU\mS^2\XtoU^\top\right)^{-1}\XtoU\mS^2\mU^{\top}\mU\mS^2\XtoU^\top\left(\XtoU\mS^2\XtoU^\top\right)^{-1}\right)\right]\nonumber\\
		&= \ex_{\XtoU}\left[\tr\left(\left(\XtoU\mS^2\XtoU^\top\right)^{-1}\XtoU\mS^4\XtoU^\top\left(\XtoU\mS^2\XtoU^\top\right)^{-1}\right)\right]
		\label{eqn:CXU_term}
	\end{align}
	We will encounter this function again in the variance term and we will tackle this later.
	At a high level, we will show that this term is going to be at least as large as the term for $\mS=\mI_r$, which reduces to $\ex_{\XtoU}\left[\tr\left(\left(\XtoU\XtoU^\top\right)^{-1}\right)\right]$ which has a closed form expression, again using inverse Wishart distribution.
	For now we will first deal with the $\mB_{\XtoU}$ term.
	\begin{align*}
		\ex_{\tclf,\XtoU} \left[\|\mB_{\XtoU}\mU^\top\tclf\|^2\right] 
		&= \ex_{\substack{\vu\sim\gN(\vzero_r,\mU^\top\Astar\Astar^\top\mU),\\\XtoU}} \left[\|\mB_{\XtoU}\vu\|^2\right]
		= \ex_{\vu,\XtoU} \left[\|\mU(\mS^2\XtoU^\top\left(\XtoU\mS^2\XtoU^\top\right)^{-1}\XtoU - \mI_r)\vu\|^2\right]\\
		&= \ex_{\vu,\XtoU} \left[\|(\mS^2\XtoU^\top\left(\XtoU\mS^2\XtoU^\top\right)^{-1}\XtoU - \mI_r)\vu\|^2\right]\\
	\end{align*}
	We are now going to exploit the symmetry in Gaussian distribution once again.
	Recall that $\XtoU\sim\gN(\vzero_r,\mI_r)$.
	We notice that $\XtoU \equiv_{D} \XtoU\mP\mD$, where $\mP$ is a random permutation matrix and $\mD$ is a diagonal matrix with random entries in $\pm1$.
	Essentially this is saying that randomly shuffling the coordinates and multiplying each coordinate by a random sign results in the same isotropic Gaussian distribution.
	We observe that $\mP\mD\mS^2\mD\mP^\top$ for diagonal matrix $\mS$ and that $\mP\mP^\top = \mP^\top\mP = \mD^2 = \mI_r$ rewrite the above expectation.
	\begin{align*}
		&\ex_{\tclf,\XtoU} \left[\|\mB_{\XtoU}\mU^\top\tclf\|^2\right]
		= \ex_{\vu,\XtoU} \left[\|(\mS^2\XtoU^\top\left(\XtoU\mS^2\XtoU^\top\right)^{-1}\XtoU - \mI_r)\vu\|^2\right]\\
		&= \ex_{\vu,\XtoU,\mP,\mD} \left[\|(\mS^2\mD\mP^\top\XtoU^\top\left(\XtoU\mP\mD\mS^2\mD\mP^\top\XtoU^\top\right)^{-1}\XtoU\mP\mD - \mI_r)\vu\|^2\right]\\
		&= \ex_{\vu,\XtoU,\mP,\mD} \left[\|(\mS^2\mD\mP^\top\XtoU^\top\left(\XtoU\mS^2\XtoU^\top\right)^{-1}\XtoU\mP\mD - \mD\mP^\top\mP\mD)\vu\|^2\right]\\
		&= \ex_{\vu,\XtoU,\mP,\mD} \left[\|\mD\mP^\top(\mP\mD\mS^2\mD\mP^\top\XtoU^\top\left(\XtoU\mS^2\XtoU^\top\right)^{-1}\XtoU - \mI_r)\mP\mD\vu\|^2\right]\\
		&= \ex_{\vu,\XtoU,\mP,\mD} \left[\|\mD\mP^\top(\mS^2\XtoU^\top\left(\XtoU\mS^2\XtoU^\top\right)^{-1}\XtoU - \mI_r)\mP\mD\vu\|^2\right]\\
		&= \ex_{\vu,\XtoU,\mP,\mD} \left[\|(\mS^2\XtoU^\top\left(\XtoU\mS^2\XtoU^\top\right)^{-1}\XtoU - \mI_r)\mP\mD\vu\|^2\right]\\
		&= \ex_{\vu,\XtoU} \left[\tr\left((\mS^2\XtoU^\top\left(\XtoU\mS^2\XtoU^\top\right)^{-1}\XtoU - \mI_r)\ex_{\mP,\mD}\left[\mP\mD\vu\vu^\top\mD\mP^\top\right](\mS^2\XtoU^\top\left(\XtoU\mS^2\XtoU^\top\right)^{-1}\XtoU - \mI_r)^\top\right)\right]
	\end{align*}
	It is not hard to see that randomly multiplying coordinates of $\vu$ by $\pm 1$ and then shuffling the coordinates will lead to $\ex_{\mP,\mD}\left[\mP\mD\vu\vu^\top\mD\mP^\top\right] = \frac{\|\vu\|^2}{r}\mI_r$.
	\begin{align}
		\ex_{\tclf,\XtoU} \left[\|\mB_{\XtoU}\mU^\top\tclf\|^2\right]
		&= \ex_{\vu\sim\gN(\vzero_r,\mU^\top\Astar\Astar^\top\mU)}\left[\frac{\|\vu\|^2}{r}\right]\ex_{\XtoU}\left[\|\mS^2\XtoU^\top\left(\XtoU\mS^2\XtoU^\top\right)^{-1}\XtoU - \mI_r\|^2\right]\nonumber\\
		&= \frac{\|P_{\mU}\Astar\|^2}{r}\ex_{\XtoU}\left[\|\mS^2\XtoU^\top\left(\XtoU\mS^2\XtoU^\top\right)^{-1}\XtoU P_{\XtoU} - P_{\XtoU} - P^{\perp}_{\XtoU}\|^2\right]\nonumber\\
		&= \frac{\|P_{\mU}\Astar\|^2}{r}\ex_{\XtoU}\left[\|\mS^2\XtoU^\top\left(\XtoU\mS^2\XtoU^\top\right)^{-1}\XtoU - P_{\XtoU}\|^2 + \|P^{\perp}_{\XtoU}\|^2\right]\nonumber\\
		&= \beta_1(\mS) + \frac{\|P_{\mU}\Astar\|^2}{r}\ex_{\XtoU}\left[\rank(P^{\perp}_{\XtoU})\right]\nonumber\\
		&= \beta_1(\mS) + \frac{r-\ntr}{r}\|P_{\mU}\Astar\|^2\label{eqn:BXU_term}
	\end{align}
	where $\beta_1(\mS) = \frac{\|P_{\mU}\Astar\|^2}{r}\ex_{\XtoU}\left[\|\mS^2\XtoU^\top\left(\XtoU\mS^2\XtoU^\top\right)^{-1}\XtoU - P_{\XtoU}\|^2\right]$ is a function that satisfies $\beta_1(\mS) \ge 0$ and $\beta_1(\kappa\mI_r) = 0$ for any $\kappa$.
	Also we used that $\rank(P^{\perp}_{\XtoU}) = r - \rank(\XtoU) = r - \ntr$ with probability 1.

	Plugging \twoEqref{eqn:CXU_term}{eqn:BXU_term} into \Eqref{eqn:trva_case2_total}, we get the following final expression for the bias
	\begin{align}
		&\text{bias}(\flayer)\nonumber\\
		&= \left(1 + \ex_{\XtoU}\left[\tr\left(\left(\XtoU\mS^2\XtoU^\top\right)^{-1}\XtoU\mS^4\XtoU^\top\left(\XtoU\mS^2\XtoU^\top\right)^{-1}\right)\right]\right)\|P^{\perp}_{\mU}\Astar\|^2 + \frac{r-\ntr}{r}\|P_{\mU}\Astar\|^2 + \beta_1(\mS)\nonumber\\
		&= \left(1 + \ex_{\XtoU}\left[\tr\left(\left(\XtoU\mS^2\XtoU^\top\right)^{-1}\XtoU\mS^4\XtoU^\top\left(\XtoU\mS^2\XtoU^\top\right)^{-1}\right)\right]\right)\|P^{\perp}_{\mU}\Astar\|^2 + \frac{r-\ntr}{r}\left(1 - \|P^{\perp}_{\mU}\Astar\|^2\right) + \beta_1(\mS)\nonumber\\
		&= \left(1 + \ex_{\XtoU}\left[\tr\left(\left(\XtoU\mS^2\XtoU^\top\right)^{-1}\XtoU\mS^4\XtoU^\top\left(\XtoU\mS^2\XtoU^\top\right)^{-1}\right)\right] - \frac{r-\ntr}{r}\right)\|P^{\perp}_{\mU}\Astar\|^2 + \frac{r-\ntr}{r} + \beta_1(\mS)\nonumber\\
		&= \left(\frac{\ntr}{r} + \ex_{\XtoU}\left[\tr\left(\left(\XtoU\mS^2\XtoU^\top\right)^{-1}\XtoU\mS^4\XtoU^\top\left(\XtoU\mS^2\XtoU^\top\right)^{-1}\right)\right]\right)\|P^{\perp}_{\mU}\Astar\|^2 + \frac{r-\ntr}{r} + \beta_1(\mS)\label{eqn:trval_bias_final}
	\end{align}

	\textbf{Variance:}
	We now move to the variance term
	\begin{align}
		\text{variance}(\flayer)
		&= \ex_{\veta,\Xto} \left[\|\flayer\flayer^\top\Xto^\top\left(\Xto\flayer\flayer^\top\Xto^\top\right)^{-1}\veta\|^2\right]\nonumber\\
		&= \sigma^2\ex_{\Xto} \left[\tr\left(\flayer\flayer^\top\Xto^\top\left(\Xto\flayer\flayer^\top\Xto^\top\right)^{-2}\Xto\flayer\flayer^\top\right)\right]\nonumber\\
		&= \sigma^2\ex_{\Xto} \left[\tr\left(\mU\mS^2\mU^\top\Xto^\top\left(\Xto\mU\mS^2\mU^\top\Xto^\top\right)^{-2}\Xto\mU\mS^2\mU^\top\right)\right]\nonumber\\
		&= \sigma^2\ex_{\XtoU} \left[\tr\left(\mS^2\XtoU^\top\left(\XtoU\mS^2\XtoU^\top\right)^{-2}\XtoU\mS^2\right)\right]\nonumber\\
		&= \sigma^2\ex_{\XtoU} \left[\tr\left(\left(\XtoU\mS^2\XtoU^\top\right)^{-1}\XtoU\mS^4\XtoU^\top\left(\XtoU\mS^2\XtoU^\top\right)^{-1}\right)\right]\label{eqn:trval_var_final}
	\end{align}
	
	To further simply both, the bias and variance terms, we need the following result
	\begin{lemma}
	\label{lem:S_lower_bound}
		For $\XtoU\sim\gN(\vzero_r, \mI_r)^{\ntr}$ and diagonal matrix $\mS\in\R^{r\times r}$
		\begin{align*}
			\ex_{\XtoU} \left[\tr\left(\left(\XtoU\mS^2\XtoU^\top\right)^{-1}\XtoU\mS^4\XtoU^\top\left(\XtoU\mS^2\XtoU^\top\right)^{-1}\right)\right]
			&= \begin{cases}
				\alpha(r,\ntr) + \beta_2(\mS) & \text{if } r > \ntr + 1\\
				\infty & \text{if } r \in \{\ntr,\ntr+1\}
			\end{cases}
		\end{align*}
		where $\alpha(r,\ntr)$ is defined in \Eqref{eqn:alpha_def} and $\beta_2(\mS)\ge0$ and $\beta_2(\kappa\mI_r)=0$ for any $\kappa>0$.
	\end{lemma}
	Using \Lemref{lem:S_lower_bound} and plugging it into \twoEqref{eqn:trval_bias_final}{eqn:trval_var_final}, we get 
	\begin{align*}
		\Ltrvarep(\flayer;(\ntr,\va)) - \sigma^2 = \text{bias}(\flayer) + \text{variance}(\flayer)
		&= \left(\frac{\ntr}{r} + \alpha(r,\ntr)\right)\|P^{\perp}_{\mU}\Astar\|^2 + \frac{r-\ntr}{r} + \sigma^2\alpha(r,\ntr) + \beta(\mS)
	\end{align*}
	where $\beta(\mS) = \beta_1(\mS) + \beta_2(\mS)\left(\|P^{\perp}_{\mU}\Astar\|^2 + \sigma^2\right)$ is a non-negative function that is 0 at $\kappa\mI_r$ for all $\kappa>0$.
	
	We now complete the proof by proving \Lemref{lem:S_lower_bound}
	\begin{proof}[Proof of \Lemref{lem:S_lower_bound}]
		Let the L.H.S. be $\gamma(\mS) = \ex_{\XtoU} \left[\tr\left(\left(\XtoU\mS^2\XtoU^\top\right)^{-1}\XtoU\mS^4\XtoU^\top\left(\XtoU\mS^2\XtoU^\top\right)^{-1}\right)\right]$.
		We first show that equality holds for $\mS=\kappa\mI_d$.
		In this case, we have
		\begin{align}
			\gamma(\kappa\mI_r) = \ex_{\XtoU} \left[\tr\left(\left(\XtoU\XtoU^\top\right)^{-1}\XtoU\XtoU^\top\left(\XtoU\XtoU^\top\right)^{-1}\right)\right] = \ex_{\XtoU} \left[\tr\left(\left(\XtoU\XtoU^\top\right)^{-1}\right)\right].
		\end{align}
		Using the closed-form expression for inverse Wishart distribution once again, we conclude
		\begin{align*}
			\gamma(\kappa\mI_r)
			=\ex_{\XtoU} \left[\tr\left(\left(\XtoU\XtoU^\top\right)^{-1}\right)\right]
			&= \tr\left(\ex_{\XtoU}\left[\left(\XtoU\XtoU^\top\right)^{-1}\right]\right) 
			= \begin{cases}
				\tr\left(\frac{\mI_{\ntr}}{r - \ntr - 1}\right) = \frac{\ntr}{r - \ntr - 1} & \text{if } r > \ntr + 1\\
				\infty & \text{otherwise}
			\end{cases}
		\end{align*}
		Thus we get $\gamma(\kappa\mI_r) = \alpha(r,\ntr)$ when $r > \ntr + 1$ and unbounded otherwise.

		We will show for an arbitrary diagonal matrix $\mS$ that the value is at least as large as $\mI_r$, i.e. $\gamma(\mS) \ge \gamma(\kappa \mI_r)$, which will complete the proof.
		We first observe that $\XtoU\mS^4\XtoU^\top = \XtoU\mS^2\mS^2\XtoU^\top \psd \XtoU\mS^2P_{\XtoU}\mS^2\XtoU^\top = \XtoU\mS^2\XtoU^\top\left(\XtoU\XtoU^\top\right)^{-1}\XtoU\mS^2\XtoU^\top$.
		Using this, we get
		\begin{align*}
			\gamma(\mS)
			&= \ex_{\XtoU}\left[\tr\left(\left(\XtoU\mS^2\XtoU^\top\right)^{-1}\XtoU\mS^4\XtoU^\top\left(\XtoU\mS^2\XtoU^\top\right)^{-1}\right)\right]\\
			&\ge \ex_{\XtoU} \left[\tr\left(\left(\XtoU\mS^2\XtoU^\top\right)^{-1}\XtoU\mS^2\XtoU^\top\left(\XtoU\XtoU^\top\right)^{-1}\XtoU\mS^2\XtoU^\top\left(\XtoU\mS^2\XtoU^\top\right)^{-1}\right)\right]\\
			&\ge \ex_{\XtoU} \left[\tr\left(\left(\XtoU\XtoU^\top\right)^{-1}\right)\right]\\
			&= \gamma(\mI_r)
		\end{align*}		
	\end{proof}
\end{proof}


\section{Experiments}
\label{asec:exps}
This section contains experimental details and also additional experiments on more datasets and settings.
The code for all experiments will be made public.

\subsection{RepLearn Algorithm and Datasets}
\label{asec:replearn}

\paragraph{RepLearn}
We describe the inner and outer loops for the RepLearn algorithm that we use for experiments in Algorithms~\ref{alg:inner} and \ref{alg:outer} respectively.
Recall the definitions for inner and outer losses.

\begin{algorithm}[tb]
   \caption{RepLearn($\repparam$, $TaskLoader$, $variant$)}
   \label{alg:outer}
\begin{algorithmic}
        \STATE {\bfseries Parameters:} Regularization ($\lambda$), Outer steps ($\outstep$), outer lr ($\outlr$), batch size ($b$), Inner steps ($\instep$), inner lr ($\inlr$)
        \STATE {\bfseries Input:} Representation model $\repparam$, $TaskLoader$, $variant$ \COMMENT{tr-tr or tr-val}
        \STATE $\repparam(0)$ $\leftarrow$ $RandInit$
        \STATE $ $
        \FOR{$t=0$ {\bfseries to} $\outstep-1$}
        		\STATE $TaskBatch$ $\leftarrow$ $TaskLoader(batch\_size=b)$
		\STATE $ $
		\FOR{$i=1$ {\bfseries to} $b$}
			\STATE $\Sto$ $\leftarrow$ $TaskBatch[i]$   \COMMENT{Dataset of size $n$}
			\STATE $ $
			\IF{$variant$ is `tr-val'}
				\STATE $(\Str,\Sva)$ $\leftarrow_{\text{split}}$ $\Sto$    \COMMENT{Split into tr-val sets of sizes $\ntr+\nva = n$}
				\STATE $\widetilde{\mW}$ $\leftarrow_{\text{stop gradient}}$ InnerLoop$(\repparam(t), \Str, \lambda)$ \COMMENT{Ignore the dependence of $\widetilde{\mW}$ on $\repparam(t)$}
				\STATE $\nabla_i$ $\leftarrow$ $\nabla_{\repparam}~\ell_{\rep}(\widetilde{\mW};\repparam,\Sva) \large|_{\repparam(t)}$   
				\STATE $ $
			\ELSIF{$variant$ is `tr-tr'}
				\STATE $\widetilde{\mW}$ $\leftarrow_{\text{stop gradient}}$ InnerLoop$(\repparam(t), \Sto, \lambda)$
				\STATE $\nabla_i$ $\leftarrow$ $\nabla_{\repparam}~\ell_{\rep}(\widetilde{\mW};\repparam,\Sto) \large|_{\repparam(t)}$   \COMMENT{Defined in \Eqref{aeqn:inner_loss}}
				\STATE $ $
			\ENDIF
			\STATE $ $
			\STATE $\nabla = \frac{1}{b}\sum_{i=1}^b \left[\nabla_i\right]$
			\STATE $\repparam(t+1) \leftarrow \textbf{Adam}(\repparam(t), \nabla, \inlr)$
		\ENDFOR
        \ENDFOR
        \STATE \textbf{return } $\repparam(\outstep)$
\end{algorithmic}
\end{algorithm}

\begin{algorithm}[!tb]
   \caption{InnerLoop($\repparam$, $\Sto$)}
   \label{alg:inner}
\begin{algorithmic}
        \STATE {\bfseries Parameters:} Regularization ($\lambda$), Inner steps ($\instep$), inner lr ($\inlr$)
        \STATE {\bfseries Input:} Representation model $\repparam$, $\Sto$
        \STATE $\mW(0)$ $\leftarrow$ $\vzero_{d\times k}$
        \FOR{$t=0$ {\bfseries to} $\instep-1$}
        		\STATE $\nabla \leftarrow \nabla_{\mW} \{\ell_{\rep}(\mW;\repparam,\Sto) + \frac{\lambda}{2}\|\mW\|_F^2\} \large|_{\mW(t)}$
		\STATE $\mW(t+1) \leftarrow  \textbf{SGD}(\mW(t), \nabla, \inlr, \text{momentum}=0.9)$
        \ENDFOR
        \STATE \textbf{return } $\mW(\instep)$
\end{algorithmic}
\end{algorithm}

\begin{align}
	\ell_{\rep}(\Clf; \repparam,\Sto) &= \ex_{(\vx,y)\sim \Sto} [\ell(\Clf^\top f_{\repparam}(\vx), y)]
	\label{aeqn:inner_loss}\\
	\ell_{\lambda,\rep}(\Clf; \repparam,\Sto) &= \ex_{(\vx,y)\sim \Sto} [\ell(\Clf^\top f_{\repparam}(\vx), y)]  +  \frac{\lambda}{2} \|\Clf\|_2^2
	\label{aeqn:inner_reg_loss}\\
	\Alg_{\lambda,\rep}(\repparam;\Sto) &= \argmin_{\Clf} \ell_{\lambda,\rep}(\Clf; \repparam,\Sto)
	\label{aeqn:inner_alg}
\end{align}

tr-tr Inner loop:
For dataset $\Sto$ and current initialization $\repparam_i$, run $\instep$ of gradient descent (with momentum 0.9) with learning rate $\inlr$ on $\ell_{\lambda,\rep}(\Clf; \repparam,\Sto)$ to get an approximation $\widetilde{\mW}_{\lambda,\rep}(\repparam;\Sto) \approx \Alg_{\lambda,\rep}(\repparam;\Sto)$.
Compute the gradient: $\gradtrva(\repparam) = \nabla \ell_{\lambda,\rep}(\widetilde{\mW}_{\lambda,\rep}\left(\repparam_i,\Sto); \cdot, \Sto\right) |_{\repparam_i}$

Outer loop: Run Adam with learning rate $\outlr$ (other parameters at default value) with batch size $b$ for $\outstep$ steps by using the gradient

Meta-testing: Tune $\instep$ and $\telambda$ using validation tasks

\paragraph{Datasets}
We conduct experiments on the Omniglot \citep{lake2015human}  and MiniImageNet \citep{vinyals2016matching} datasets. The Omniglot dataset consists of 1623 different handwritten characters from 50 different alphabets.  Each character was hand drawn by 20 different people. The original Omniglot dataset was split into a background set comprised of 30 alphabets and an evaluation set of 20 alphabets. We use the split recommended by  \citet{vinyals2016matching}, which contains of a training split of 1028 characters, a validation split of 172 characters, and test split of 423 characters. 
 \citet{vinyals2016matching}  construct the MiniImageNet dataset by sampling 100 random classes from ImageNet. We use 64 classes for training, 16 for validation, and 20 for testing.
We use torchmeta \citep{deleu2019torchmeta} to load datasets.
All our evaluations in meta-test time are conducted in the transductive setting.

\subsection{Omniglot Experiments}

\paragraph{Replearn on Omniglot}
We use a batch size $b=32$. We use the standard 4-layer convolutional backbone with each layer having 64 output filters followed by batch normalization and ReLU activations.
We resize the images to be 28x28 and apply 90, 180, and 270 degree rotations to augment the data, as in prior work, during training and evaluation.
We train for $\outstep=30000$ meta-steps and use $\instep=100$ inner steps regardless of model, and use an inner learning rate $\inlr=$ 0.05, unless there is a failure of optimization, in which case we reduce the learning rate to 0.01. We use an outer learning rate $\outlr=0.001$. We evaluate on 600 tasks at meta-test time.
At meta-test time, for each model, we pick the best $\telambda$ and inner step size based on the validation set, where we explore $\telambda \in [0,.1,.3, 1.0, 2.0, 3.0, 10.0, 100.0]$ and inner step size in $[50,100,200]$ and evaluate on the test set.

\paragraph{Increasing width of a FC network on Omniglot}
We examine the performance gap between tr-val versus tr-tr as we increase the expressive power of the representation network.
We use a baseline 4 hidden layer fully connected network (FCN) inspired by \citet{finn2017model}, but with $64 \ell$ nodes at all 4 layers for different values of $\ell\in\{1,4,8,16,32\}$. We set inner learning rates to be .05, except for $\ell \in \{16,32\}$ where we found a smaller inner learning rate of .01 was needed for convergence.

\paragraph{iMAML}
We use the original author code\footnote{\url{https://github.com/aravindr93/imaml_dev}} as a starting point, which creates a convolutional neural network with four convolutional layers, followed by Batch Normalization and ReLU activations. 
We modify their code to add a tr-tr variant by using the combined data for the inner loop and the outer loop updates.
We apply 90, 180, and 270 degree rotations to Omniglot data, resizing each image to 28x28 pixels. We use 5 conjugate gradient steps. 
For meta-testing we pick the best $\telambda \in \{0,.1,.3,1.0,2.0,3.0,10.0,100.0\}$ and $n_{\text{in}} \in. \{8,16,32, 64\}$ that maximizes accuracy on the validation set.
All other hyper-parameters at meta-test time are equal to the values used during training.
We use an outer learning rate of 1e-3.  We train for 30000 outer steps for all models tested, and set the number of inner steps n\_steps = 16 for 5-way 1-shot  and 25 for 20-way 1-shot. 
We investigate the performance of tr-val versus tr-tr by examining different settings of the regularization parameter $\lambda$. We report our results for tr-val versus tr-tr for Omniglot 5-way 1-shot and 20-way 1-shot in \Tableref{table:imaml}.
We find that tr-val significantly outperforms tr-tr in all settings, and the gap is much larger than for RepLearn.

\begin{center}
\begin{table}[!t]
\caption{meta-test accuracies in \% for a CNN trained with iMAML for tr-val versus tr-tr on Omniglot 5-way 1-shot and Omniglot 20-way 1-shot.
We find a huge gap in performance between tr-val and tr-tr, even larger than that for RepLearn.}
 \medskip
  \centering
  \small
\begin{tabular}{ c c c }
    \toprule
  & 5-way 1-shot & 20-way 1-shot \\ 
       \midrule
$\lambda=$ 2.0 tr-val &	97.90 $\pm$ 0.58   & 91.0 $\pm$ 0.54 \\  
\midrule
$\lambda=$ 10.0 tr-tr  &	49.22 $\pm$ 1.83 & 14.45 $\pm$  0.61 \\
$\lambda=$ 2.0 tr-tr  &	43.71 $\pm$ 1.92 & 16.18 $\pm$ 0.65\\
$\lambda=$ 1.0 tr-tr &	47.18 $\pm$ 1.90 & 16.96 $\pm$ 0.66\\
$\lambda=$ 0.3 tr-tr &	48.61 $\pm$ 1.90  & 16.50 $\pm$ 0.64 \\
$\lambda=$ 0.1 tr-tr  &	48.60 $\pm$ 1.93   & 17.70 $\pm$ 0.69\\
$\lambda=$ 0.0 tr-tr &	49.21 $\pm$ 1.92  &18.30 $\pm$ 0.67 
\end{tabular}
\label{table:imaml}
\end{table}
\end{center}

\begin{figure}
\includegraphics[scale=.16]{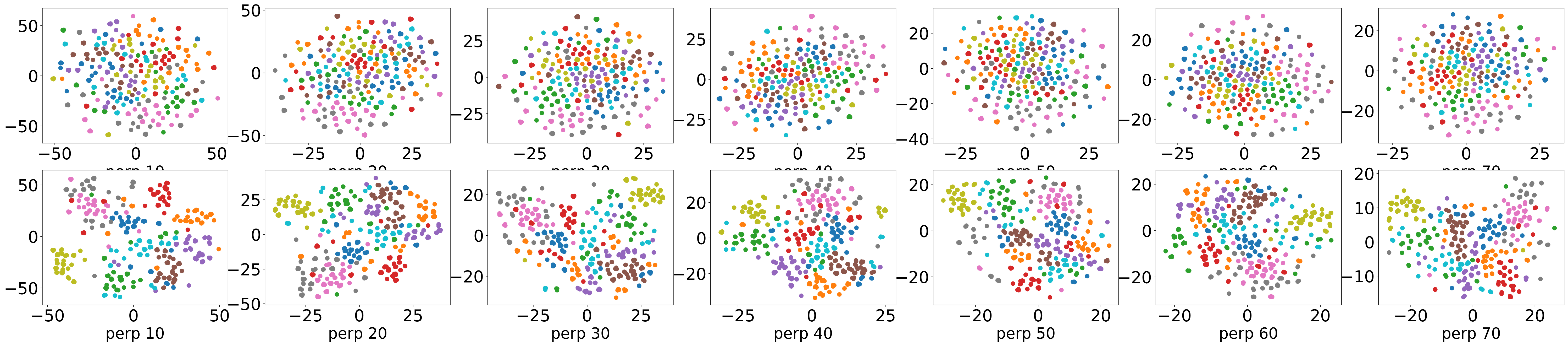}
\caption{t-SNE plots for tr-val versus tr-tr for varying values of the perplexity for a CNN model trained with iMAML  on Omniglot 5-way 1-shot. Plots shown for 10 randomly selected classes from the test split of Omniglot. Top row depicts tr-tr and bottom row depicts tr-val. We find that the tr-val representations appear to be more clustered for all values of perplexity of t-SNE.}
\label{fig:perpsiMAML}
\end{figure}

\begin{figure}[!t]
\centering
\begin{subfigure}[t]{0.22\textwidth}
\includegraphics[scale=.15]{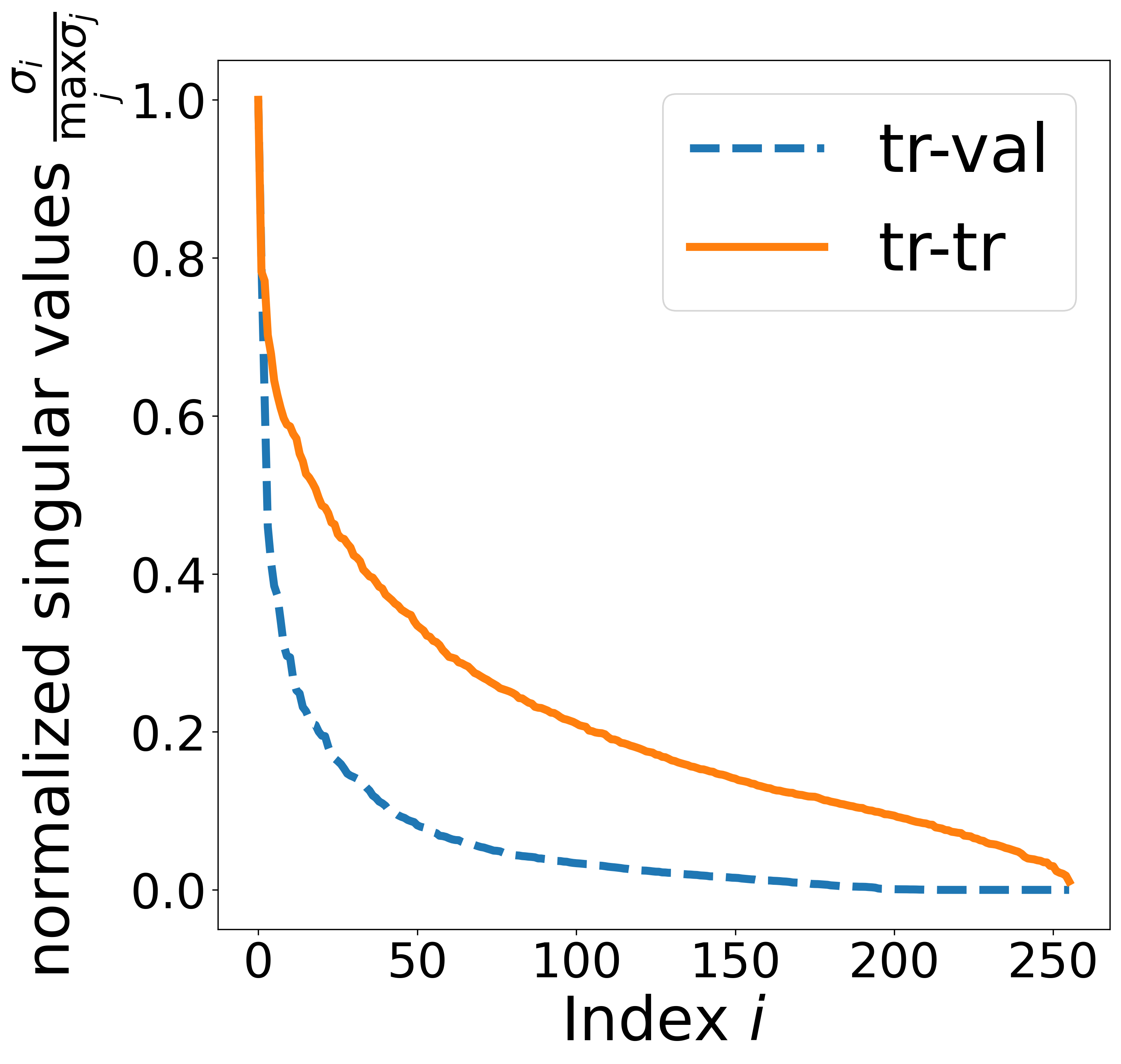}
 \caption{}
 \end{subfigure}%
 \begin{subfigure}[t]{0.22\textwidth}
\includegraphics[scale=.14]{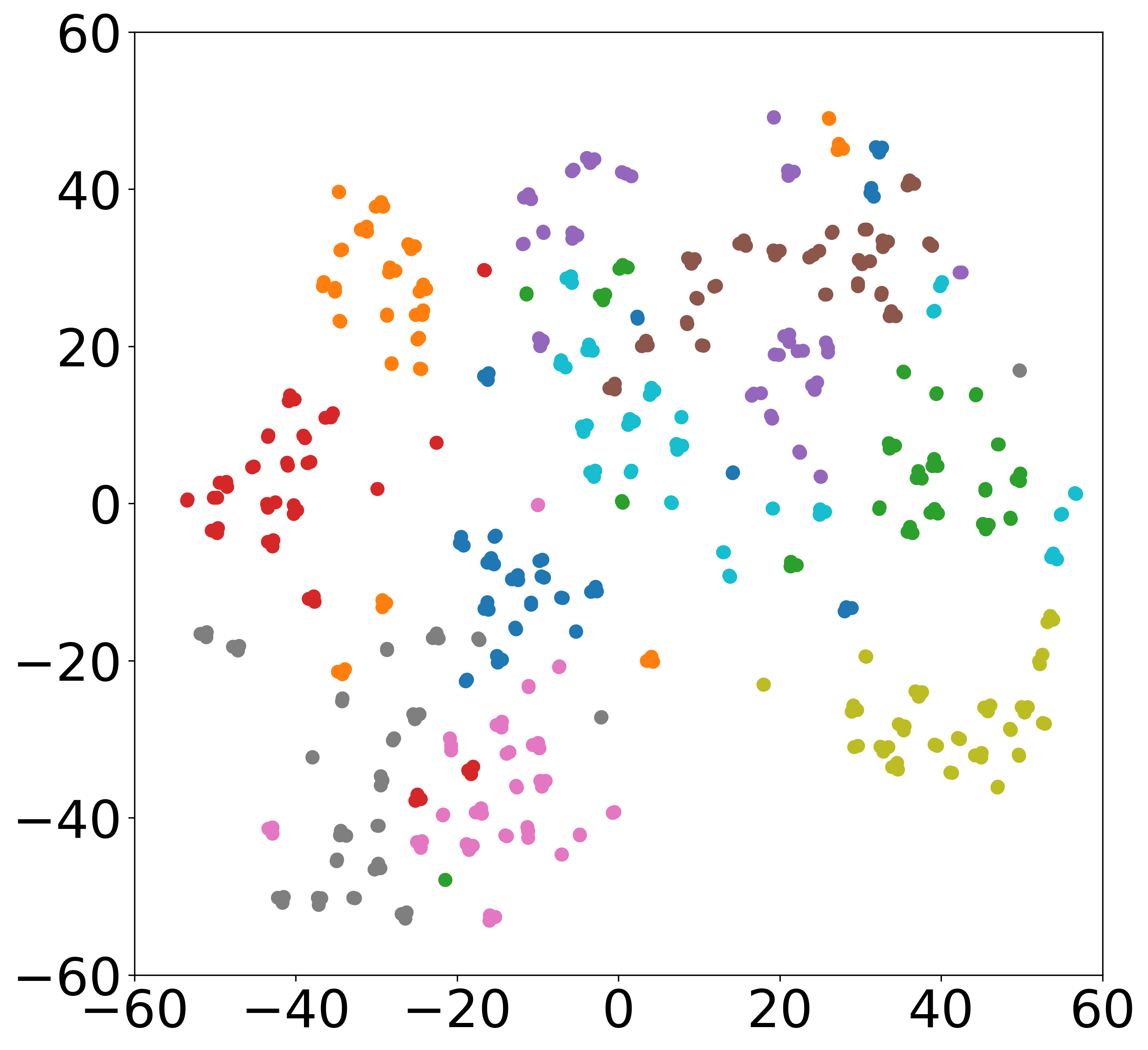}
 \caption{}
 \end{subfigure}%
 \begin{subfigure}[t]{0.22\textwidth}
\includegraphics[scale=.14]{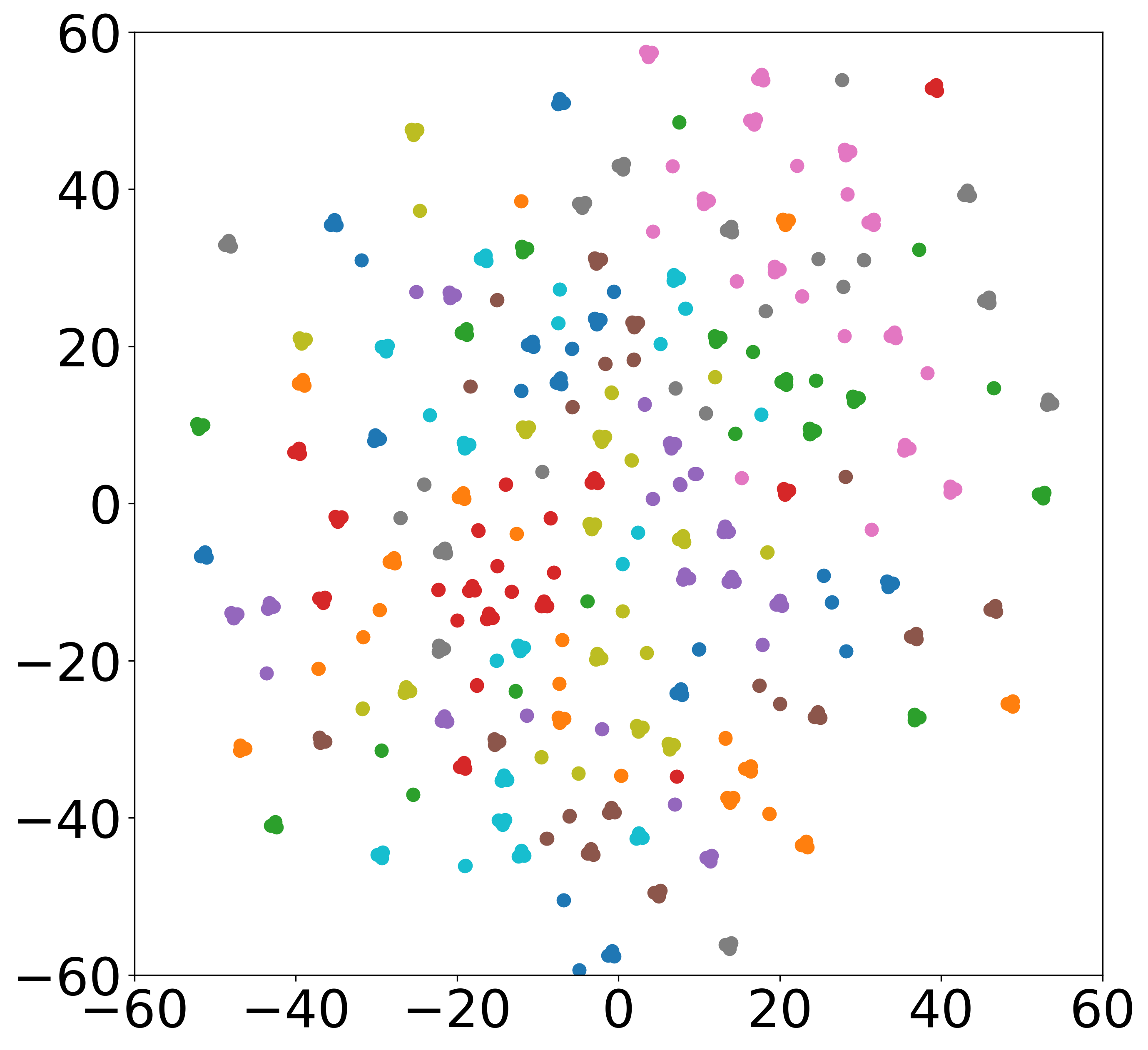}
 \caption{}
 \end{subfigure}%
 \begin{subfigure}[t]{0.26\textwidth}
\includegraphics[scale=.14]{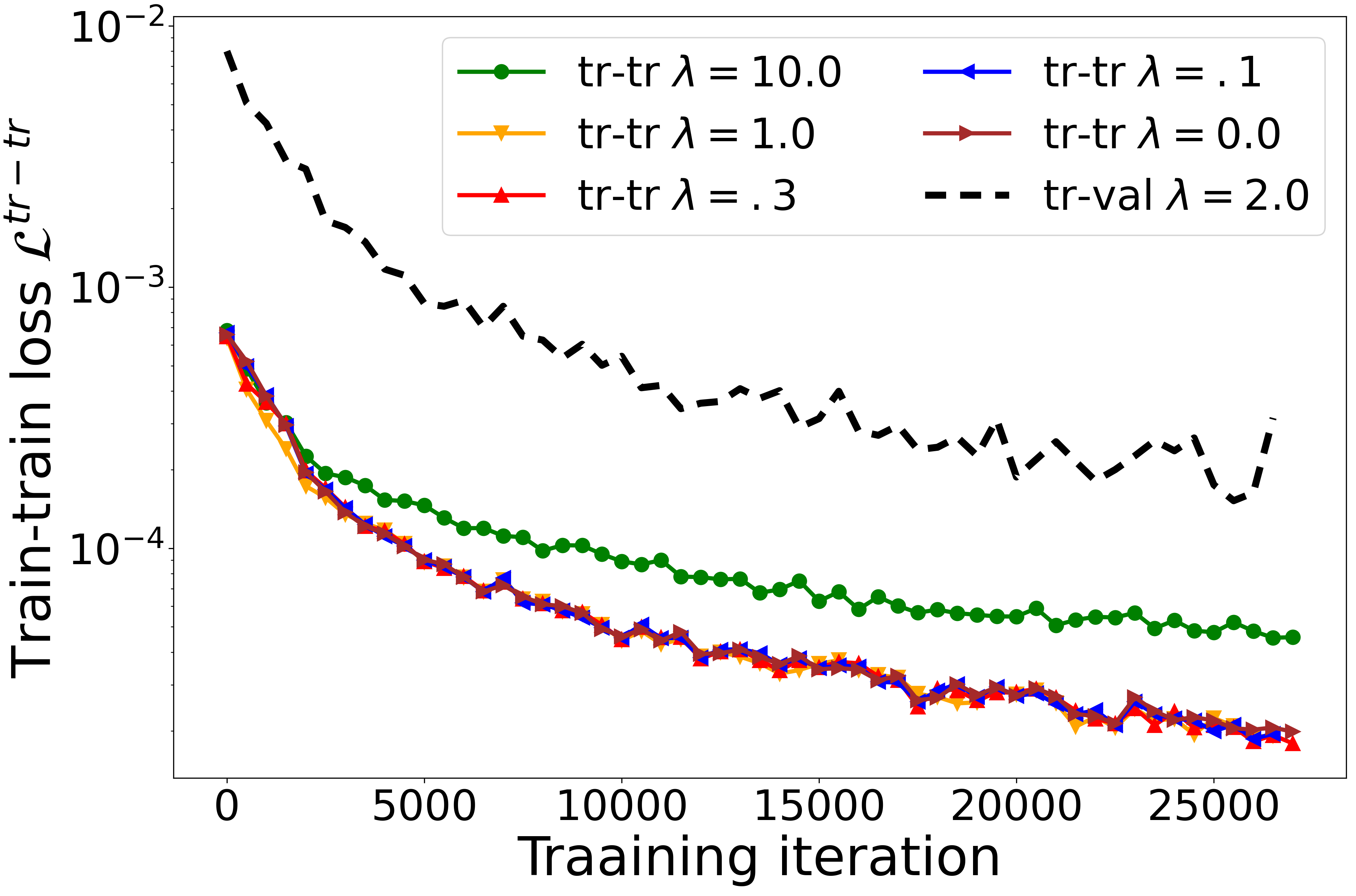}
 \caption{}
 \end{subfigure}%
\caption{Plot of singular values and t-SNE plots for iMAML tr-val versus tr-tr. \textbf{(a)}:  Singular values for representations produced by tr-val versus tr-tr of a CNN model trained with iMAML on Omniglot 5-way 1-shot.  \textbf{(b)}: t-SNE plot of representations produced by tr-val CNN model trained with iMAML on Omniglot 5-way 1-shot for 10 randomly selected classes of test split. \textbf{(c)}: t-SNE plot of representations produced by tr-tr CNN trained with iMAML model on Omniglot 5-way 1-shot for the same 10 randomly selected classes of test split. We find that the tr-val representations appear to be more clustered than the tr-tr representations.  \textbf{(d)}: Tr-tr loss values for tr-tr versus tr-val models.}
\label{fig:tsneiMAML}
\end{figure}

\paragraph{iMAML representations and t-SNE}
We train a model on Omniglot 5-way 1-shot using iMAML with batch size 32, inner learning rate .05, and meta steps 30000, again using the original author code convolutional neural network.  We use $\lambda=2.0$ for tr-val and $\lambda = 1.0$ for tr-tr. We use the tr-val and tr-tr CNN models to get image representations for input images, and then perform t-SNE on 10 randomly selected classes. 
We report t-SNE and singular value decay results for iMAML in \Figref{fig:tsneiMAML}.
We also plot the t-SNE representations for varying values of the perplexity in \Figref{fig:perpsiMAML}.
As in the case of RepLearn, the tr-val representations are much better clustered in the tr-tr representations.
Furthermore the tr-val representations have a sharper drop in singular values, suggesting that they have lower effective rank than tr-tr representations.

\paragraph{Rank and Expressivity}
For a fully connected model of width factor $\ell = 32$ trained with RepLearn on Omniglot 5-way 1-shot, we conduct linear regression twice. The first regression predicts the tr-tr representations given the tr-val representations, and the second predicts the tr-val representations given the tr-tr representations. We find that the $R^2$ scores were  $0.0973$ and $0.0967$, respectively. 
Thus the two sets of representations can express each other well enough, even though tr-val representations have lower effective rank.

We conduct the same experiment for a CNN model trained with iMAML on Omniglot 5-way 1-shot, and find that the $R^2$ scores were 0.0103 and 0.171, respectively, thus suggesting that tr-val representations are a bit more expressive than tr-tr representations, in addition to having lower effective rank.

\paragraph{Adding explicit regularization to RepLearn}
For a CNN model trained with RepLearn on Omniglot 5-way 1-shot, we add explicit regularization to tr-tr by adding the Frobenius norm of the representation to the loss. We report the accuracy in percentage evaluated on 1200 tasks  in \Tableref{table:withexpreg} in the top row.
We find that this significantly improves the performance of the tr-tr method, compared to the tr-tr models without explicit regularization in \Tableref{table:omni5w1s}, or the bottom row of \Tableref{table:withexpreg}.
This fits our intuition that the tr-tr method requires some form of regularization to learn low rank representations and to have guaranteed good performance on new tasks.

\subsection{MiniImageNet Experiments}

\paragraph{RepLearn on MiniImageNet}
As standard \citep{rajeswaran2019meta}, we resize the data to 84x84 pixel images and apply 90, 180, and 270 degree rotations and use a batch size of 16 during training. We use a convolutional neural network with four convolutional layers, with output filter sizes of 32,, 64, 128, and 128 each followed by batch normalization and ReLU activations.
We investigate the performance of tr-val versus tr-tr for RepLearn on the MiniImageNet 5-way 1-shot and 5-way 5-shot setting and report our results in \Tableref{table:miniImageNetrep}.
The findings are very similar to those from the Omniglot dataset, suggesting that our insights hold across multiple benchmark datasets.

\begin{center}
\begin{table}[!t]
\caption{Tr-val versus tr-tr meta-test accuracies in \% for a CNN model trained with RepLearn on MiniImageNet 5-way 1-shot  and 5-way 5-shot for varying values of the regularization parameter, $\lambda$. The final value of the tr-tr objective is depicted in the last two columns for MiniImageNet 5-way 1-shot and MiniImageNet 5-way 5-shot, respectively.
The tr-tr models make the tr-tr loss very small, which is what they were trained to minimize. Thus their failure on few-shot learning is not due to failure of optimization.}
 \medskip
  \centering
  \small
\begin{tabular}{ c | c | c | c | c}
    \toprule
  & 5-way 1-shot&  5-way 5-shot & tr-tr loss 5-way 1-shot & tr-tr loss 5-way 5-shot\\ 
      \midrule
$\lambda=$ 0.0 tr-val 	& 46.16 $\pm$ 1.67		& 65.36 $\pm$ 0.91 	& 0.01 & 0.01 \\  
\midrule
$\lambda=$ 0.0 tr-tr  		& 25.53 $\pm$ 1.43     	& 33.49 $\pm$ 0.82  	& 1.1e-8 & 2.1e-6\\
$\lambda=$ 0.1 tr-tr  		& 24.69 $\pm$ 1.32       	& 34.91 $\pm$ 0.85  	& 3.5e-8 & 5.5e-7\\
$\lambda=$ 1.0 tr-tr  		& 25.88 $\pm$ 1.45    	& 40.19 $\pm$ 1.12 	& 1.9e-6 & 9.3e-5\\
\end{tabular}
\label{table:miniImageNetrep}
\end{table}
\end{center}

\paragraph{Increasing capacity of a CNN model on MiniImageNet}
We start with a CNN model trained on MiniImageNet with four convolutional layers with 32, 64, 128, and 128 filters, respectively. Each convolution is followed by batch normalization and ReLU activations. We train with RepLearn. 
We increase the capacity by increasing the number of filters by a capacity factor, $\ell$, so that the convolutional layers contain $32 \ell$, $64 \ell$, $128 \ell$, and $128 \ell$ output filters, respectively.  We depict our results in \Tableref{table:widthtablecnn}.
We find that increasing the network capacity improves the performance of tr-val representations, but slightly hurts tr-tr performance, just like the findings for Omniglot dataset with fully-connected networks.
Thus the tr-val method is more robust to architecture choice/capacity and datasets.

\begin{figure}[!t]
\centering
\begin{subfigure}[t]{0.25\textwidth}
\includegraphics[scale=.15]{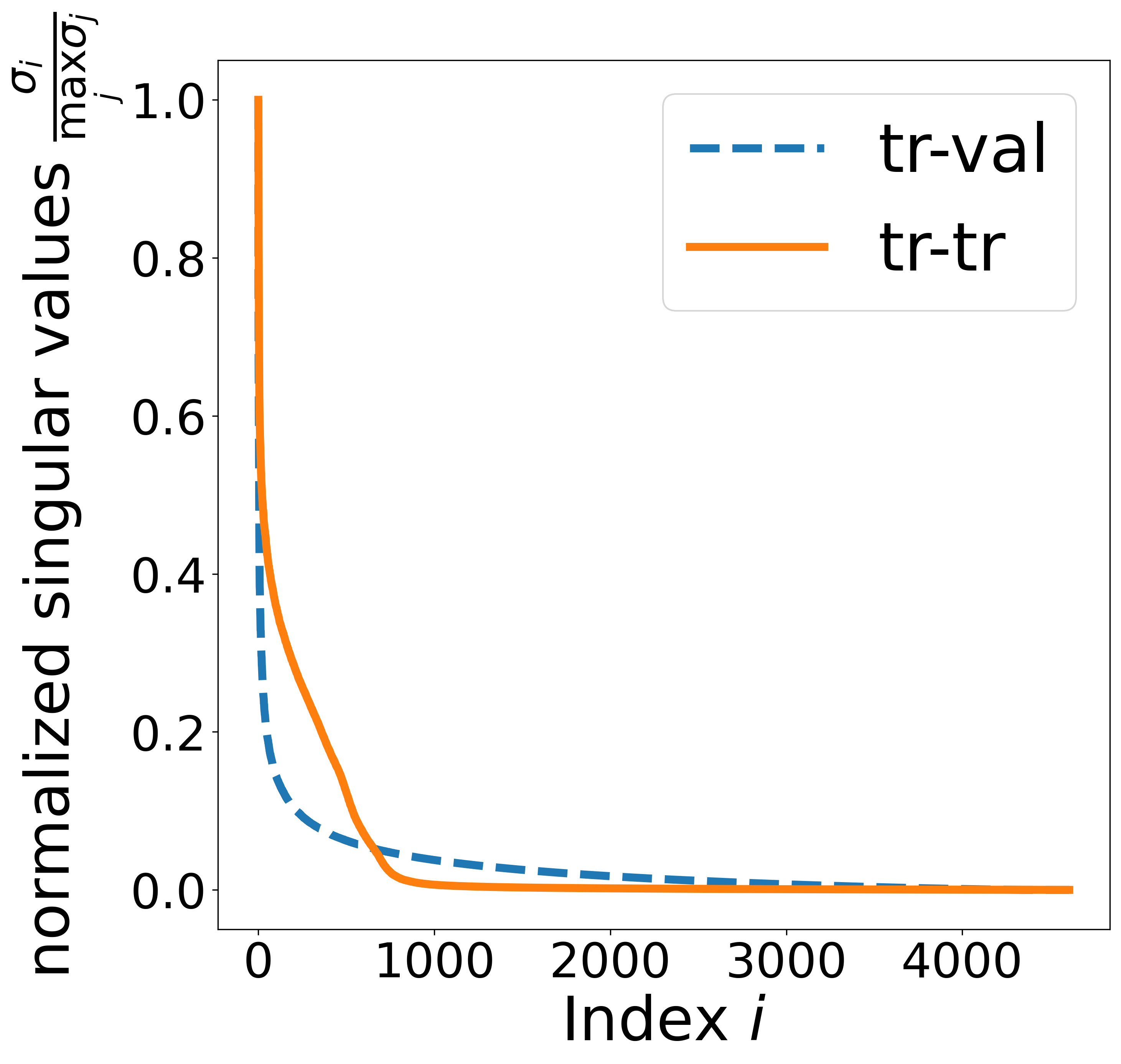}
 \caption{}
 \end{subfigure}%
 \begin{subfigure}[t]{0.25\textwidth}
\includegraphics[scale=.15]{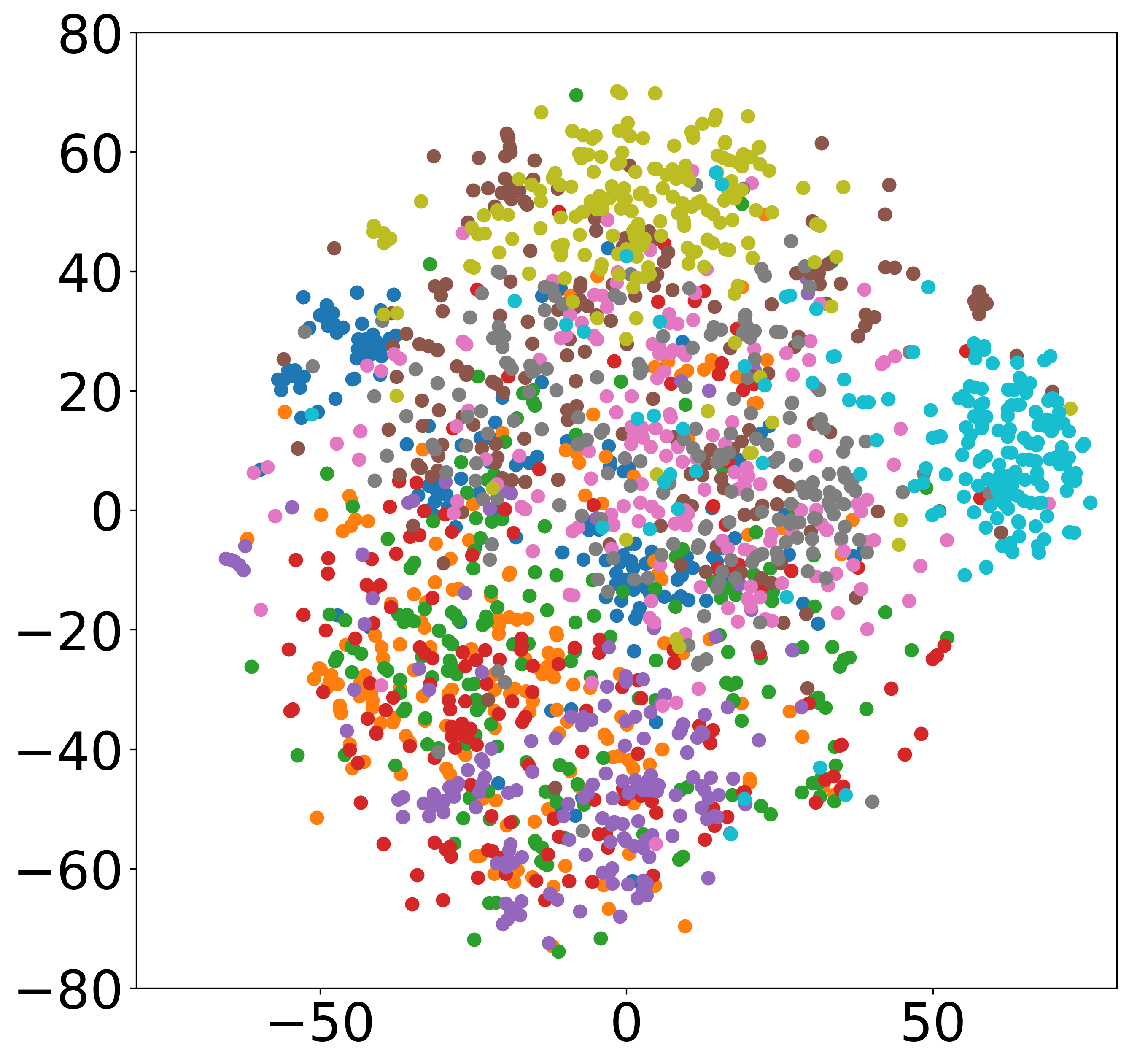}
 \caption{}
 \end{subfigure}%
 \begin{subfigure}[t]{0.25\textwidth}
\includegraphics[scale=.15]{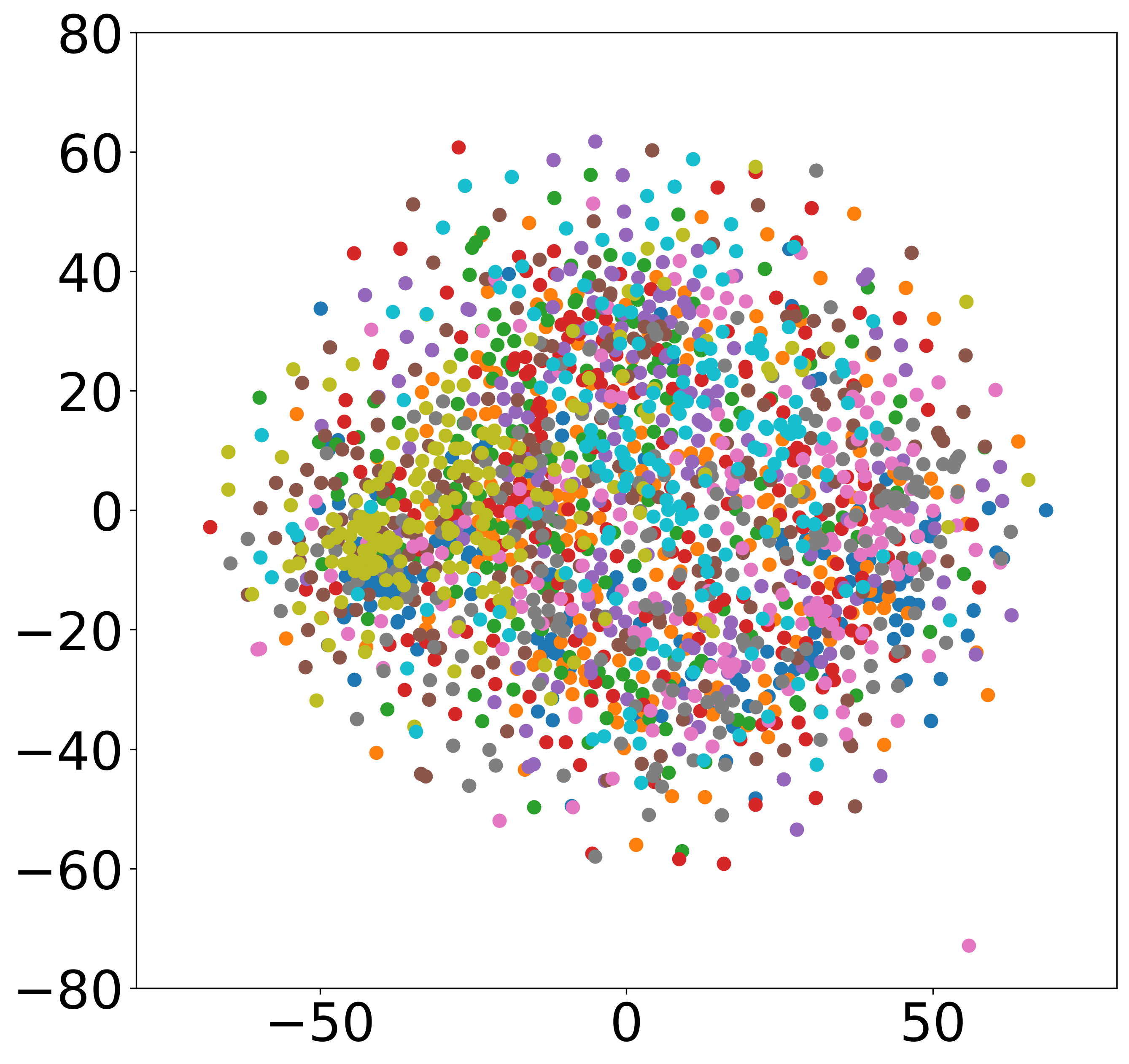}
 \caption{}
 \end{subfigure}%
\caption{Plot of singular values and t-SNE plots for MiniImageNet tr-val versus tr-tr. \textbf{(a)}:  Singular values for representations produced by tr-val versus tr-tr of a CNN model trained with RepLearn on MiniImagement. \textbf{(b)}: t-SNE plot of representations produced by tr-val CNN model trained with RepLearn on MiniImageNet for 10 randomly selected classes of test split. \textbf{(c)}: t-SNE plot of representations produced by tr-tr CNN model trained with RepLearn on MiniImageNet for the same 10 randomly selected classes of test split. We find that the tr-val representations appear to be more clustered than the tr-tr representations.  }
\label{fig:tsneminicnn}
\end{figure}

\paragraph{t-SNE on MiniImageNet}
We take the baseline convolutional neural network with capacity factor $\ell = 1$ from the previous section, and conduction t-SNE on the representations produced by the tr-val versus the tr-tr model. We report our results in \Figref{fig:tsneminicnn}.

\begin{center}
\begin{table}[!t]
\caption{Meta-test accuracies in \% for tr-tr RepLearn trained on Omniglot 5-way 1-shot with explicit Frobenius norm regularization added to the representations. }
 \medskip
  \centering
  \small
\begin{tabular}{ c | c | c | c }
    \toprule
  & $\lambda=$ 0.0 (tr-tr)		&  $\lambda=$ 0.1 (tr-tr)  		&  $\lambda=$ 3.0(tr-tr)\\ 
      \midrule
  with representation regularization & 94.73 $\pm$ 0.55  &  95.05 $\pm$ 0.55	&	94.74 $\pm$ 0.55 \\  
    no representation regularization  & 67.78 $\pm$ 1.60  &  67.53 $\pm$ 1.66	&	89.00 $\pm$ 1.08 \\  
\end{tabular}
\label{table:withexpreg}
\end{table}
\end{center}

\begin{table}[!t]
\caption{Accuracies in \% of representations parameterized by CNN networks of varying number of filters on MiniImageNet. Representations trained using tr-val objective consistently outperforms those learned using tr-tr objective, and the gap increases as width increases.}
\label{table:widthtablecnn}
    \medskip 
  \centering
  \small
\begin{tabular}{ c | c c | c c}
    \toprule
  capacity $=$	& \multicolumn{2}{c}{MiniImageNet 5-way 1-shot} & \multicolumn{2}{c}{Supervised 20-way} \\
  $num\_filters^*\ell$	& tr-val 			& tr-tr 			& tr-val 	& tr-tr\\
  \midrule
 $\ell = 0.5$ 	& 46.66  $\pm$ 1.69	  & 26.25 $\pm$ 1.45		& 1.		& 1.\\  
 $\ell = 1$ 		& 48.44 $\pm$ 1.62  	  & 26.81 $\pm$ 1.44		& 1.  		& 1.\\  
 $\ell = 4$ 		& 52.22 $\pm$ 1.68.	  & 24.66 $\pm$ 1.26		& 1. 		& 1.\\  
 $\ell = 8$ 		& 52.25 $\pm$ 1.71	  &25.28 $\pm$ 1.37		& 1.	 	& 1.\\  
 \bottomrule
\end{tabular}
\end{table}

\iftrue
\begin{table}[!t]
\caption{Tr-val versus tr-tr meta-test accuracies in \% for a CNN model trained with RepLearn on Omniglot 5-way 1-shot for varying values of the regularization parameter, $\lambda$.}
 \medskip
  \centering
  \small
\begin{tabular}{ c | c c }
    \toprule
  & 5-way 1-shot tr-val & 5-way 1-shot tr-tr \\ 
       \midrule
$\lambda=$ 0.0  & 97.25	$\pm$ 0.57    &  67.78 $\pm$ 1.60\\  
$\lambda=$ 0.1    & 97.34	$\pm$ 0.59  & 67.53 $\pm$ 1.64  \\
$\lambda=$ 0.3 & 97.59 $\pm$ 0.55 & 66.06 $\pm$ 1.67  \\
$\lambda=$ 1.0  &  97.66 $\pm$ 0.52  & 87.25 $\pm$ 1.13\\
$\lambda=$ 3.0  & 97.19 $\pm$ 0.59 & 89.00 $\pm$ 1.08\\
$\lambda=$ 10.0  & 96.50 $\pm$ 0.61 & 85.41 $\pm$ 1.22 \\
\end{tabular}
\label{table:omni5w1s}
\end{table}

\begin{table}[!t]
\caption{Tr-val versus tr-tr meta-test accuracies in \% for a CNN model trained with RepLearn on Omniglot 20-way 1-shot for varying values of the regularization parameter, $\lambda$.}
 \medskip
  \centering
  \small
\begin{tabular}{ c | c c }
    \toprule
  & 20-way 1-shot trval & 20-way 1-shot tr-tr \\ 
       \midrule
$\lambda=$ 0.0  &92.26 $\pm$ 0.45  	&  49.00 $\pm$ 0.88 \\  
$\lambda=$ 0.1    & 92.38 $\pm$ 0.44   	& 50.84 $\pm$ 0.85  \\
$\lambda=$ 0.3 &  92.21 $\pm$ 0.47		& 55.38 $\pm$ 0.92  \\
$\lambda=$ 1.0  &  92.44 $\pm$ 0.47	&84.14 $\pm$ 0.63\\
$\lambda=$ 3.0  & 92.70 $\pm$ 0.48 	&  88.20 $\pm$ 0.55\\
$\lambda=$ 10.0  & 91.50 $\pm$ 0.48 	&85.85 $\pm$ 0.58 \\
\end{tabular}
\end{table}

\fi

\end{document}

\fi

